\documentclass[10pt,twocolumn,letterpaper]{article}

\usepackage[pagenumbers]{cvpr} 

\usepackage{verbatim}

\usepackage[dvipsnames]{xcolor}
\usepackage[normalem]{ulem} 
\newcommand*{\dif}{\mathop{}\!\mathrm{d}}
\newcommand*{\vect}[1]{\boldsymbol{#1}}
\newcommand*{\mat}[1]{\textbf{#1}}

\usepackage{setspace}
\usepackage{array}
\usepackage{multirow}
\usepackage{graphicx}
\usepackage{booktabs}
\usepackage{pgffor}
\usepackage{adjustbox}
\usepackage{xifthen}
\usepackage{tikz}
\usepackage{pgfplots}
\pgfplotsset{compat=1.18}

\newcommand{\imagecell}[2][0.12]{%
    \begin{minipage}[b]{#1\columnwidth} 
        \raisebox{-0.5\height}{\includegraphics[width=\linewidth]{#2.png}}%
    \end{minipage} 
}

\definecolor{cvprblue}{rgb}{0.21,0.49,0.74}
\usepackage[pagebackref,breaklinks,colorlinks,citecolor=cvprblue]{hyperref}

\def\paperID{199} 
\def\confName{3DV\xspace}
\def\confYear{2026\xspace}

\title{Recovering 3D Shapes from Ultra-Fast Motion-Blurred Images}

\author{
    Fei Yu\textsuperscript{1}, 
    Shudan Guo\textsuperscript{1}, 
    Shiqing Xin\textsuperscript{1}, 
    Beibei Wang\textsuperscript{2}, 
    Haisen Zhao\textsuperscript{1\dag},
    Wenzheng Chen\textsuperscript{3}
    \\
    \textsuperscript{1}Shandong University \quad
    \textsuperscript{2}Nanjing University \quad
    \textsuperscript{3}Peking University \quad
}

\begin{document}

\maketitle
\begin{abstract}

We consider the problem of 3D shape recovery from ultra-fast motion-blurred images.
While 3D reconstruction from static images has been extensively studied, recovering geometry from extreme motion-blurred images remains challenging.
Such scenarios frequently occur in both natural and industrial settings, such as fast-moving objects in sports (\eg, balls) or rotating machinery, where rapid motion distorts object appearance and makes traditional 3D reconstruction techniques like Multi-View Stereo (MVS) ineffective.

In this paper, we propose a novel inverse rendering approach for shape recovery from ultra-fast motion-blurred images.
While conventional rendering techniques typically synthesize blur by averaging across multiple frames, we identify a major computational bottleneck in the repeated computation of barycentric weights. To address this, we propose a fast barycentric coordinate solver, which significantly reduces computational overhead and achieves a speedup of up to \textbf{4.57$\times$}, enabling efficient and photorealistic simulation of high-speed motion.
Crucially, our method is fully differentiable, allowing gradients to propagate from rendered images to the underlying 3D shape, thereby facilitating shape recovery through inverse rendering.

We validate our approach on two representative motion types: rapid translation and rotation. 
Experimental results demonstrate that our method enables  efficient and realistic modeling of ultra-fast moving objects in the forward simulation. 
Moreover, it successfully recovers 3D shapes from 2D imagery of objects undergoing extreme translational and rotational motion, advancing the boundaries of vision-based 3D reconstruction.
Project page can be found at \url{https://maxmilite.github.io/rec-from-ultrafast-blur/}.

\end{abstract}

\vspace{-15pt}

\renewcommand{\thefootnote}{\dag}\footnotetext{Corresponding author.}

\section{Introduction}
\label{sec:intro}

Estimating the shape of an object from image collections is crucial for numerous applications, including film production, gaming, and AR/VR.
As a long-standing goal in computer vision and graphics, extensive research has leveraged geometric and learning-based priors for object shape recovery~\cite{hartley2003multiple,furukawa2015multi,kato2018neural,DIBR19,liu2019softras}.
However, most existing methods focus on static objects or those with low-speed motion~\cite{zhang2022deep,chen2024deblur,lu2025bard}, leaving shape recovery of high-speed objects largely underexplored. 

\begin{figure}
    \centering
    \begin{spacing}{1}
    \setlength\tabcolsep{0pt}
    \begin{tabular}{cccc}
    %

    \rotatebox[origin=c]{90}{Translation} \hspace{2pt} &
    \imagecell[0.3]{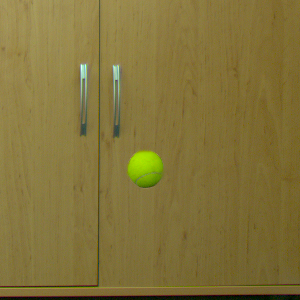} & 
    \imagecell[0.3]{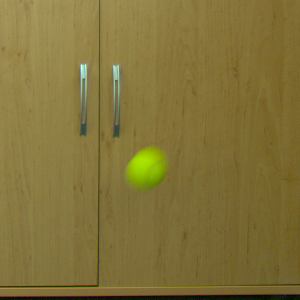} & 
    \imagecell[0.3]{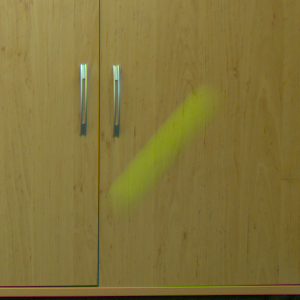} \\
    \vspace{-9pt} \\

    \rotatebox[origin=c]{90}{Rotation} \hspace{2pt} &
    \imagecell[0.3]{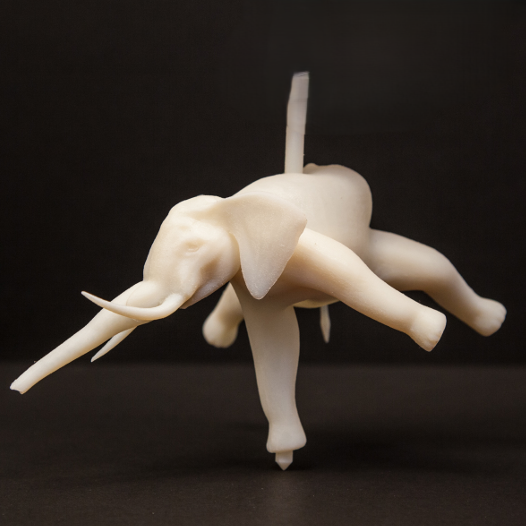} & 
    \imagecell[0.3]{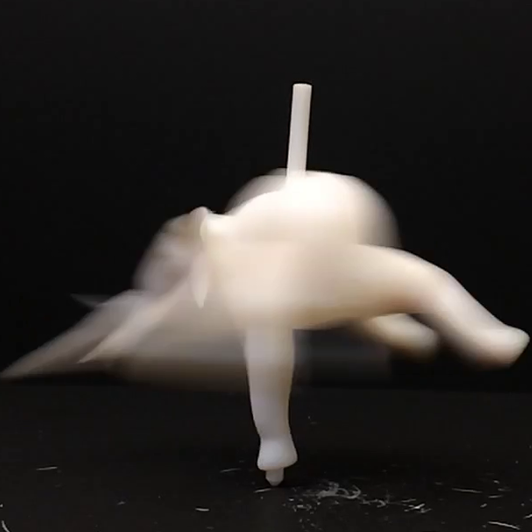} & 
    \imagecell[0.3]{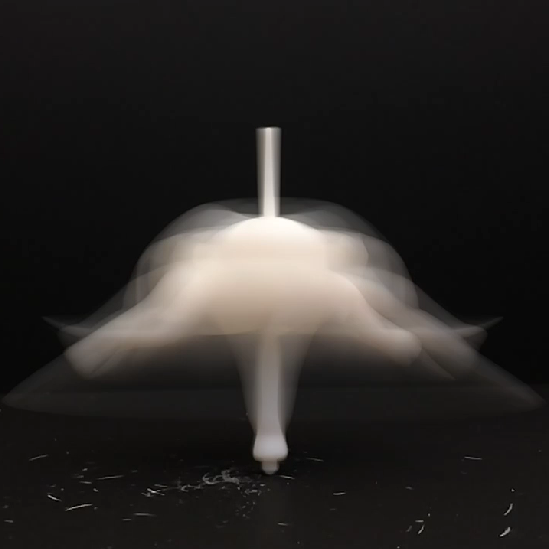} \\  
    \vspace{-9pt} \\
    &  Static &  Low-speed &  High-speed
    \end{tabular}
    \end{spacing}
    \vspace{-7pt}
    \caption{
        \textbf{Ultra-fast motion blur} is common in real-world scenarios. 
       Top: A ball undergoing translational motion \cite{rozumnyi2021defmo}. 
       Bottom: A spinning top in rotation \cite{bacher2014spin}. 
        In this paper, our goal is to recover 3D shapes from the high-speed translational and rotational motion.
    }
    \label{fig:teaser-1}
    \vspace{-20pt}
\end{figure}

In this work, we investigate an extremely challenging question: Can we recover the shape of an object undergoing ultra-fast motion?
Fast motion is prevalent in real-world scenarios, such as flying balls in sports, rotating machinery, or high-speed robotics. 
While reducing exposure time can mitigate blur, it often leads to extremely low signal-to-noise ratios in low-light conditions, making motion blur physically unavoidable in many practical scenarios.
However, extreme motion blur severely distorts the object's appearance, often obscuring the underlying shape.
As shown in \cref{fig:teaser-1}, the object's shape is barely perceptible in the captured blurry images, making traditional multi-view geometry-based methods, such as Structure from Motion (SfM)~\cite{hartley2003multiple,furukawa2015multi}, ineffective.
SfM-based techniques rely on sharp feature correspondences across views, but when motion blur obscures these features, shape recovery becomes highly challenging.

\begin{figure}
    \centering
    \begin{spacing}{1}
    \setlength\tabcolsep{0pt}
    \begin{tabular}{ccc}
    \imagecell[0.3]{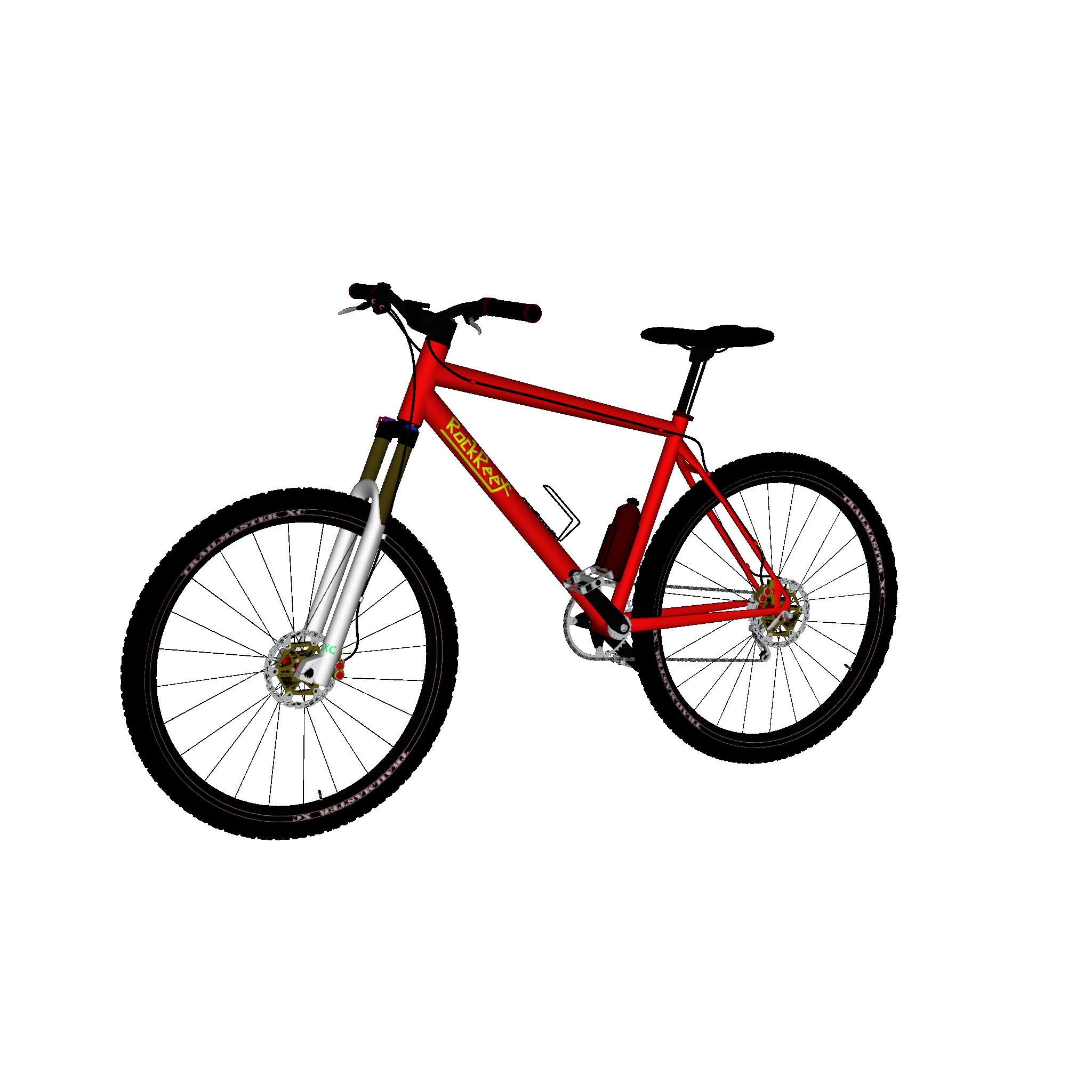} 
     & 
    \imagecell[0.3]{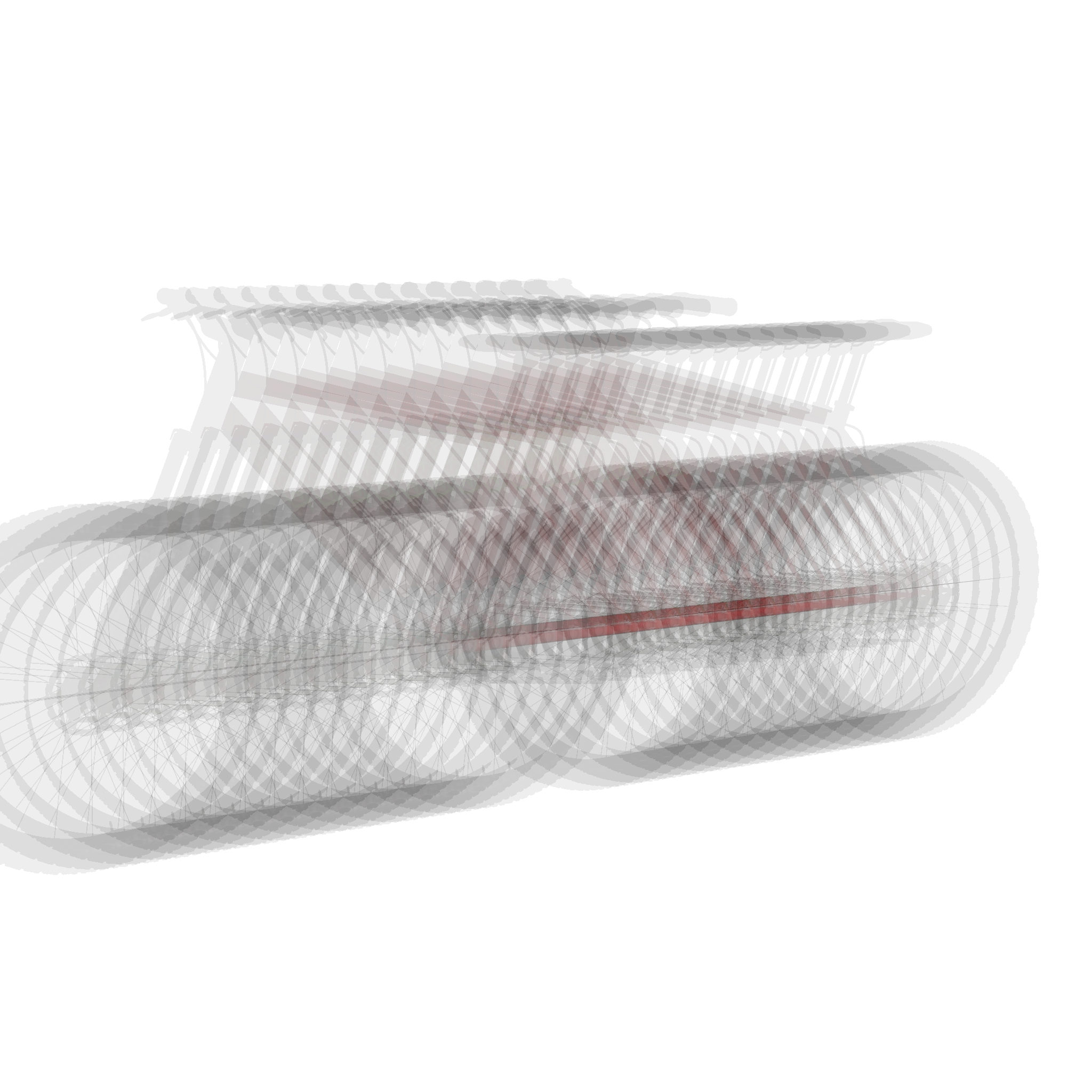} 
     & 
    \imagecell[0.3]{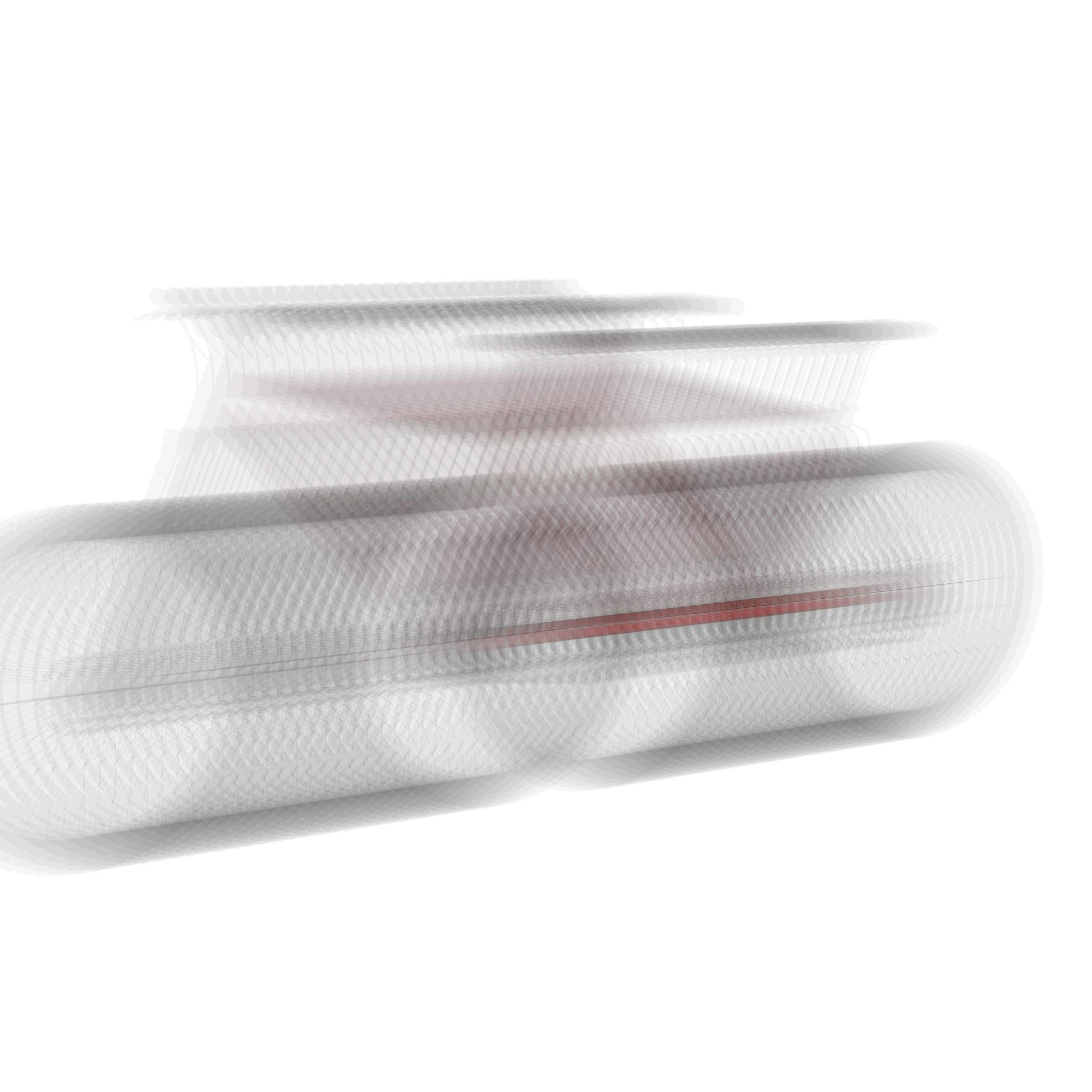} \\ 
    (a) Static & (b) 15 Samples & (c) 30 Samples \\
    \imagecell[0.3]{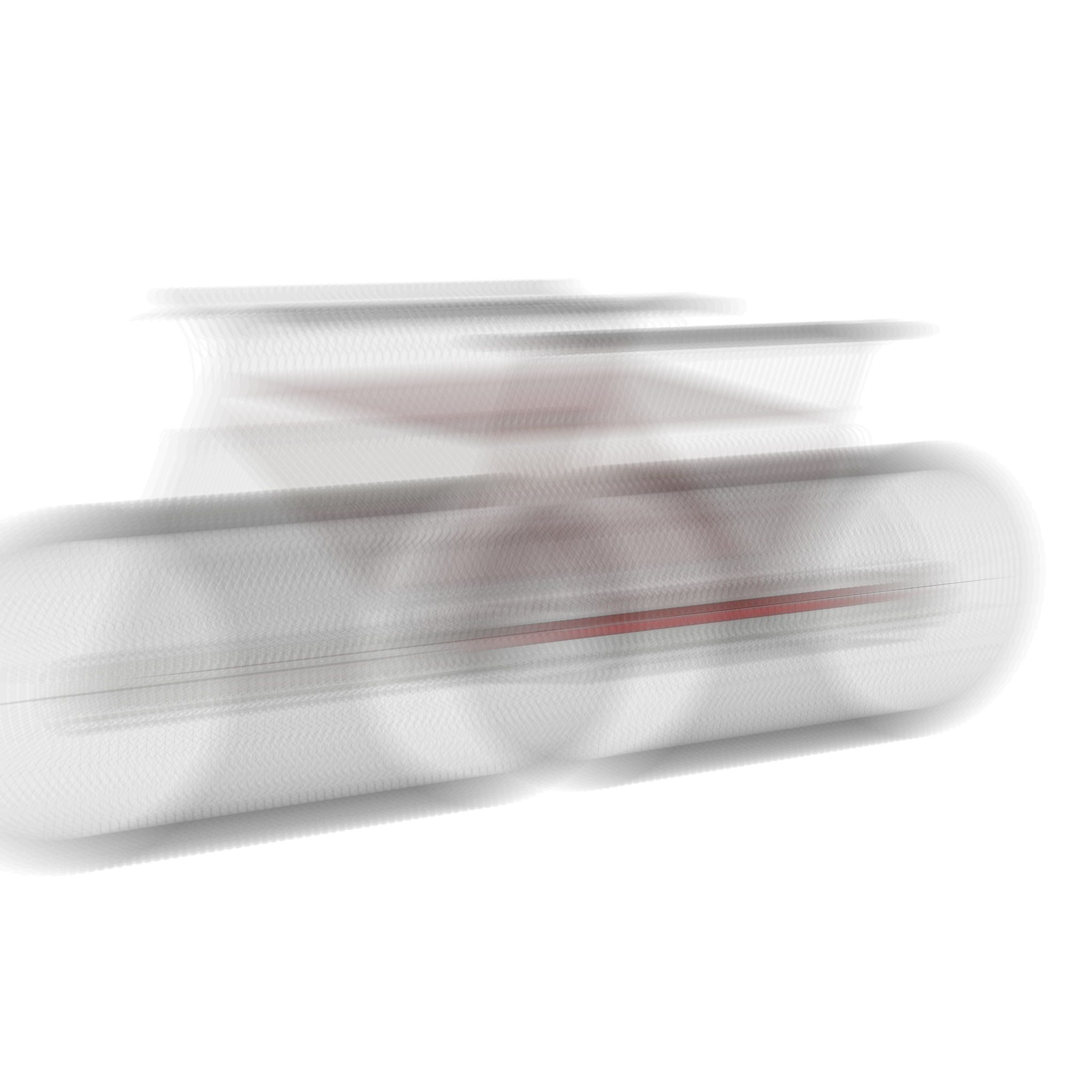} 
     & 
    \imagecell[0.3]{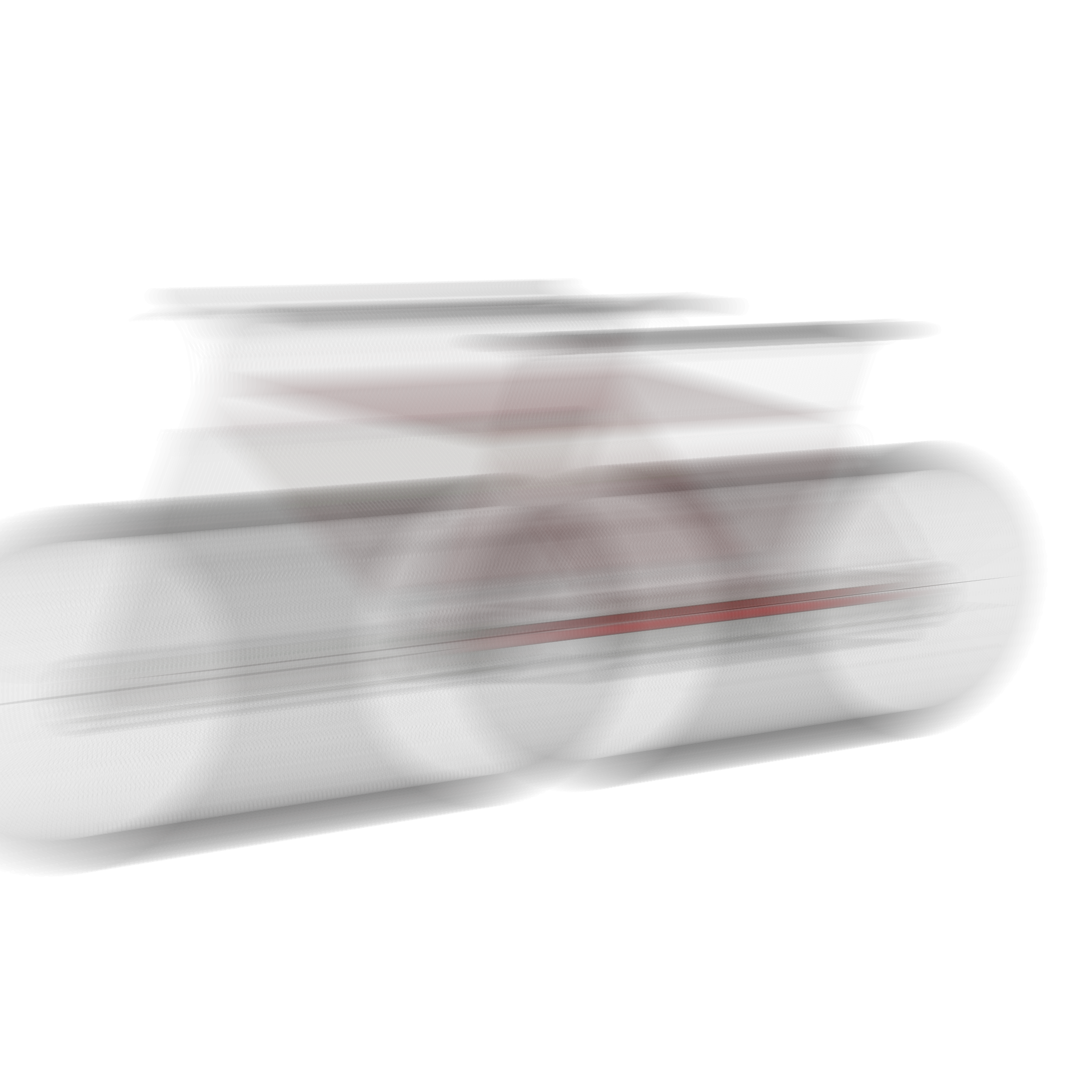} 
     & 
    \imagecell[0.3]{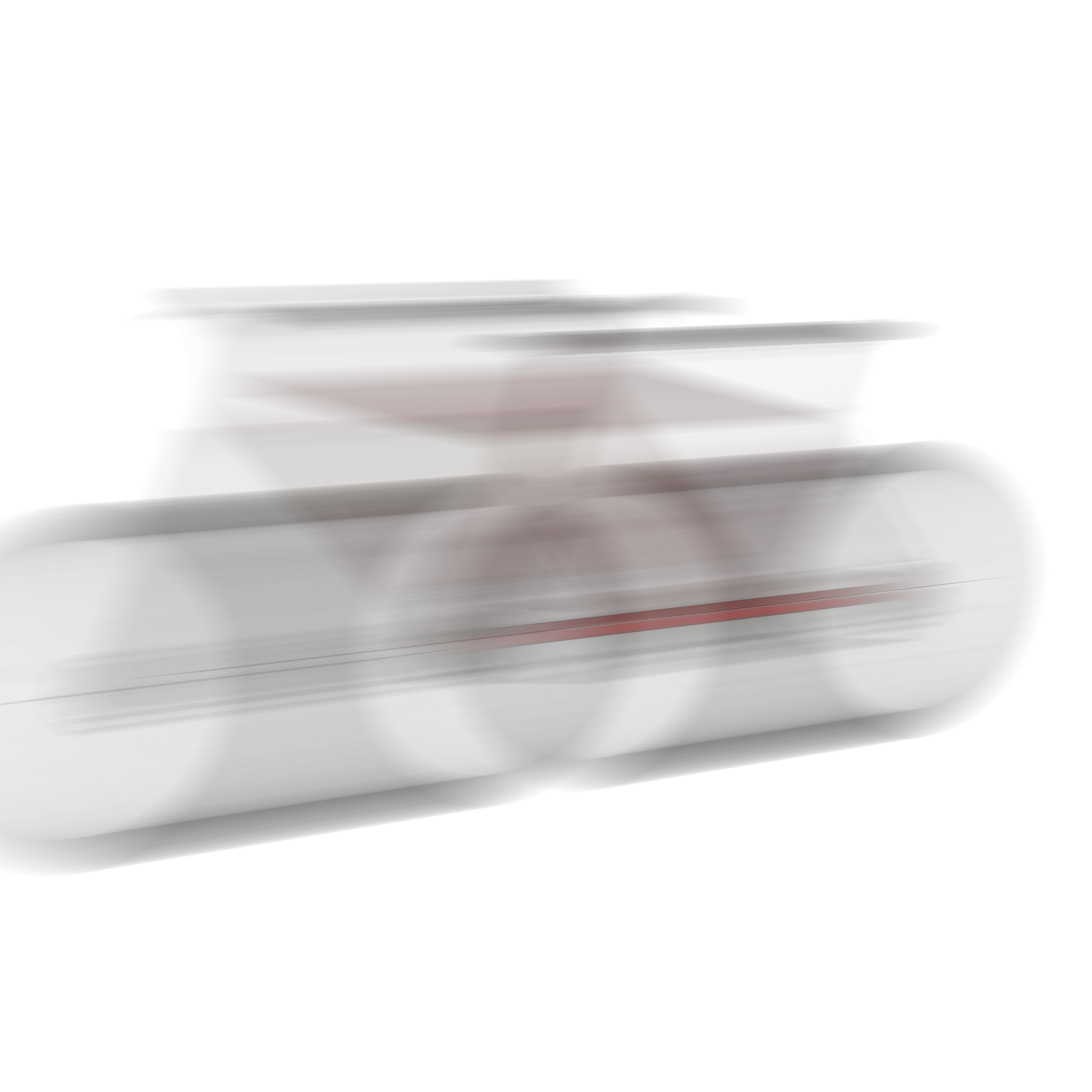} \\      
    (d) 50 Samples & (e) 100 Samples & (f) 180 Samples
    \end{tabular}
    \end{spacing}
    \vspace{-7pt}
    \caption{
            Motion blur is typically synthesized by rendering and averaging multiple frames. However, for extreme motion, a large number of frames are required to achieve realistic results. Here, we illustrate a bicycle undergoing extreme translation. Noticeable artifacts appear when using fewer samples, and at least 50 frames are needed to produce a realistic motion-blurred image.
        }
    \label{fig:teaser-2}
    \vspace{-15pt}
\end{figure}

\vspace*{1pt}
Alternatively, recovering 3D shape from 2D image collections can be formulated as an inverse problem, where the objective is to optimize the shape so that its renderings match the observed images~\cite{liu2019softras}.
Leveraging this paradigm, we formulate shape recovery under ultra-fast motion as an inverse rendering problem, where both geometry and appearance are estimated by simulating the blurry process.
Typically, motion blur can be approximated by rendering multiple static frames and averaging them~\cite{rozumnyi2021defmo,rozumnyi2021shape,rozumnyi2022motion,spetlik2024single}.
However, this method becomes computationally expensive for ultra-fast translational or rotational motions.
As shown in \cref{fig:teaser-2}, generating realistic motion blur under such conditions requires synthesizing and averaging over 50 individual static frames per blurry image, leading to excessive rendering costs and memory consumption.

\vspace*{1pt}
We carefully analyze the computational bottleneck in motion blur synthesis.
While a single barycentric computation is inexpensive, we identify that the repetitive calculation of these weights required for temporal integration becomes a primary source of inefficiency.
This is because barycentric weights must be computed for every pixel with respect to all triangles, and the synthesis of motion blur further amplifies the computational cost by requiring these computations across all sampled frames, leading to a significant overhead.  
To address this issue, inspired by analytic motion approximation techniques~\cite{gribel2010analytical}, we propose a fast barycentric coordinate solver that significantly reduces computational complexity.  
By integrating this solver into our differentiable rasterization framework, our approach achieves significant speedup while preserving the accuracy of motion blur simulation.  
Furthermore, we reformulate the rendering process in a soft, fully differentiable manner, allowing gradients to propagate through motion-blurred images to the underlying 3D shapes. 

\vspace*{1pt}
With its differentiable capabilities, our framework enables 3D shape recovery through an inverse rendering pipeline:
Beginning with an initial 3D shape, we render motion-blurred images and compare them with the observed ground-truth (GT) images.
The shape is then iteratively refined by minimizing the discrepancy between the rendered and GT images.
This analysis-by-synthesis approach allows for shape recovery from multi-view blurred images, even under extreme translational and rotational motion.

\vspace*{1pt}
We evaluate our method on a wide range of testing cases across various shapes and categories. 
Our method successfully recovers shapes from heavily blurred images caused by ultra-fast motion.
Additionally, we demonstrate 3D shape recovery from real-world motion-blurred images, showcasing the effectiveness of our method in challenging real-world scenarios.
Our work pushes the boundaries of 3D recovery from  ultra-fast motion-blurred images.

\vspace{-5pt}

\section{Related Work}
\label{sec:related-work}

\vspace{-3pt}

\subsection{General Deblurring Methods}

\vspace{-3pt}

Motion blur arises when multiple scene contents are projected onto the same pixel due to motion during image capture~\cite{zhang2022deep}. This blur can originate from various sources, including camera motion, object motion, or long exposures in low-light conditions. Typically, motion blur is modeled as a convolution of a clean image with a blur kernel. Numerous methods have been developed to address this issue, leveraging various priors such as total variation (TV) and phase information~\cite{pan2019phase}, deep neural networks~\cite{jin2018learning, pan2020cascaded, zhong2020efficient, shen2020blurry, zhou2019spatio, jin2019learning}, generative adversarial networks~\cite{kupyn2018deblurgan, Kupyn_2019_ICCV, zhang2020deblurring}, and, more recently, diffusion models~\cite{chung2022diffusion, spetlik2024single, chung2023parallel, song2023solving, bai2024blind}. However, these methods primarily focus on low-speed motion, where the blur kernels remain relatively small. Furthermore, most approaches are confined to 2D image space, making them ineffective for handling more complex, non-linear motion patterns, such as rotations.

\vspace{-3pt}

\subsection{Shape Recovery from Blurry Images or Videos}

\vspace{-3pt}

Our objective is to recover 3D shapes from blurry images of objects undergoing extremely fast motion~\cite{gribel2010analytical, shkurko2017time}. 
Objects exhibiting motion blur are often categorized as Fast Moving Objects (FMOs)~\cite{rozumnyi2017world}. 
Prior works have explored reconstructing both shape and motion from images or videos~\cite{kotera2019intra, rozumnyi2020sub, kotera2020restoration}.  
\citet{rozumnyi2021shape, rozumnyi2022motion} pioneered methods to recover 2D and 3D shapes along with motion from blurry images and videos.
These approaches effectively leverage neural network-based learned priors, such as the DeFMO network \cite{rozumnyi2021defmo}, to predict per-timestamp static silhouettes of the object. 
Combined with differentiable rendering techniques~\cite{DIBR19, liu2019softras}, they jointly estimate shape and motion via optimization, enabling robust solutions across a broad range of practical scenarios.

While impressive results are achieved, these methods also largely rely on accurately estimating object motion and static silhouettes, which might struggle with challenging, ultra-fast motion inputs.
In this scenario, the resulting blur introduces an unprecedented level of visual ambiguity, making the accurate recovery of static object silhouettes particularly challenging for ~\cite{rozumnyi2021defmo}.
In contrast, our method enables shape recovery under significantly more extreme high-speed motion conditions, demonstrating new capabilities in reconstructing objects undergoing ultra-fast motion.

\vspace{-3pt}

\subsection{Inverse Rendering}

\vspace{-3pt}

Shape recovery through inverse rendering has made rapid progress in recent years. A variety of 3D representations, including meshes, neural radiance fields (NeRF)~\cite{mildenhall2020nerf}, and Gaussian splatting~\cite{kerbl20233d}, have been combined with differentiable rendering techniques such as mesh rendering~\cite{kato2018neural, DIBR19, liu2019softras, laine2020modular}, volume rendering~\cite{drebin1988volume}, and surface rendering~\cite{pfister2000surfels} to jointly estimate shape, texture, lighting, and material directly from images~\cite{Munkberg_2022_CVPR, wang2021neus}. However, these methods are generally designed for clean, static images, and their applicability to motion-blurred scenes remains limited. 
In this paper, we propose a differentiable, rasterization-based renderer specifically designed to handle ultra-fast motion. Our approach extends inverse rendering to extreme motion-blurred conditions, making it a promising solution for high-speed shape recovery.

\section{Method}
\label{sec:methodology}

We now describe our method.  
We first provide the preliminaries of traditional rasterization algorithms in \cref{sec:pre} and analyze their computational bottleneck.  
We then present our solution: a fast barycentric coordinate solver in \cref{sec:solver}.  
With this new solver, we detail our differentiable motion-blur rendering algorithm in \cref{sec:render}.  

\subsection{Preliminaries of Rasterization}
\label{sec:pre}

Rasterization is a fundamental rendering technique that projects 3D triangle meshes onto a 2D image plane.  
Typically, it operates on a per-pixel basis by computing its barycentric coordinates with respect to each triangle.  
For a screen pixel $\vect{p}_i$ and a projected triangle face $\mat{F}_j$ with three vertices $\begin{bmatrix} \vect{v}_0 & \vect{v}_1 & \vect{v}_2  \end{bmatrix}$,  
we denote the barycentric coordinates of $\vect{p}_i$ with respect to $\mat{F}_j$ as $\vect{w} = \begin{bmatrix} w_0 & w_1 & w_2 \end{bmatrix}^T$,  
which satisfies the equation:
\begin{equation}
   \vect{p}_i = w_0 \vect{v}_0 + w_1 \vect{v}_1 + w_2 \vect{v}_2.
\end{equation}

The vector $\vect{w}$ is then used to interpolate vertex attributes, such as colors or texture UV mappings.  

Traditional differentiable rasterizers (\eg, \cite{liu2019softras,DIBR19}) compute $\vect{w}$ by solving a linear system.  
For example, if we define $\vect{p}_i= \begin{bmatrix} u & v & 1 \end{bmatrix}^T$ and each vertex $\vect{v} = \begin{bmatrix} x & y & 1 \end{bmatrix}^T$, the triangle $\mat{F}_j$ can be expressed as:
\begin{equation}
    \mat{F}_j = \begin{bmatrix} 
        x_0 & x_1 & x_2 \\
        y_0 & y_1 & y_2 \\
        1 & 1 & 1 
    \end{bmatrix}.
\end{equation}

The barycentric coordinates can then be obtained by solving:
\begin{equation}
    \mat{F}_j \vect{w} = \vect{p}_i  \Rightarrow  \vect{w} = \mat{F}_j^{-1} \vect{p}_i.
    \label{eq:barycentric}
\end{equation}

If all $w_0, w_1, w_2$ fall within the range $[0, 1]$, the pixel $\vect{p}_i$ is covered by the triangle $\mat{F}_j$. Its final color is then determined using the Z-buffer algorithm, which selects the closest surface among all overlapping triangles.  

\vspace{-13pt}

\paragraph{Discussion} We observe that barycentric coordinate computation constitutes a significant computational bottleneck in the context of differentiable motion-blur rasterization.
Since barycentric weights for every pixel must be computed with respect to relevant triangles, the cost scales linearly with the number of temporal samples required to generate smooth, realistic blur effects.
To overcome this limitation, we propose a fast barycentric coordinate solver that drastically reduces computational complexity and significantly accelerates rendering speed.

\subsection{Fast Barycentric Coordinate Solver}
\label{sec:solver}

Traditional rasterization methods~\cite{DIBR19,liu2019softras} synthesize motion blur by rendering multiple $K$ frames and averaging them.
However, by assuming that each triangle moves \emph{linearly} in time, we propose a fast barycentric coordinate solver that avoids the heavy cost of repeated $K$ barycentric computation.  

Consider a 3D mesh object $M$ moving linearly from time $T = 0$ to $T = 1$, where our goal is to render $K$ frames at time steps $T = \frac{0}{K-1}, \frac{1}{K-1}, \frac{2}{K-1}, \dots, \frac{K-1}{K-1}$.
For a triangle $\mat{F}_j$ moving from $T = 0$ to $T = 1$, at a specific time $T=t$, its time-dependent vertex positions can be defined as:
\begin{equation}
    \mat{F}_j(t) = \begin{bmatrix} \vect{v}_0(t) & \vect{v}_1(t) & \vect{v}_2(t) \end{bmatrix}.
\end{equation}

Since we assume linear motion, each vertex position follows linear interpolation between the starting point $\vect{v}(0)$ and ending point $\vect{v}(1)$:
\begin{equation}
    \vect{v}(t) = (1 - t)\vect{v}(0) + t \vect{v}(1).
\end{equation}

Thus, the matrix representation of $\mat{F}_j(t)$ can be represented as:
\begin{equation}
    \mat{F}_j(t) = \begin{bmatrix} 
        x_0(t) & x_1(t) & x_2(t) \\ 
        y_0(t) & y_1(t) & y_2(t) \\ 
        1 & 1 & 1 
    \end{bmatrix}.
\end{equation}

We then compute the barycentric weights $\vect{w}(t)$ as:
\begin{equation}
	\vect{w}(t) = \mat{F}_j(t)^{-1} \vect{p}_i =\frac{\operatorname{adj} (\mat{F}_j(t))}{\det(\mat{F}_j(t))} ~\vect{p}_i,
	\label{eq:time-dependent1}
\end{equation}
where $\operatorname{adj} (\mat{F}_j(t))$ and $\det(\mat{F}_j(t))$ are the adjugate matrix and determinant of $\mat{F}_j(t)$. 

Moreover, with the assumption of linear motion, they could be written as quadratic functions of $t$:
\begin{equation}
	\vect{w}(t) = \frac{\vect{A}_1 t^2 + \vect{A}_2 t + \vect{A}_3}{a_1 t^2 + a_2 t + a_3},
	\label{eq:time-dependent2}
\end{equation}
where $\vect{A}_1, \vect{A}_2, \vect{A}_3, a_1, a_2, a_3$ are precomputed $3 \times 1$ vectors and values that are independent of $t$ and depend solely on $\mat{F}_j(0),\mat{F}_j(1)$ and $\vect{p}_i$. Consequently, for the total $K$ frames, these coefficients ($\vect{A}_1, \vect{A}_2, \vect{A}_3, a_1, a_2, a_3$) can be computed \emph{only once}, and the barycentric coordinate $\vect{w}(t)$ can then be evaluated using \cref{eq:time-dependent2}.
This allows for efficient barycentric computation without per-frame solving barycentric linear equations. The full derivation is provided in \Cref{sec:supp:derivation}.

\subsection{Differentiable Rasterization}
\label{sec:render}

With our fast solver, we now describe our differentiable motion-blur rasterization, which is built on prior state-of-the-art differentiable rasterization works SoftRas~\cite{liu2019softras} and DIB-R~\cite{DIBR19}. 

We first decompose the entire motion into several segments, assuming that inside each segment, all faces move linearly, which is a common assumption in previous motion-blur simulation work~\cite{gribel2010analytical,shkurko2017time,ronnow2021fast}.
Note that our method can support complex motions, \eg, a motion composed of rotation and translation, as long as it can be divided into linear motion segments (see \cref{sec:supp:parabolic-recovery}).
For linear motions, such as translation, larger segments can be used, whereas for non-linear motions, like rotation, smaller segments are employed. In our experiments, we find that even for the extreme rotational motion, we can divide the full rotation into 12 segments and render smooth results.

\vspace*{1pt}
Next, for each segment, we treat its start and end as keyframes and render intermediate frames with our fast solver. These rendered frames are averaged to generate the segment blurry image. Subsequently, all segment images are further averaged to produce the final blurry image.

\vspace*{1pt}
Following DIB-R~\cite{DIBR19}, we also separately process \emph{foreground} pixels (covered by one or more faces) and \emph{background} pixels (not covered by any faces) attributes.

\vspace{-13pt}

\paragraph{Foreground Pixels} For foreground pixels, we perform barycentric interpolation for each frame at time $T=t$ on the closest covering face using the Z-buffer:
\begin{equation}
	I(t) = w_0(t) c_0 + w_1(t) c_1 + w_2(t) c_2,
	\label{eq:fore}
\end{equation}
where $c$ represents vertex attributes (\eg, vertex colors or texture UV coordinates).

\vspace{-13pt}

\paragraph{Background Pixels} If a pixel is not covered by any triangle, in differentiable rasterization it is assumed that it could be influenced by all triangles. Similarly, we extend the probability of a triangle $\mat{F}_j$ influencing a pixel $\vect{p}_i$ to a time-dependent version:
\begin{equation}
    A_i^j(t) = \exp \left( - \frac{d \left(\vect{p}_i, \mat{F}_j(t) \right)}{\delta} \right),
    \label{eq:probability-function}
\end{equation}
where $A_i^j(t)$ is the time-dependent probability, $\delta$ is a hyperparameter~\cite{liu2019softras}, and $d \left(\vect{p}_i, \mat{F}_j(t) \right)$ is the squared Euclidean distance, which can be defined as
\begin{equation}
	d \left(\vect{p}_i, \mat{F}_j(t) \right) = \min_{\vect{p} \in \mat{F}_j(t)} \left| \left| \vect{p}_i - \vect{p} \right| \right|_2^2.
\end{equation}

The core of the Euclidean distance calculation lies in finding $\vect{p} \in \mat{F}_j(t)$ that is closest to $ \vect{p}_i$.  
By replacing $\vect{p} $ with another form $\vect{p} = \mat{F}_j(t)\vect{\hat{w}}$,  where we constrain ${\vect{\hat{w}} \in [0, 1]^3}$ to ensure $\vect{p} \in \mat{F}_j(t)$,
finding the closest $\vect{p}$ is equivalent to finding $\vect{\hat{w}^*}$, which can be written as:
\begin{equation}
\vect{\hat{w}^*} = \mathop{\arg\min}_{\vect{\hat{w}} \in [0, 1]^3} \left\| \mat{F}_j(t) \vect{w}(t) - \mat{F}_j(t) \vect{\hat{w}} \right\|_2^2,
    \label{eq:euc-dist-approx-1}
\end{equation}
where we also replace $\vect{p}_i=\mat{F}_j(t) \vect{w}(t)$. 

However, we find that evaluating \Cref{eq:euc-dist-approx-1} requires additional computational resources for computing  $\mat{F}_j(t)$ across frames. 
To further accelerate the computation, we approximate $\mat{F}_j(t)$ with either $\mat{F}_j(0)$ or $\mat{F}_j(1)$, depending on whether $t$ is closer to the start or the end:

\vspace{-20pt}
\begin{equation}
    \scriptstyle
    \boldsymbol{\hat{w}}^* = \mathop{\arg\min}\limits_{\boldsymbol{\hat{w}} \in [0, 1]^3} \left\| \mathbf{F}_j(X) \boldsymbol{w}(t) - \mathbf{F}_j(X) \boldsymbol{\hat{w}} \right\|_2^2, \; 
    X=\scriptstyle \begin{cases} \scriptstyle 0 & \scriptstyle t \leq 0.5 \\ \scriptstyle 1 & \scriptstyle t>0.5 \end{cases}
    \label{eq:euc-dist-approx-2}
\end{equation}
\vspace{-10pt}

\cref{eq:euc-dist-approx-2} requires only the evaluation of $\mat{F}_j^{(0)}, \mat{F}_j^{(1)}$, which significantly reduces computational cost while introducing only minor approximation errors. The full derivation is provided in \cref{sec:supp:para:euc-distance-approx}.
Finally, with the computed $\vect{\hat{w}^*}$, we obtain
\begin{equation}
 	d \left(\vect{p}_i, \mat{F}_j{(t)} \right) = \left\| \mat{F}_j{(t)} \vect{w}(t) - \mat{F}_j{(t)}\vect{\hat{w}^*}) \right\|_2^2.
 	\label{eq:euc-dist-approx-3}
\end{equation}
 
We then combine the probabilistic influence of all triangle faces on a particular pixel as
\begin{equation}
 	A_i(t) = 1 - \prod_{j} \left(1 - A_i^j(t) \right).
 	\label{eq:back}
\end{equation}

\vspace{-20pt}
\paragraph{Gradient Computation}  
\Cref{eq:fore,eq:back} are fully differentiable~\cite{liu2019softras,DIBR19}. Therefore, our method supports backpropagation by propagating gradients through each time-dependent intermediate frame, and ultimately into the keyframes, ensuring efficient optimization in inverse rendering tasks.

\vspace{-3pt}

\section{Analysis}
\label{sec:analysis}

\vspace{-3pt}

In this section, we evaluate the effectiveness of our method through extensive synthetic experiments.
We implemented our method based on SoftRas \cite{liu2019softras} but split the pixels into foreground and background, following DIB-R~\cite{DIBR19}. We provide the implementation details in \Cref{sec:supp:impl-details}. 

We first analyze the ultra-fast motion blur synthesis effect in \cref{sec:ana:blureffects}, including both forward rendering and backward gradients.
Then, in \cref{sec:ana:speed}, we present the computation speed, demonstrating significant acceleration over prior methods.
\vspace{-1pt}

\subsection{Qualitative Validation}
\label{sec:ana:blureffects}
\vspace{-4pt}

We first present the synthesis of ultra-fast motion blur effects, including both translational and rotational movements. We show forward rendering results and their backward gradients.
As a reference, we also apply SoftRas~\cite{liu2019softras} to render with the same settings (\eg, the same number of sampled frames) and compute the corresponding gradients. The default hyperparameter values provided in SoftRas are used for this comparison.

\begin{figure}
    \centering
    \begin{spacing}{1}
    \setlength\tabcolsep{-0.5pt}

    \newcommand{\vlinecell}[1]{\multirow{#1}{*}{\rule{0.5pt}{19.2em}\hspace{2.5pt}}}
    
    \newcolumntype{M}[1]{>{\centering\arraybackslash}m{#1}}
    \begin{tabular}{cccccccc} 
       ~ & ~ & \multicolumn{2}{c}{Translation} & \vlinecell{6} & \multicolumn{3}{c}{Rotation} \\
    & & {\small{10 Samples}} & \multicolumn{1}{c}{{\small{50 Samp.}}} & & {\small{12 Samp.}} & {\small{60 Samp.}} & {\small{240 Samp.}} \\
    
    \multirow{3}{*}{\rotatebox[origin=c]{90}{Forward} \hspace{1pt}} 
    & 
    \rotatebox[origin=c]{90}{SoftRas} \hspace{1pt} &
            \imagecell[0.17]{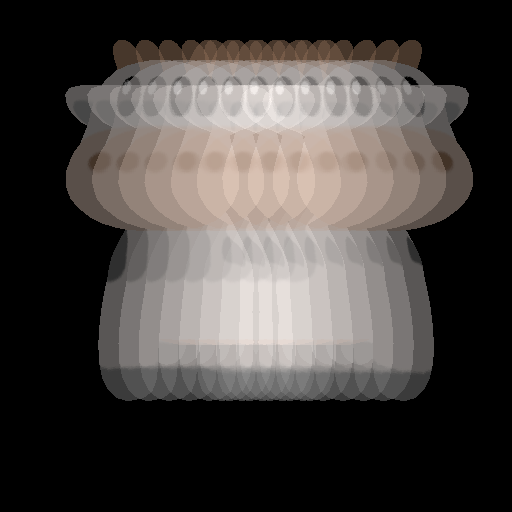} & 
            \imagecell[0.17]{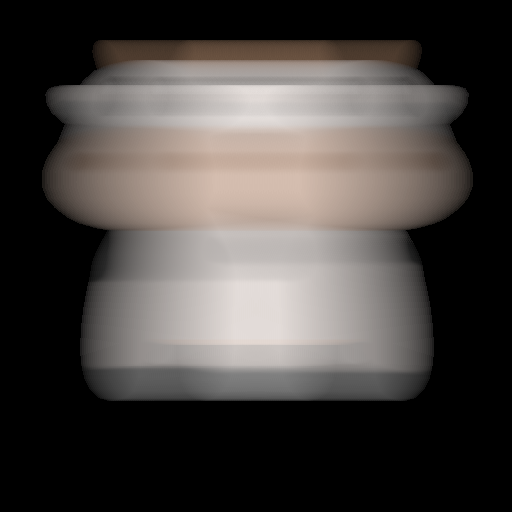} & \vspace{1.5pt}  &
            \imagecell[0.17]{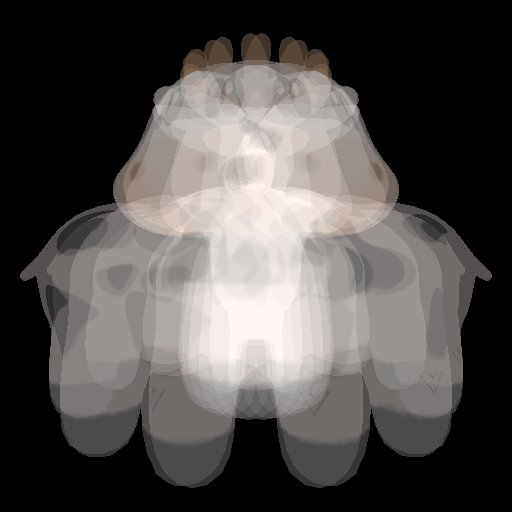} & 
            \imagecell[0.17]{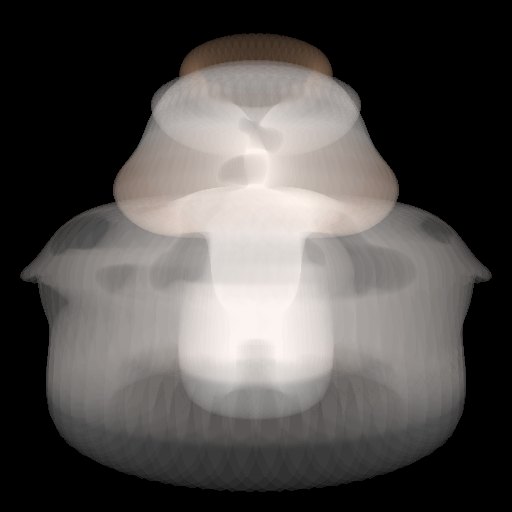} & 
            \imagecell[0.17]{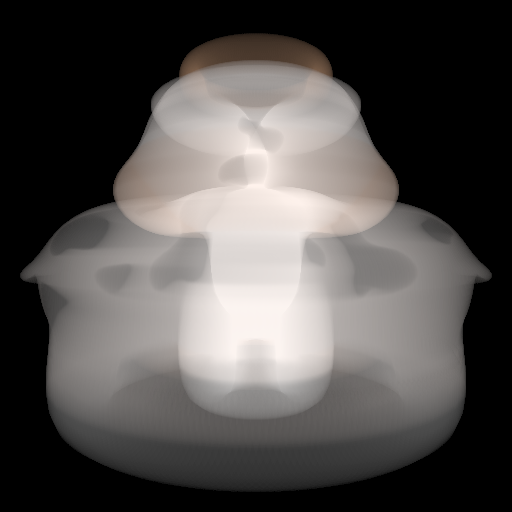}
    \\ 
    &
    \rotatebox[origin=c]{90}{Ours} \hspace{1pt} &
            \imagecell[0.17]{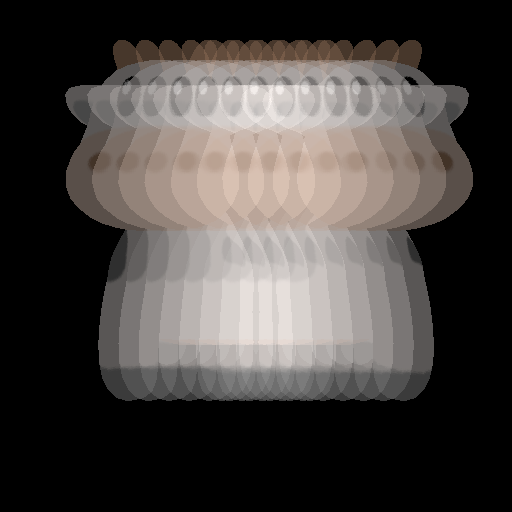} & 
            \imagecell[0.17]{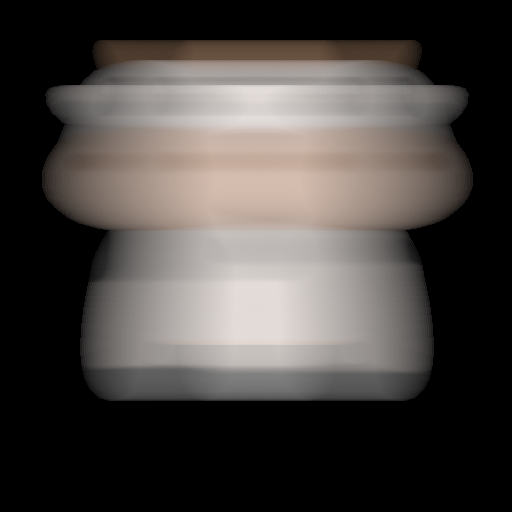} & \vspace{1.5pt} &
            \imagecell[0.17]{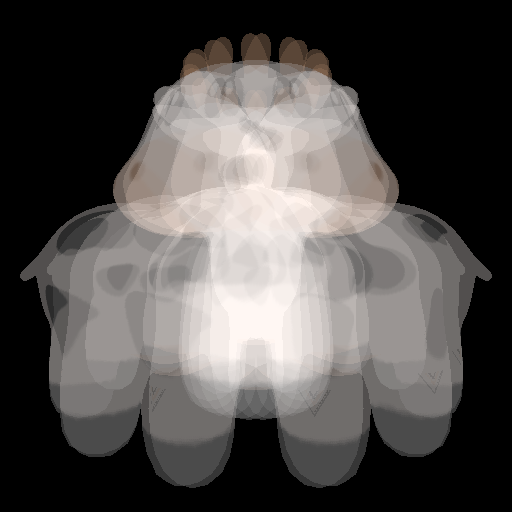} & 
            \imagecell[0.17]{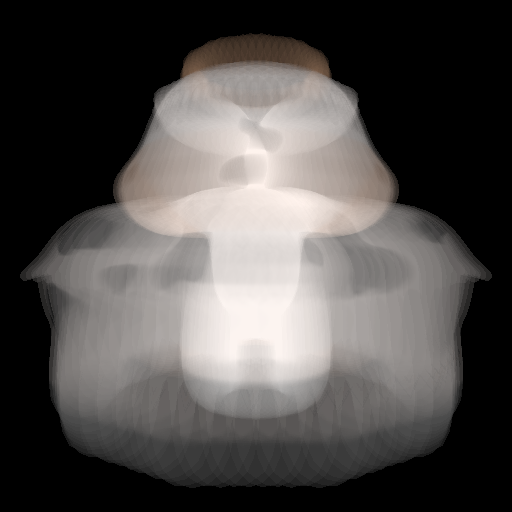} & 
            \imagecell[0.17]{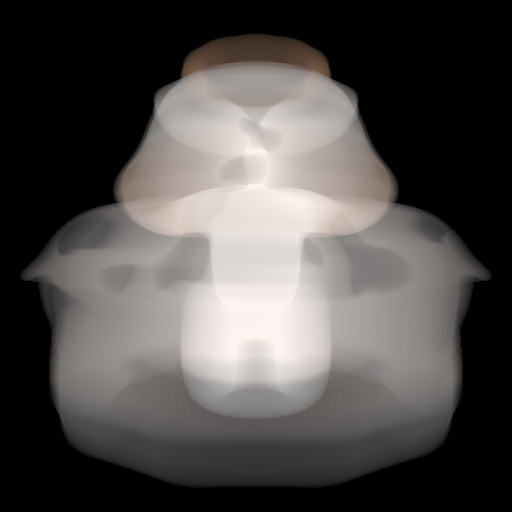}
    \\ 

    \multirow{3}{*}{\rotatebox[origin=c]{90}{Gradient} \hspace{1pt}} 
    &
    \rotatebox[origin=c]{90}{SoftRas} \hspace{1pt} &
            \imagecell[0.17]{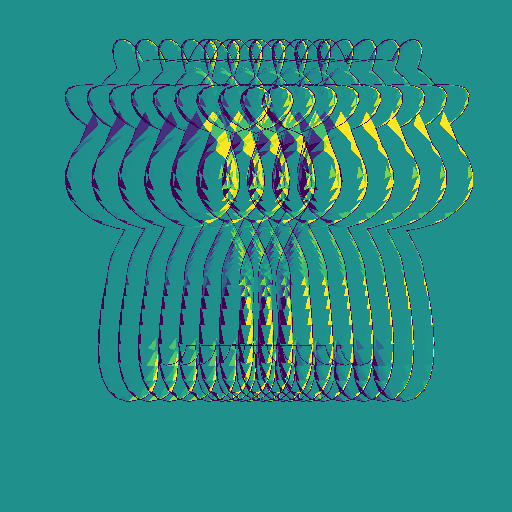} & 
            \imagecell[0.17]{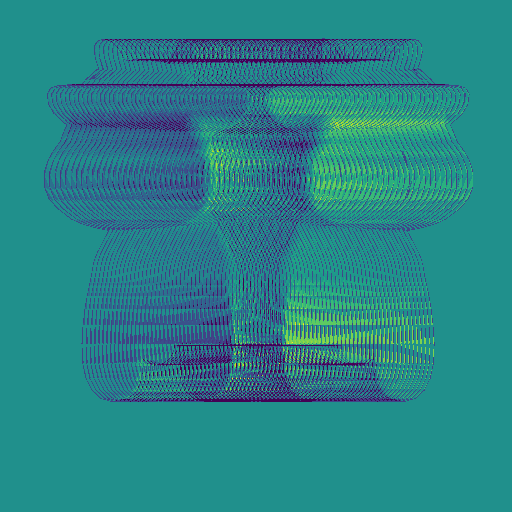} & \vspace{1.5pt} &
            \imagecell[0.17]{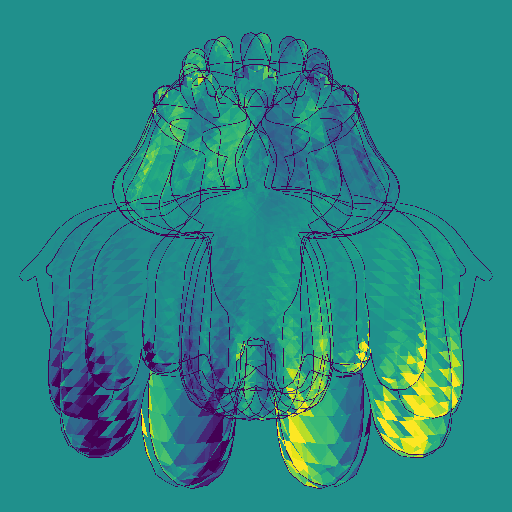} & 
            \imagecell[0.17]{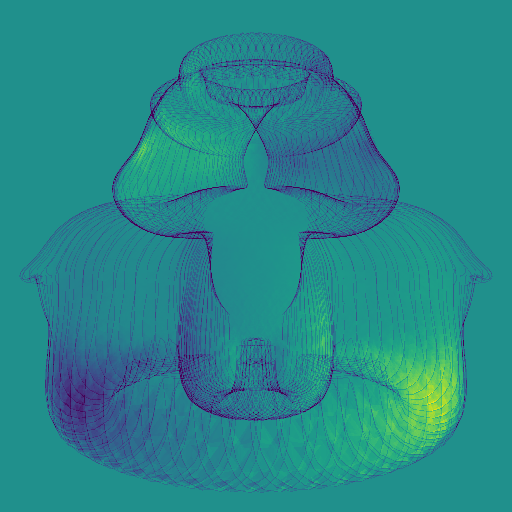} & 
            \imagecell[0.17]{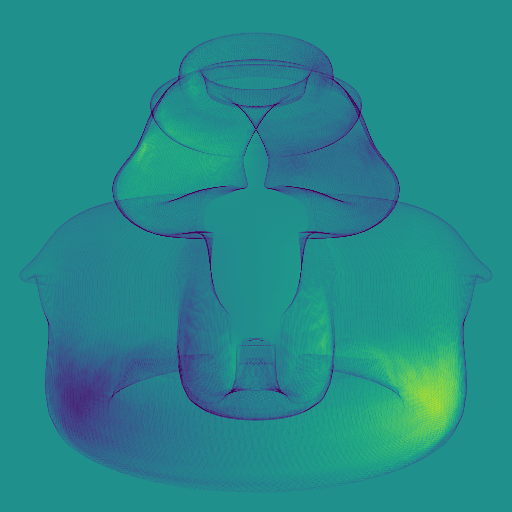}
    \\
    &
    \rotatebox[origin=c]{90}{Ours} \hspace{1pt} &
            \imagecell[0.17]{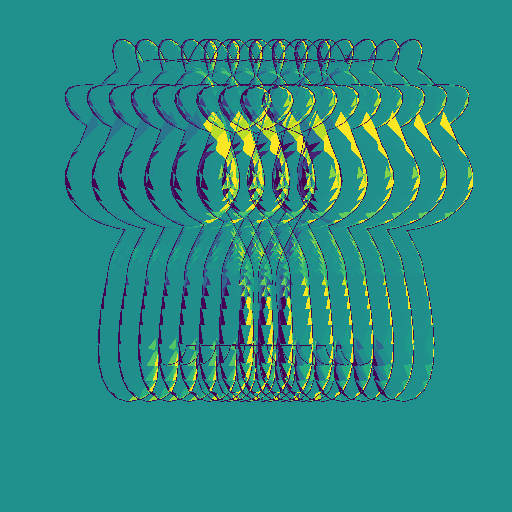} & 
            \imagecell[0.17]{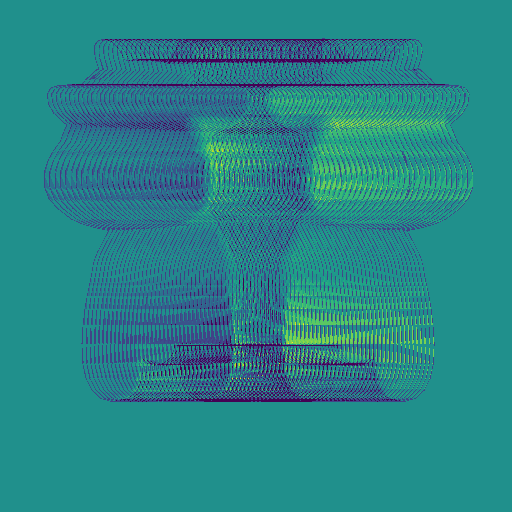} & &
            \imagecell[0.17]{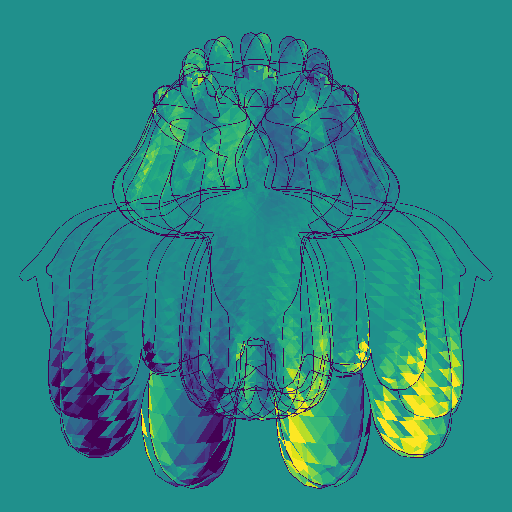} & 
            \imagecell[0.17]{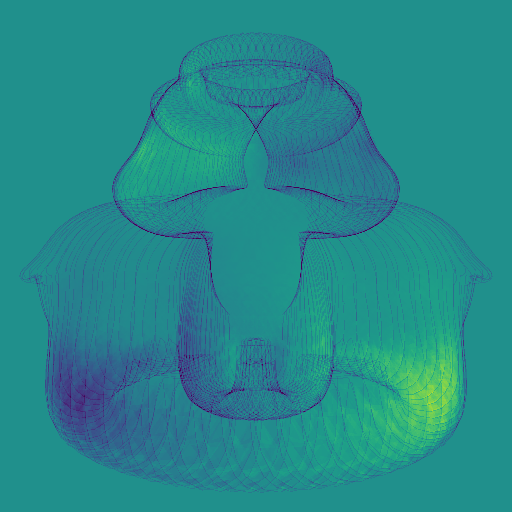} & 
            \imagecell[0.17]{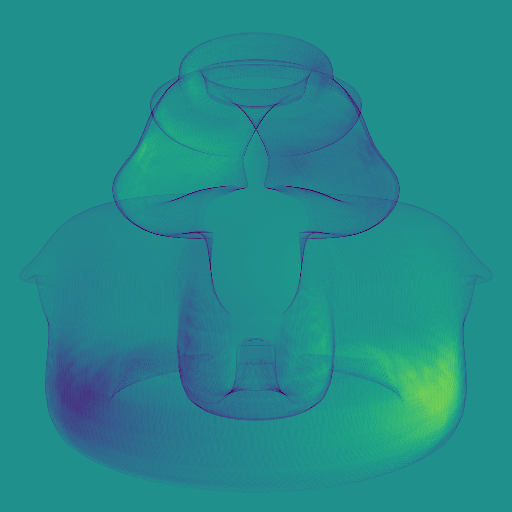}
    \\
    
    \end{tabular}
    
    \end{spacing}
    \vspace{-3pt}
    
    \caption{Forward rendering \& backward gradient visualization for ultra-fast motion-blur synthesis. 
    Our rendered images and gradients exhibit a high degree of similarity to those generated by SoftRas across various motion cases and sample numbers.
    Scene settings and render details are provided in \Cref{sec:supp:visualization-details,sec:supp:scene-settings}.}%
    
    \label{fig:image-analysis}
    \vspace{-20pt}
\end{figure}

The comparison results are shown in \cref{fig:image-analysis}. 
Across all test cases, our rendered images and gradients exhibit a high degree of similarity to those generated by SoftRas, which demonstrates the effectiveness of our method in realistic and differentiable motion blur synthesis.  
In theory, under linear motion, our foreground pixel renderings should be identical to those of SoftRas, whereas our background pixel computation shows slight discrepancies, primarily due to our Euclidean distance approximation (as detailed in \cref{eq:euc-dist-approx-1,eq:euc-dist-approx-2}). 
However, we show that these minor discrepancies have negligible impact on gradient computation of our method (\cref{fig:image-analysis} Bottom).

\vspace{-5pt}

\subsection{Speed Comparison}
\label{sec:ana:speed}

\vspace{-5pt}

Our method is significantly more efficient than traditional blur synthesis methods, \eg, applying SoftRas to render and average multiple frames.
To evaluate the running speed, we randomly select 50 models from ShapeNet \cite{chang2015shapenet}, each containing an average of 5,536 faces. 
We apply random rotations to each model, render $128 \times 128$ front-view motion-blurred silhouette and color images, and measure the time required for both forward rendering and gradient computation in a single pass.  
We render objects undergoing linear translation with a varying number of samples.
The evaluation considers the average time required for forward rendering and backward gradient computation across all models.
All experiments are conducted on a single NVIDIA RTX 4090 GPU with 24GB of memory.
More details are provided in \Cref{sec:supp:scene-settings}.

\begin{figure}
\centering

\includegraphics[width=0.8\linewidth]{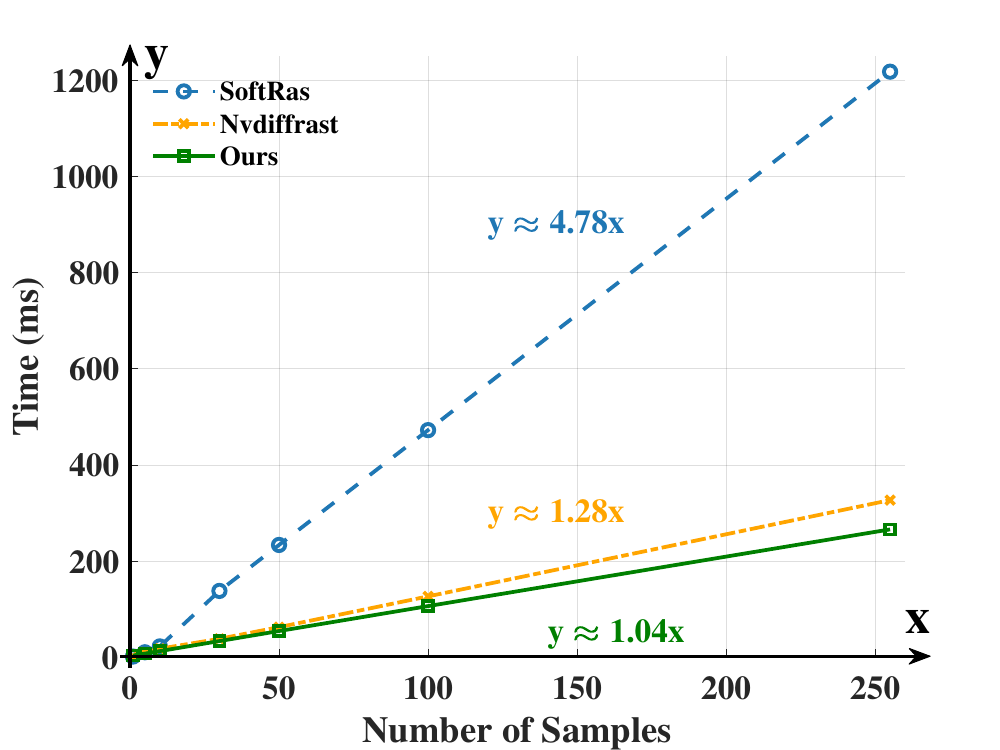}

\vspace{-8pt}

 \caption{Forward + gradient computation timing results. The slope represents the average time of sampling once. The lower the better. 
 Our method achieves speedups of up to 4.57$\times$ and 1.23$\times$ compared to SoftRas and Nvdiffrast, respectively.} 
\label{tab:speed-comparison}
\vspace{-15pt}
\end{figure}

The timing results are summarized in \cref{tab:speed-comparison}. Our method achieves speedups of up to 4.57$\times$ over SoftRas and 1.23$\times$ over Nvdiffrast.
Regarding Nvdiffrast \cite{laine2020modular}, it is a highly performant OpenGL-optimized rasterization library, whereas our implementation is built on the SoftRas implementation. Nevertheless, our method is still faster than Nvdiffrast, and incorporating our method into Nvdiffrast would likely bring further acceleration. 
Moreover, we observe that Nvdiffrast produces weak gradient signals, as its gradients are computed only near edges. This limitation can lead to slower convergence or even failure in extreme shape recovery tasks. In contrast, our method enables global gradient propagation across all triangle primitives, facilitating smoother optimization. Further discussion is provided in \cref{para:failure-cases-for-nvdiffrast}.

\vspace{-7pt}

\section{{Shape Optimization from Blurred Images}}

\label{sec:applications}
\label{sec:shape-optimization-from-blurred-images}

With our differentiable rendering pipeline, we now present inverse rendering applications, \ie, recovering 3D shapes from ultra-fast motion-blurred images. 
Thanks to our efficient framework, our method supports rendering more samples, resulting in smoother forward rendering effects and better backward gradients across the optimization process.
We demonstrate the effectiveness and advantages of our method through two challenging tasks, where the goals are to recover the 3D shape of ultra-fast moving objects from two representative types of motion blur: multi-view translational and rotational blurred images.

Similar to other inverse rendering tasks, we assume known rendering parameters for each image, including its camera viewpoint and blur settings (translation or rotation speed).

\vspace{-5pt}

\subsection{Translational Recovery}
\label{sub-sub-sec:optim-mesh-geo-col}

\vspace{-5pt}

\begin{figure}
    \centering
    \setlength\tabcolsep{0pt}
    \begin{tabular}{cccc}
    Input & \cite{rozumnyi2021shape} & Ours & G.T. \\
        \imagecell[0.23]{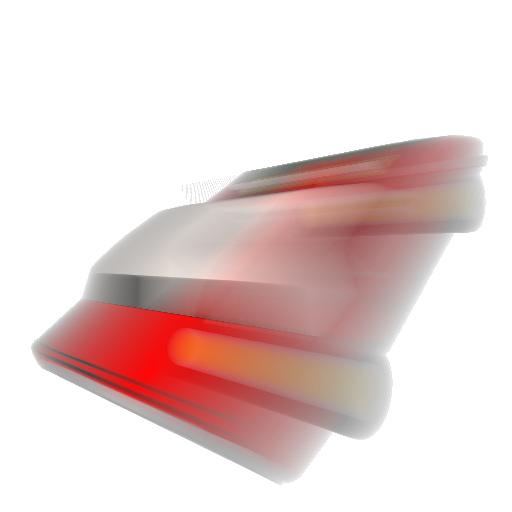} & 
        \imagecell[0.23]{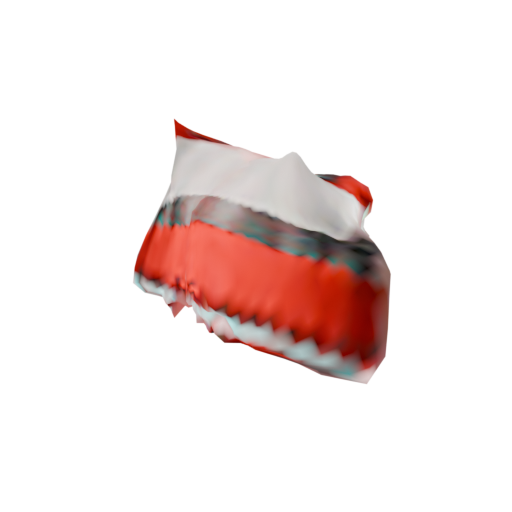} &
        \imagecell[0.23]{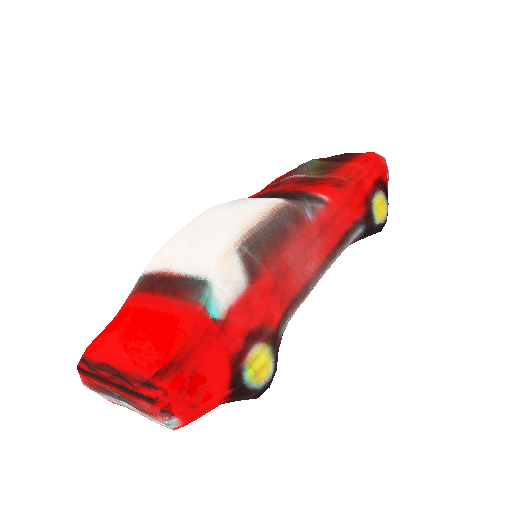} & 
        \imagecell[0.23]{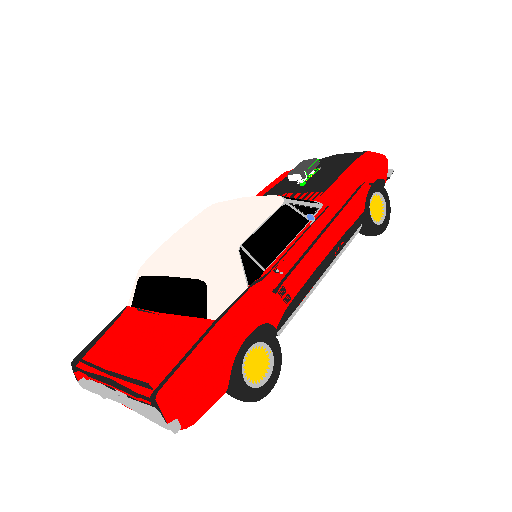} 
    \\
        \imagecell[0.23]{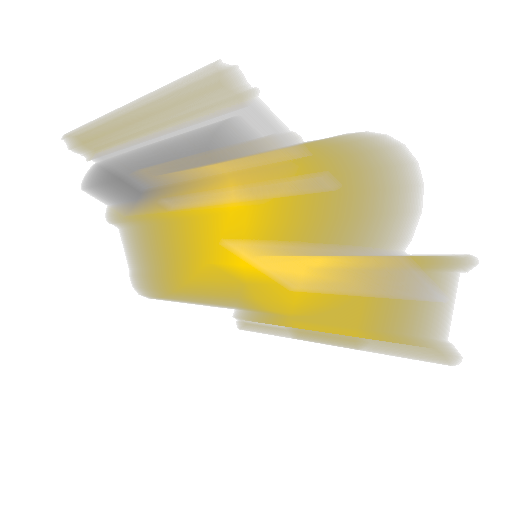} & 
        \imagecell[0.23]{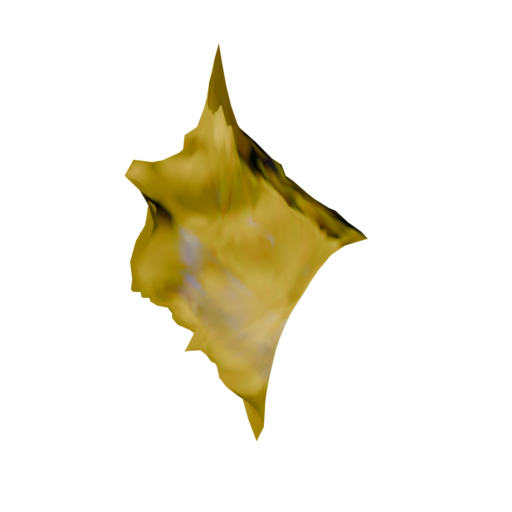} &
        \imagecell[0.23]{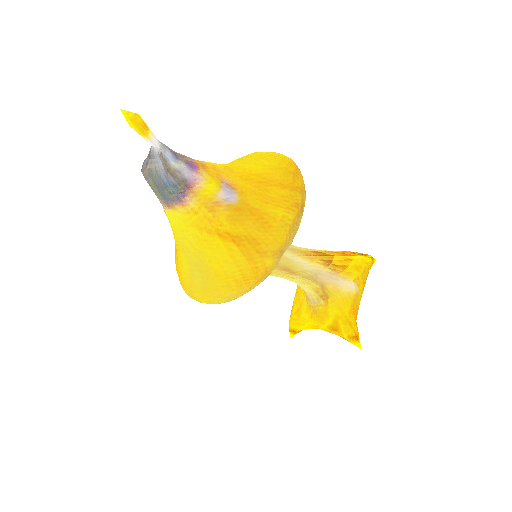} &
        \imagecell[0.23]{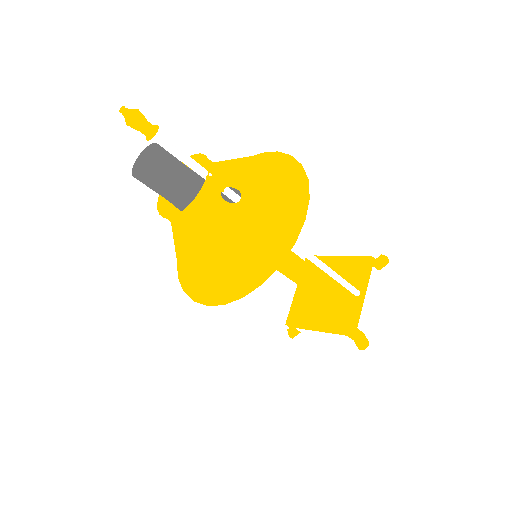} 
    \\
    %
    %
        \imagecell[0.23]{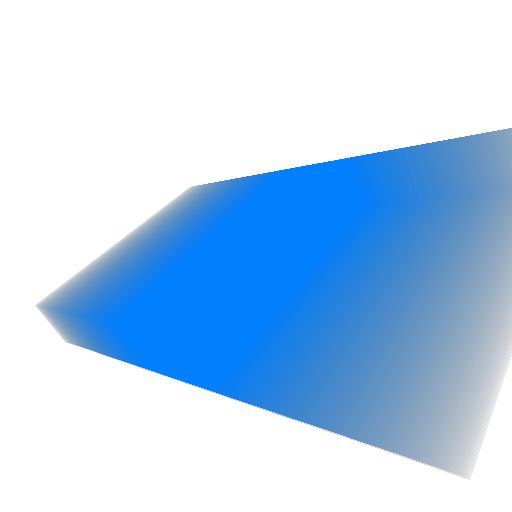} & 
        \imagecell[0.23]{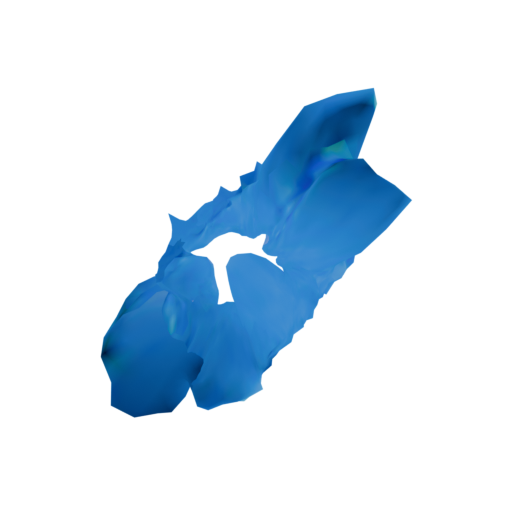} &
        \imagecell[0.23]{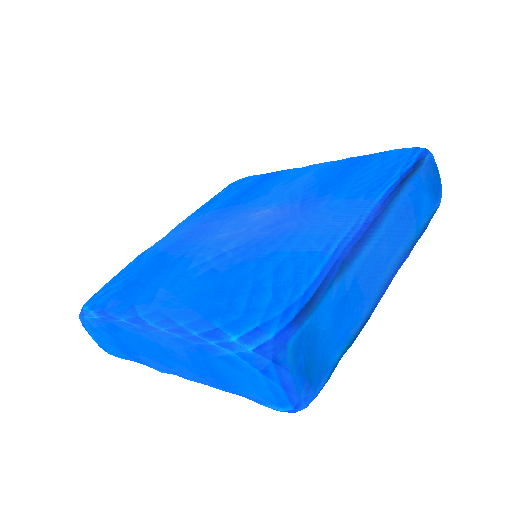} & 
        \imagecell[0.23]{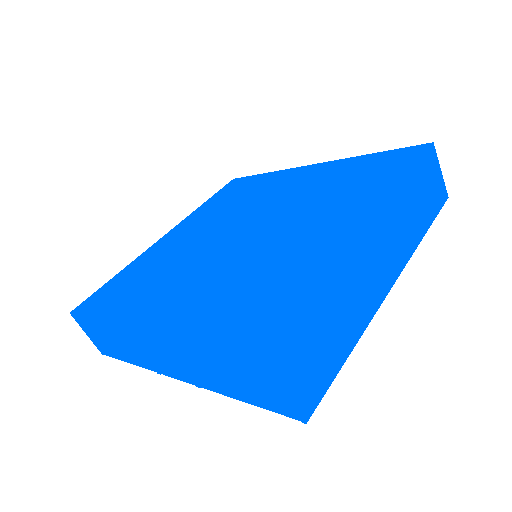} 
    \\
        \imagecell[0.23]{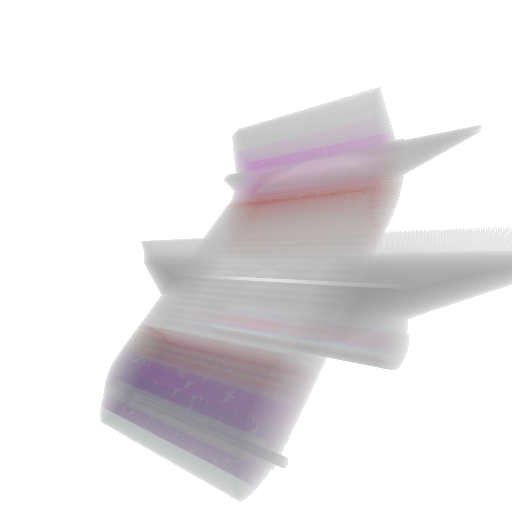} & 
        \imagecell[0.23]{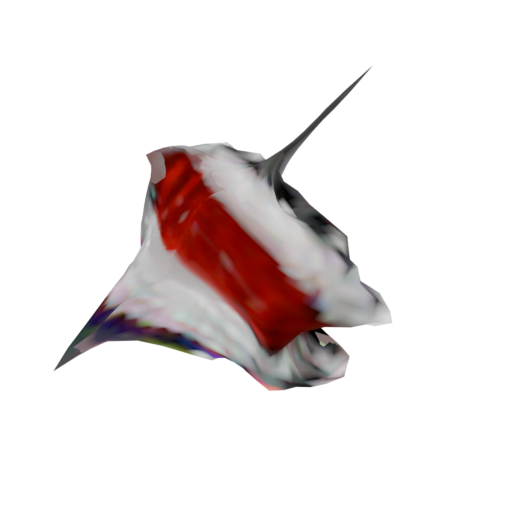} &
        \imagecell[0.23]{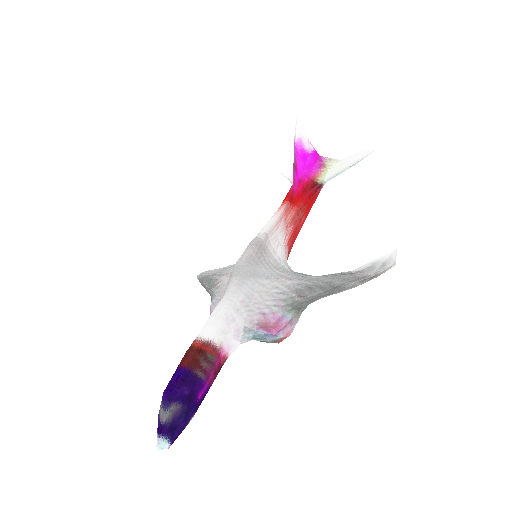} &
        \imagecell[0.23]{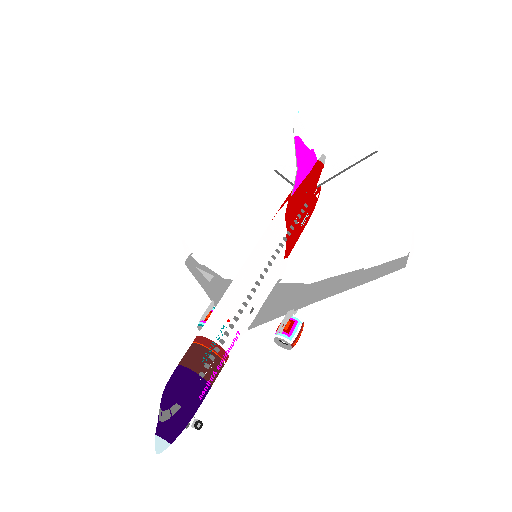}
    \\
    
    (a) & (b) & (c) & (d)
    \end{tabular}
    \caption{Qualitative results on geometry and color optimization. (a) One of blurred input images. (b) State of the Art \cite{rozumnyi2021shape} result. (c) Our optimization result. (d) Ground-truth object. Our method yields significant superior results than the state-of-the-art work. Note that the object trajectories are synthesized to test the solver's robustness to diverse motion vectors, rather than simulating realistic physical dynamics.}
    \label{fig:qualitative-optim-translation}
    \vspace{-15pt}
\end{figure}

We begin with the task of optimizing a 3D shape undergoing linear translation. 
Given multi-view RGBA translational blurred images (consisting of RGB color $I$ and transparency $\alpha$) as input, 
our method optimizes a mesh $M$ with vertex positions $V$ and a texture map $C$ such that the rendered images, $\hat{I}, \hat{\alpha} = R(V, C)$, match the input. 

The optimization of $V$ and $C$ is performed by minimizing the image loss and regularization terms. 
We adopt the $\mathcal{L}_{1}$ loss for both color $I, \hat{I}$ and transparency $\alpha, \hat{\alpha}$, formulated as:
\begin{equation}
    \mathcal{L}_{\text{img}} = \left\| I - \hat{I} \right\|_1 + \left\| \alpha - \hat{\alpha} \right\|_1.
\end{equation}

Similar to \cite{liu2019softras,DIBR19}, we incorporate a smoothness loss $\mathcal{L}_{\text{s}}$ and a Laplacian loss $\mathcal{L}_{\text{L}}$ to regularize the deformation of $V$ (details provided in \Cref{sec:supp:loss-terms}). The final loss function is defined as:
\begin{equation}
    \mathcal{L} = \mathcal{L}_{\text{img}} + \lambda_{\text{s}} \mathcal{L}_{\text{s}} + \lambda_{\text{L}} \mathcal{L}_{\text{L}}. 
    \label{eq:loss-img}
\end{equation}

\vspace{-5pt}

\subsection{Rotational Recovery}

\vspace{-5pt}

\label{sub-sub-sec:optim-rotation}

We further apply our method to a more challenging task: reconstructing fast-rotating objects from motion-blurred images. 
Similarly, given multi-view images $I$ as input, our method reconstructs a 3D mesh $M$ corresponding to the observed object.

Due to the highly non-convex nature of the rotation optimization problem, 
we observe that directly optimizing mesh vertices rarely yields well-shaped objects (illustrated in \cref{sec:supp:failure-mesh-rotational-optim}).
To mitigate this issue, we optimize a Signed Distance Function (SDF) representation instead. 
We construct an SDF field $\mathcal{S}$ following \cite{wang2023neural}, and extract a mesh $M$ using FlexiCubes \cite{shen2023flexicubes} in a differentiable manner.
Our differentiable renderer $R$ is then employed to generate corresponding rotation images $\hat{I}$, which are subsequently used to optimize $\mathcal{S}$ via loss functions. %
Given the complexity of the SDF representation, we render grayscale images in this setting and focus on shape recovery.

We retain the same $\mathcal{L}_{\text{img}}$ loss (from \cref{eq:loss-img}) for image consistency. For SDF regularization, we utilize the loss terms $\mathcal{L}_{\text{crit}}$ and $\mathcal{L}_{\text{reg}}$ from \cite{wang2023neural} and \cite{shen2023flexicubes}, respectively (details in \Cref{sec:supp:loss-terms}). The final loss function is defined as:
\begin{equation}
    \mathcal{L} = \mathcal{L}_{\text{img}} + \lambda_{\text{crit}} \mathcal{L}_{\text{crit}} + \lambda_{\text{reg}} \mathcal{L}_{\text{reg}}. 
\end{equation}

\section{Experiments}
\label{sec:experiments}

In this section, we present qualitative and quantitative experiments to evaluate the effectiveness and efficiency of our method for shape recovery from ultra-fast motion-blurred images. 
In \cref{sub-sec:exp-translation,sub-sec:exp-rotation}, we present translational and rotational blurred shape recovery, respectively.
Furthermore, we validate our method's practical applicability through real-world motion blur data in \cref{sub-sec:exp-real-world}.
Finally, we conduct an ablation study in \cref{sub-sec:exp-ablation}, to demonstrate the benefits of our method compared to the widely adopted codebase SoftRas \cite{liu2019softras} and Nvdiffrast \cite{laine2020modular} under extreme motion blur conditions. 
All hyperparameter settings and more results are provided in \Cref{sec:supp:scene-settings,sec:supp:hyperparameter,sec:supp:more-results}.

\begin{table}
\centering
\setlength\tabcolsep{3pt}
\begin{tabular}{ccccc}
\toprule
\multirow{2}{*}{\textbf{Method}} & \multicolumn{2}{c}{\textbf{Translation}} & \textbf{Rotation} \\
& 3D IoU $\uparrow$ & Static PSNR $\uparrow$ & Blurred PSNR $\uparrow$ \\
\midrule
\text{\cite{rozumnyi2021shape}} & 0.152 & 11.63 & 13.58 \\
Ours & 0.679 & 19.20 & 31.89 \\
\bottomrule
\end{tabular}
\vspace{-7pt}
\caption{
Quantitative comparison for shape optimization from blurred images. We compare the best performance between our method and the state-of-the-art \cite{rozumnyi2021shape}. Our method achieves significantly superior performance compared to \cite{rozumnyi2021shape}. 
}
\label{tab:comparison-sfb}
\vspace{-15pt}
\end{table}

\vspace{-5pt}

\subsection{Translational Recovery}
\label{sub-sec:exp-translation}

\vspace{-5pt}

In this experiment, we present our method's 3D shape reconstruction performance for objects undergoing translational motion, specifically benchmarking against the state-of-the-art \cite{rozumnyi2021shape}.
We perform optimization on 25 selected shapes from ShapeNet, and evaluate both the geometry and color recovery. For geometry quantitative evaluation, we voxelize the predicted and ground truth meshes into $32^3$ volumes, and compute the 3D IoU. For color quantitative evaluation, we compare the PSNR of multi-view static novel-view-synthesis (NVS) of the objects.
Quantitative evaluation results are presented in \cref{tab:comparison-sfb}, while qualitative results are illustrated in \cref{fig:qualitative-optim-translation}.

\label{sub:rozumnyi-comparison-details}
We assess our method's fundamental capability in 3D shape recovery from highly motion-blurred images by benchmarking against the state-of-the-art work \cite{rozumnyi2021shape}.
As presented in \cref{tab:comparison-sfb,fig:qualitative-optim-translation}, our method demonstrates a significant advantage in translational recovery.
Notably, \cite{rozumnyi2021shape}'s reliance on the learning prior \cite{rozumnyi2021defmo} to predict static silhouettes fundamentally limits its performance in the challenging blurry input. In contrast, our method successfully recovers meaningful 3D shape and appearance.
More details and analysis are provided in \Cref{sec:supp:sfb-analysis}. 

\vspace{-5pt}

\subsection{Rotational Recovery}
\label{sub-sec:exp-rotation}

\vspace{-5pt}

\newcommand{\imagecellspec}[1]{%
    \begin{minipage}[b]{0.23\columnwidth} 
        \centering
        \IfFileExists{#1.png}{%
            \raisebox{-.5\height}{\includegraphics[width=\linewidth, trim={4cm 4cm 4cm 4cm}, clip]{#1.png}}%
        }{%
            \raisebox{-.5\height}{\includegraphics[width=\linewidth]{example-image}}%
        }
    \end{minipage} 
}

\newcommand{\imagecellspecin}[1]{%
    \begin{minipage}[b]{0.23\columnwidth} 
        \centering
        \IfFileExists{#1.png}{%
            \raisebox{-.5\height}{\includegraphics[width=\linewidth, trim={1cm 1cm 1cm 1cm}, clip]{#1.png}}%
        }{%
            \raisebox{-.5\height}{\includegraphics[width=\linewidth]{example-image}}%
        }
    \end{minipage} 
}

\newcommand{\imagecellspecgt}[1]{%
    \begin{minipage}[b]{0.23\columnwidth} 
        \centering
        \IfFileExists{#1.png}{%
            \raisebox{-.5\height}{\includegraphics[width=\linewidth, trim={2cm 2cm 2cm 2cm}, clip]{#1.png}}%
        }{%
            \raisebox{-.5\height}{\includegraphics[width=\linewidth]{example-image}}%
        }
    \end{minipage} 
}

\begin{figure}
    \centering
    \setlength\tabcolsep{0pt}
    \begin{tabular}{cccc}
    Input & \cite{rozumnyi2021shape} & Ours & G.T. \\
        \imagecellspecin{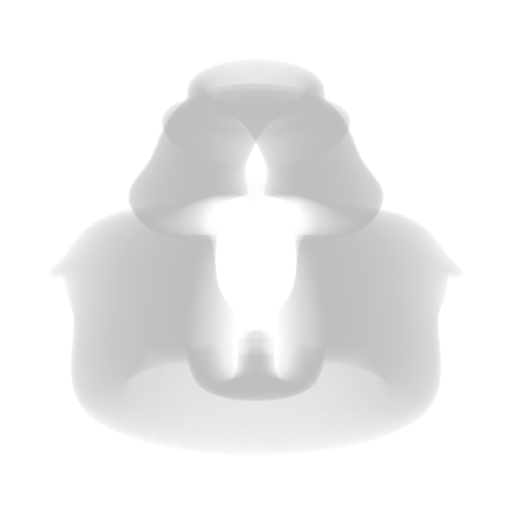} & 
        \imagecell[0.23]{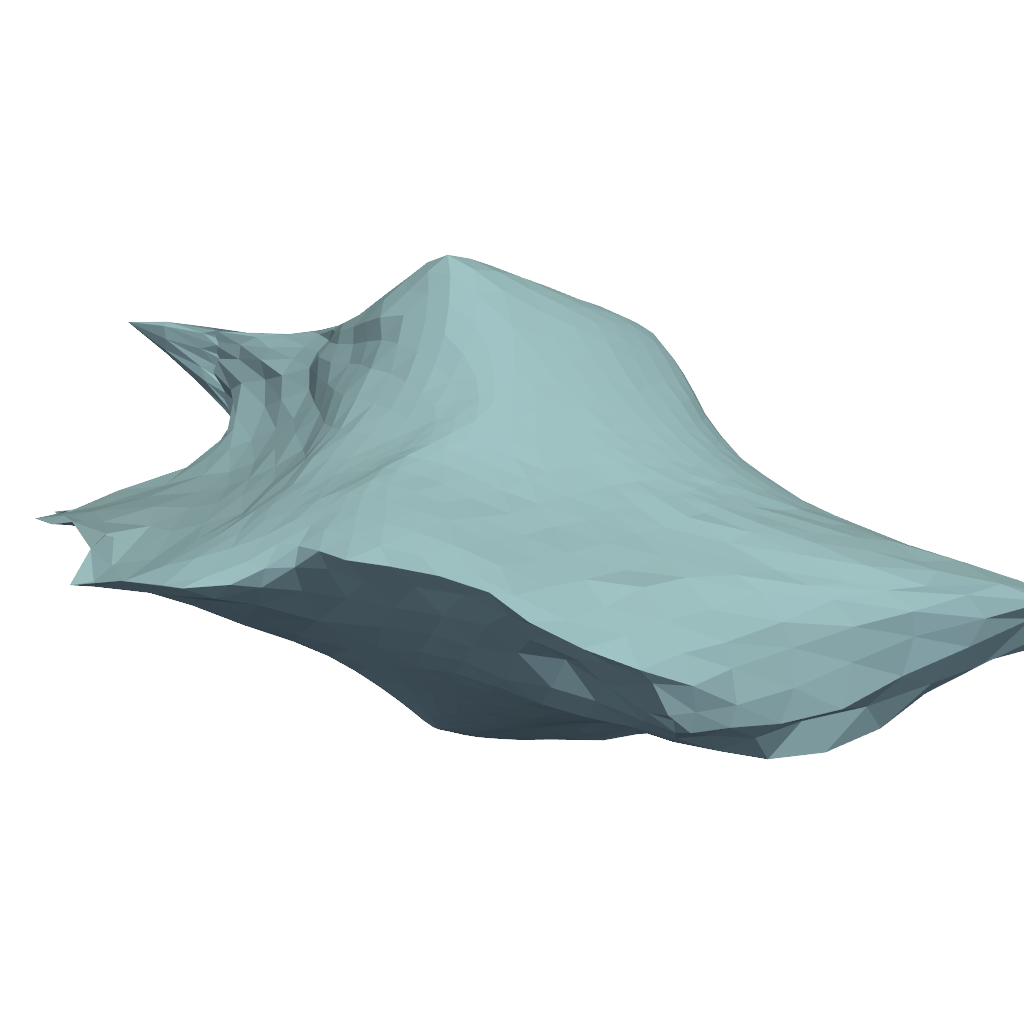} &
        \imagecell[0.23]{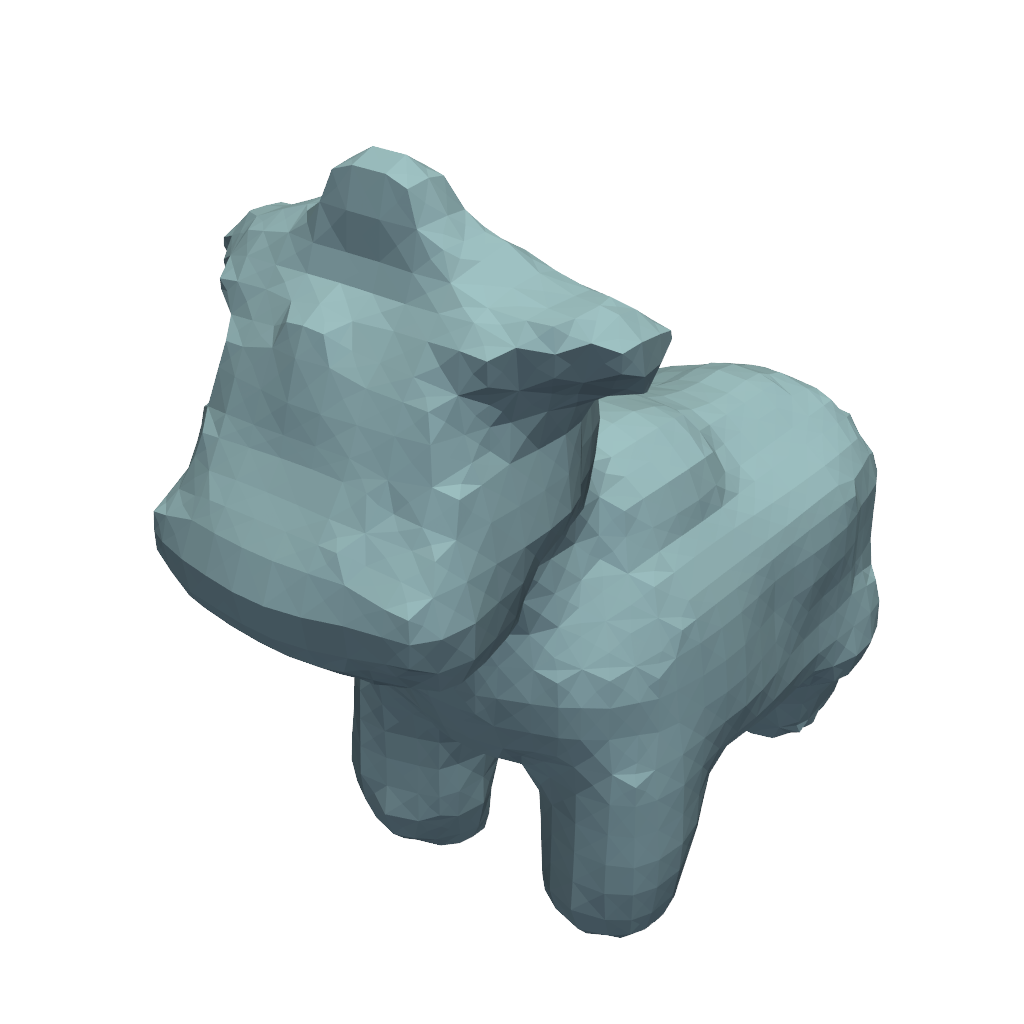} & 
        \imagecell[0.23]{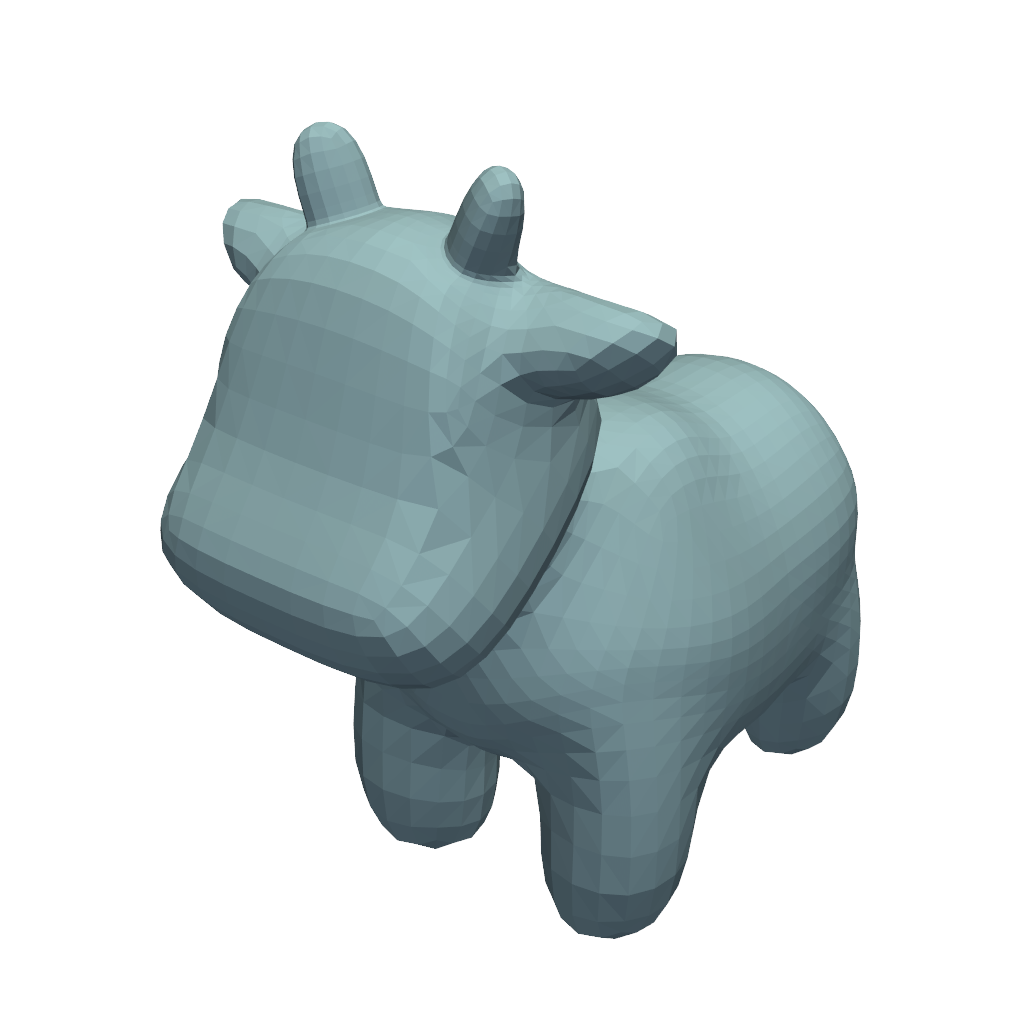}
        \\
        \imagecellspecin{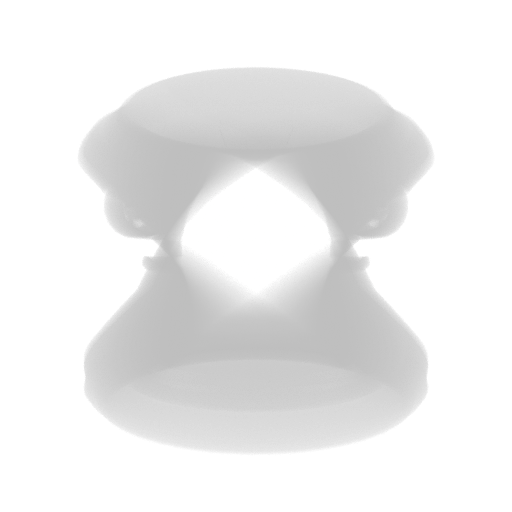} &  
        \imagecell[0.23]{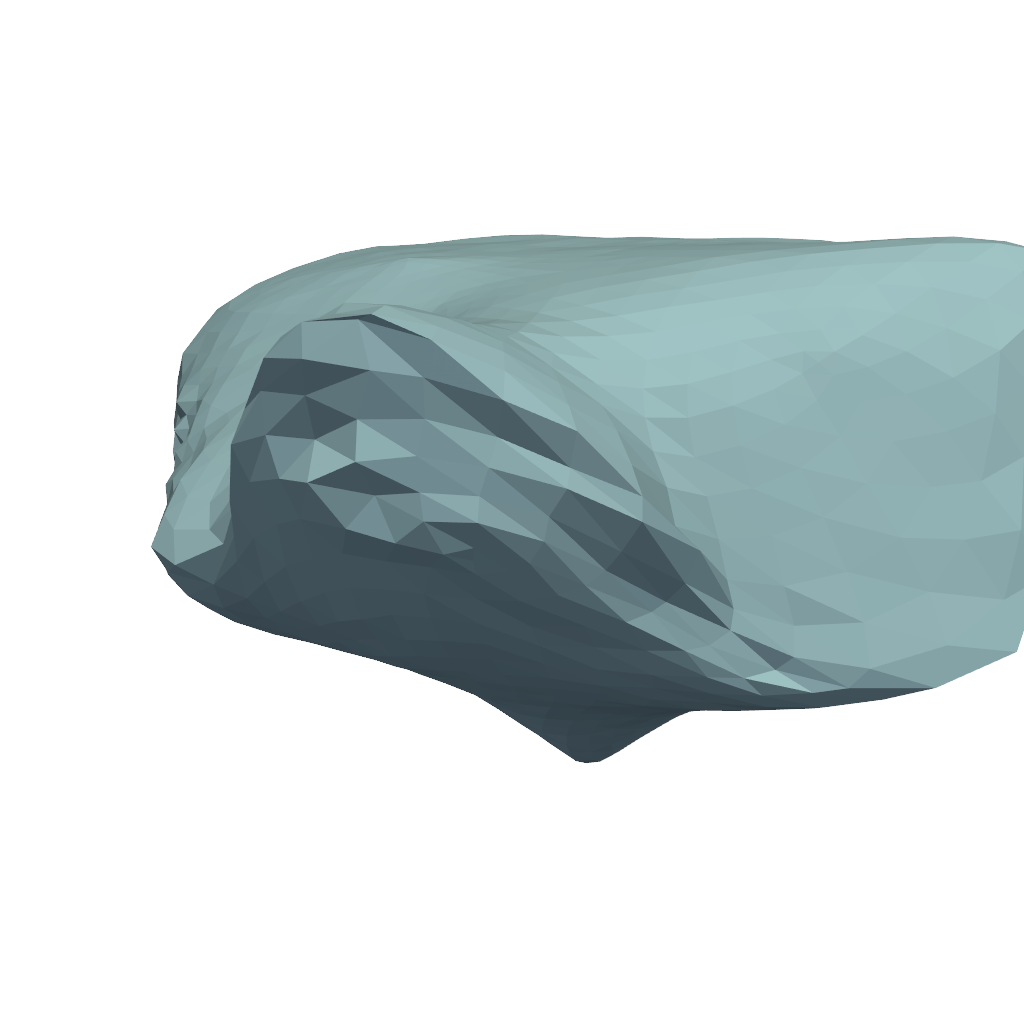} & 
        \imagecell[0.23]{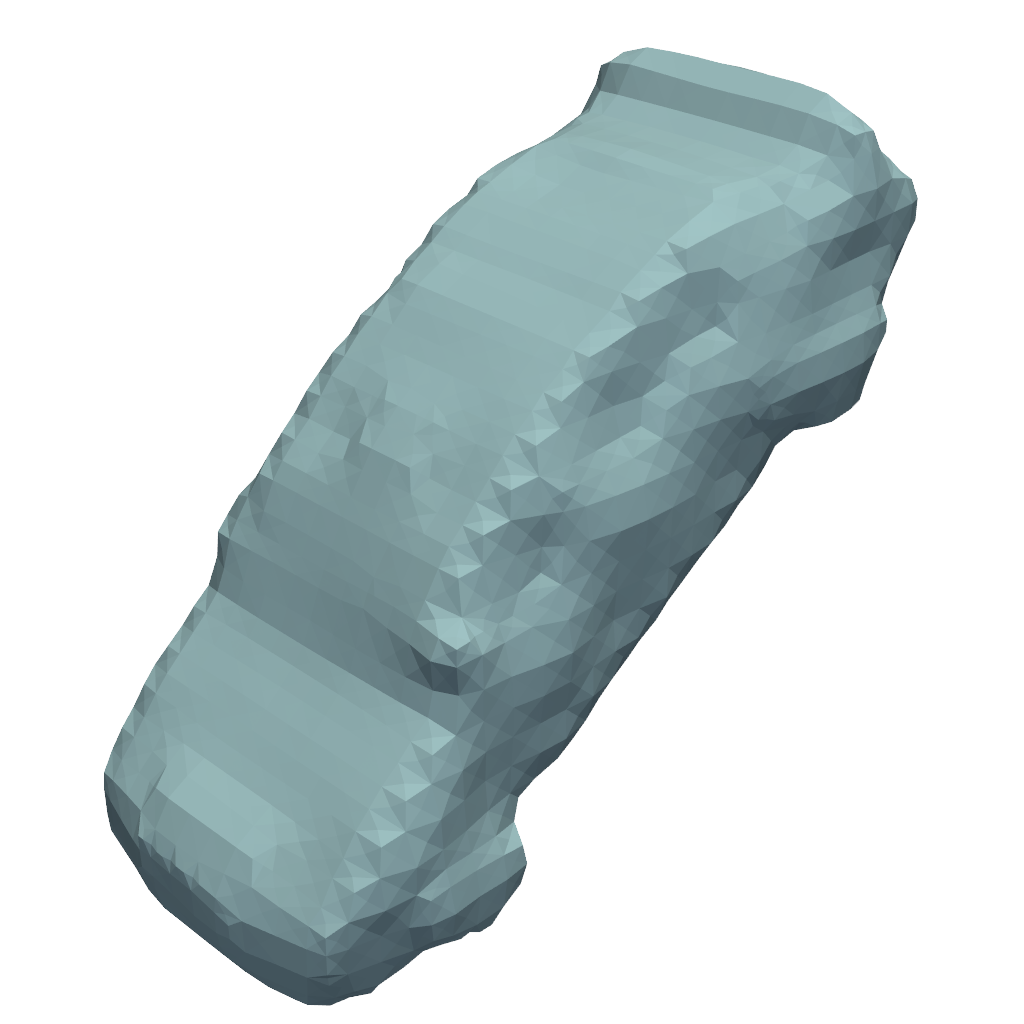} & 
        \imagecell[0.23]{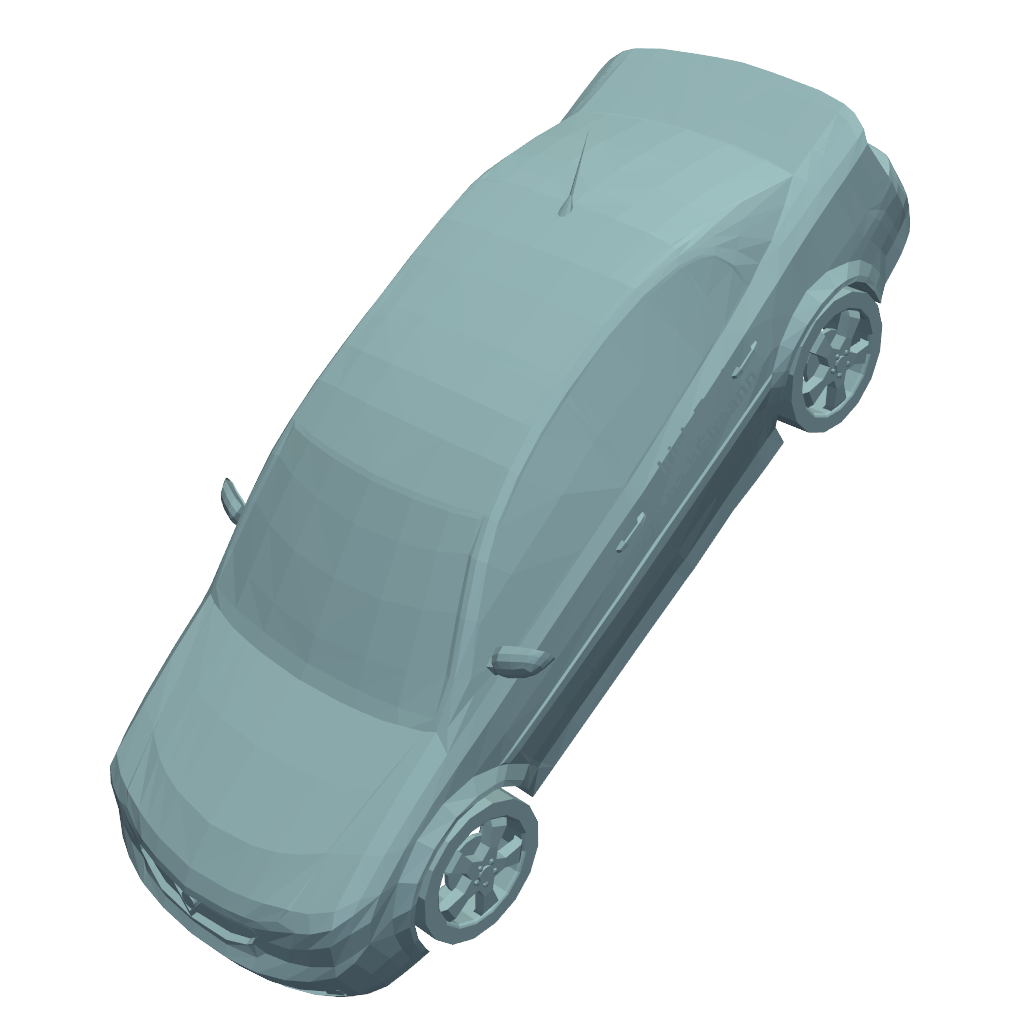}  
        \\
        \imagecellspecin{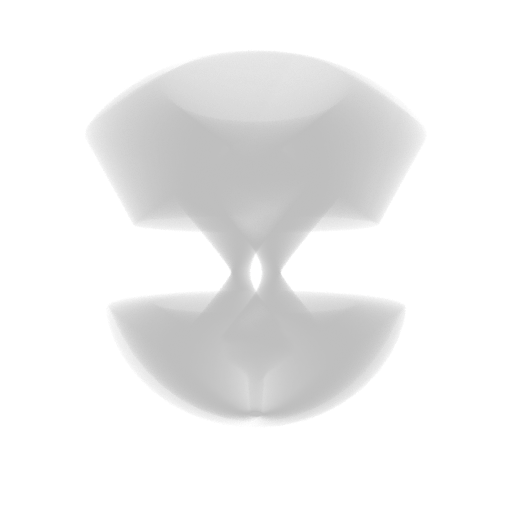} & 
        \imagecell[0.23]{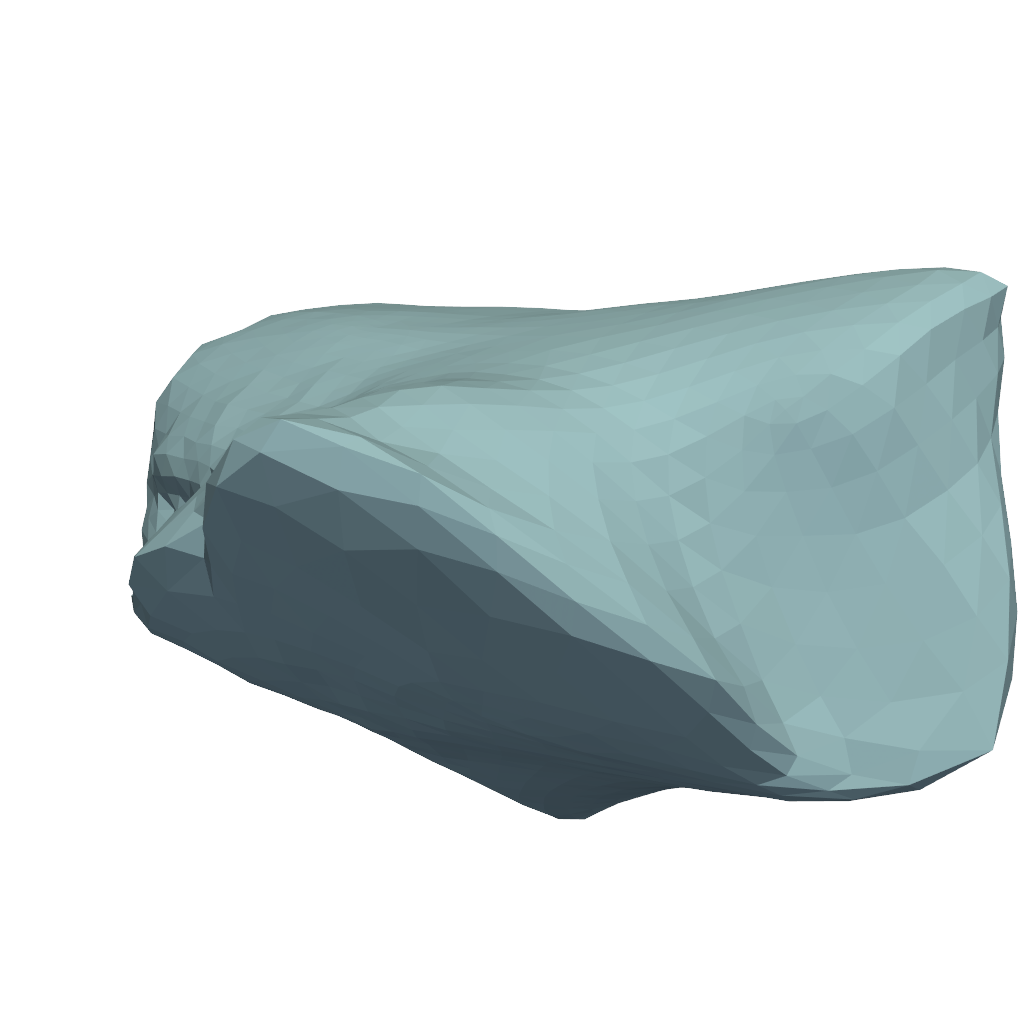} & 
        \imagecell[0.23]{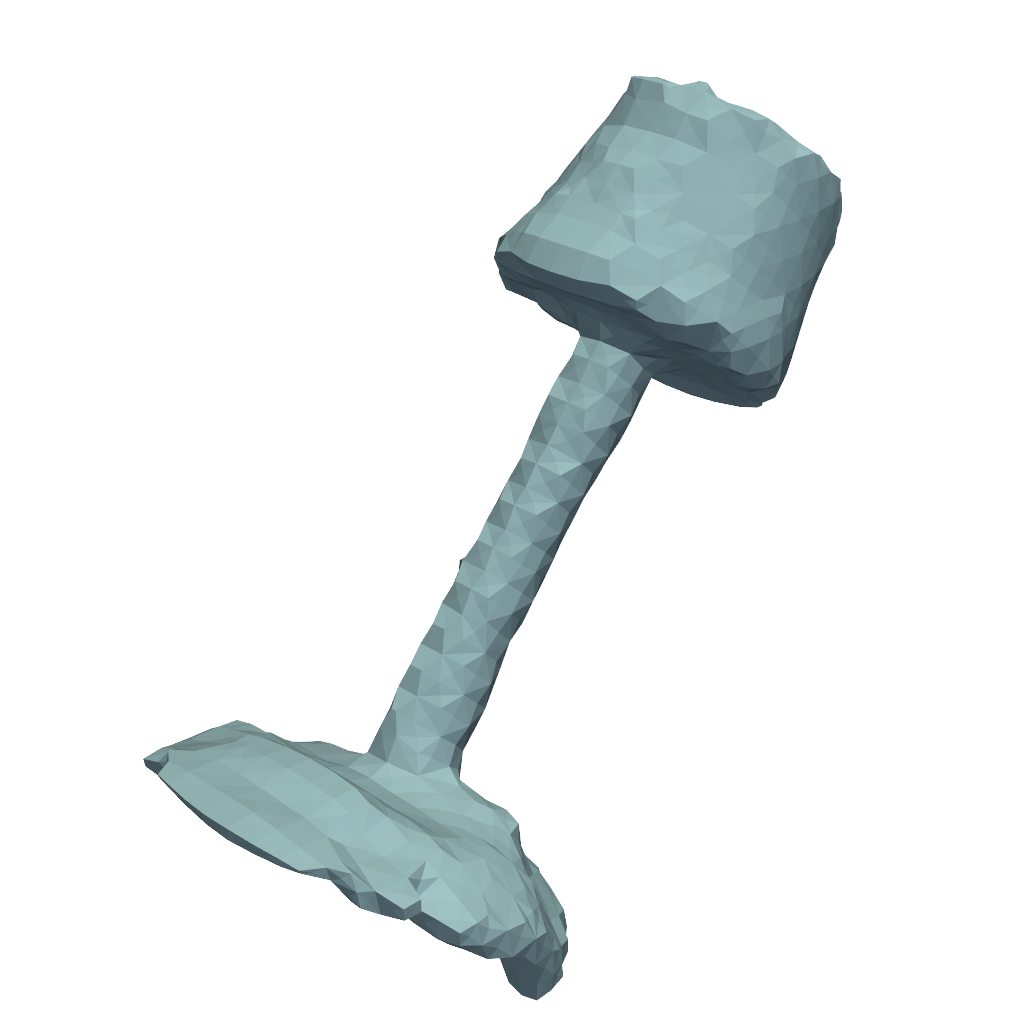} & 
        \imagecell[0.23]{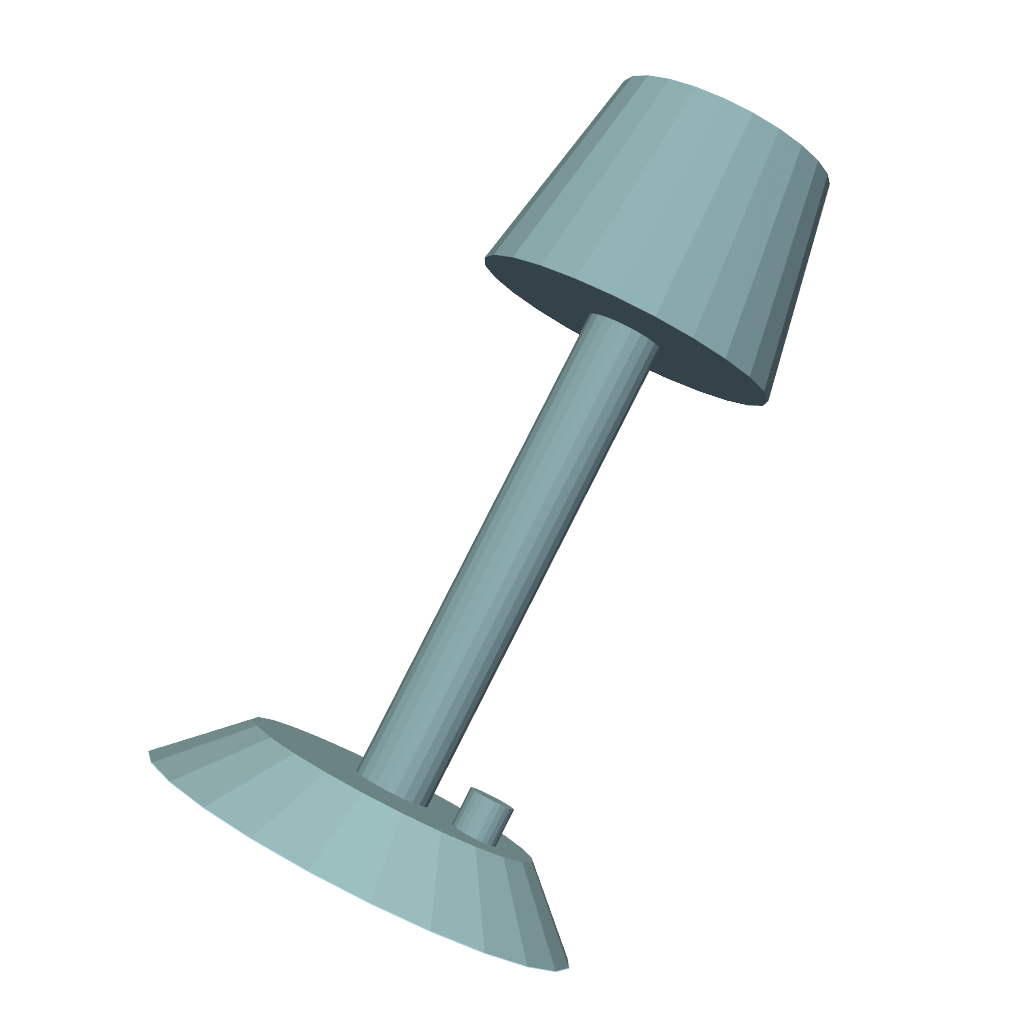}
        \\
        \imagecellspecin{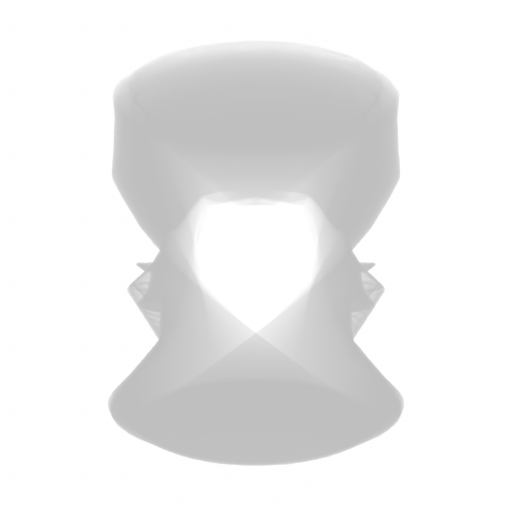} & 
        \imagecell[0.23]{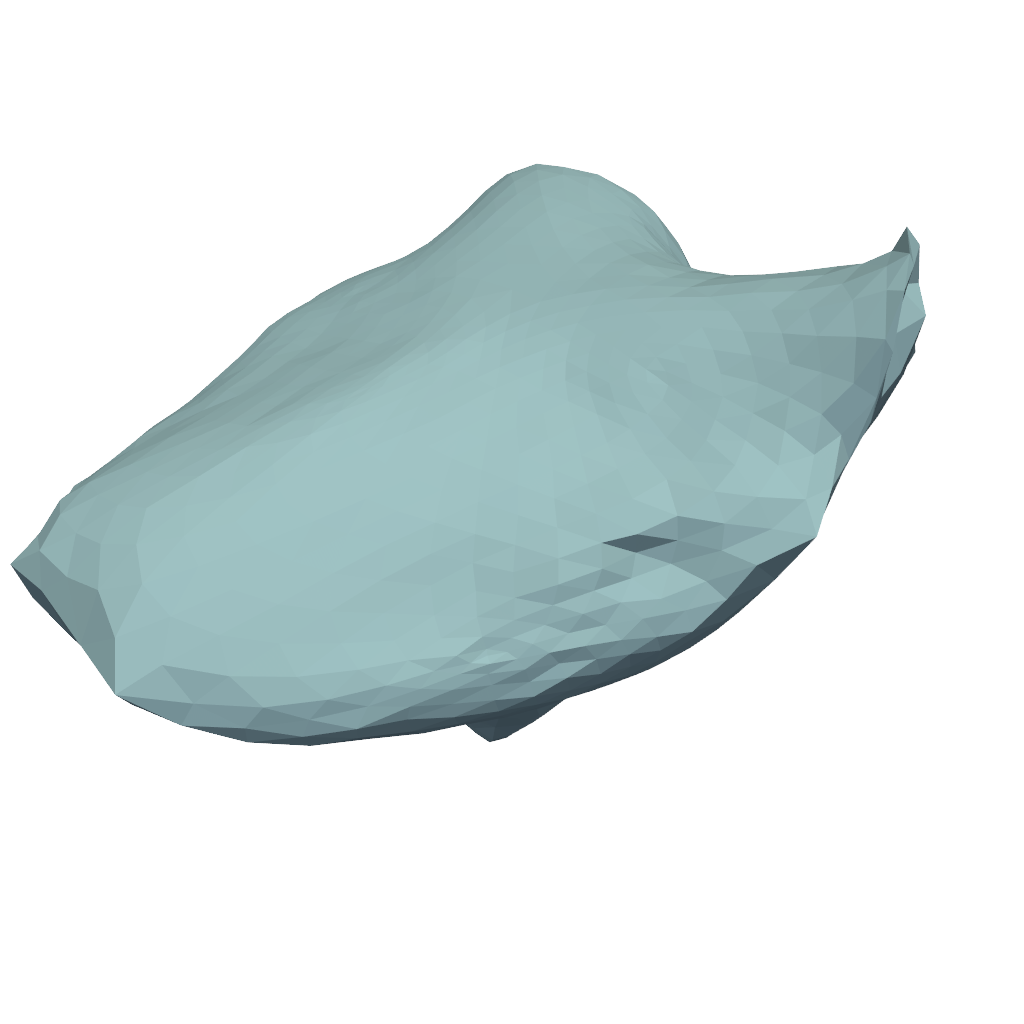} & 
        \imagecell[0.23]{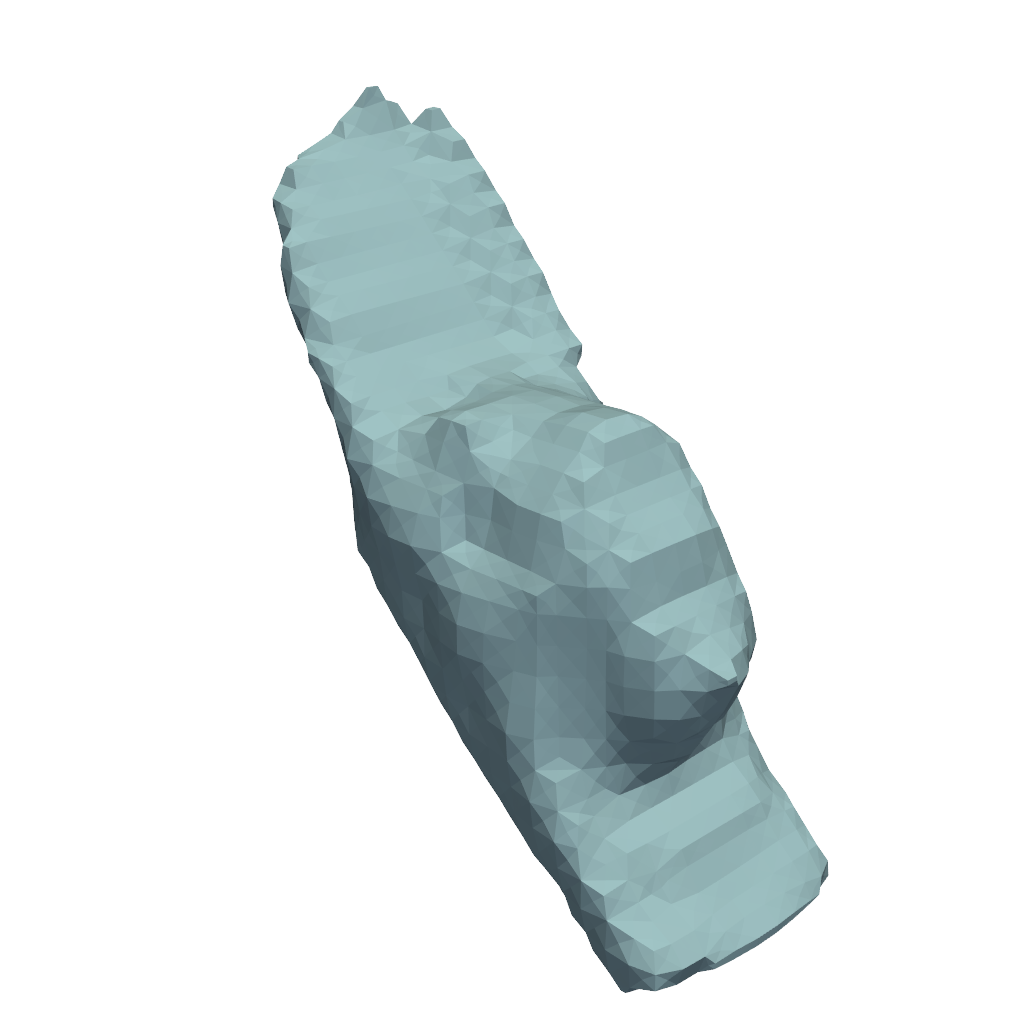} & 
        \imagecell[0.23]{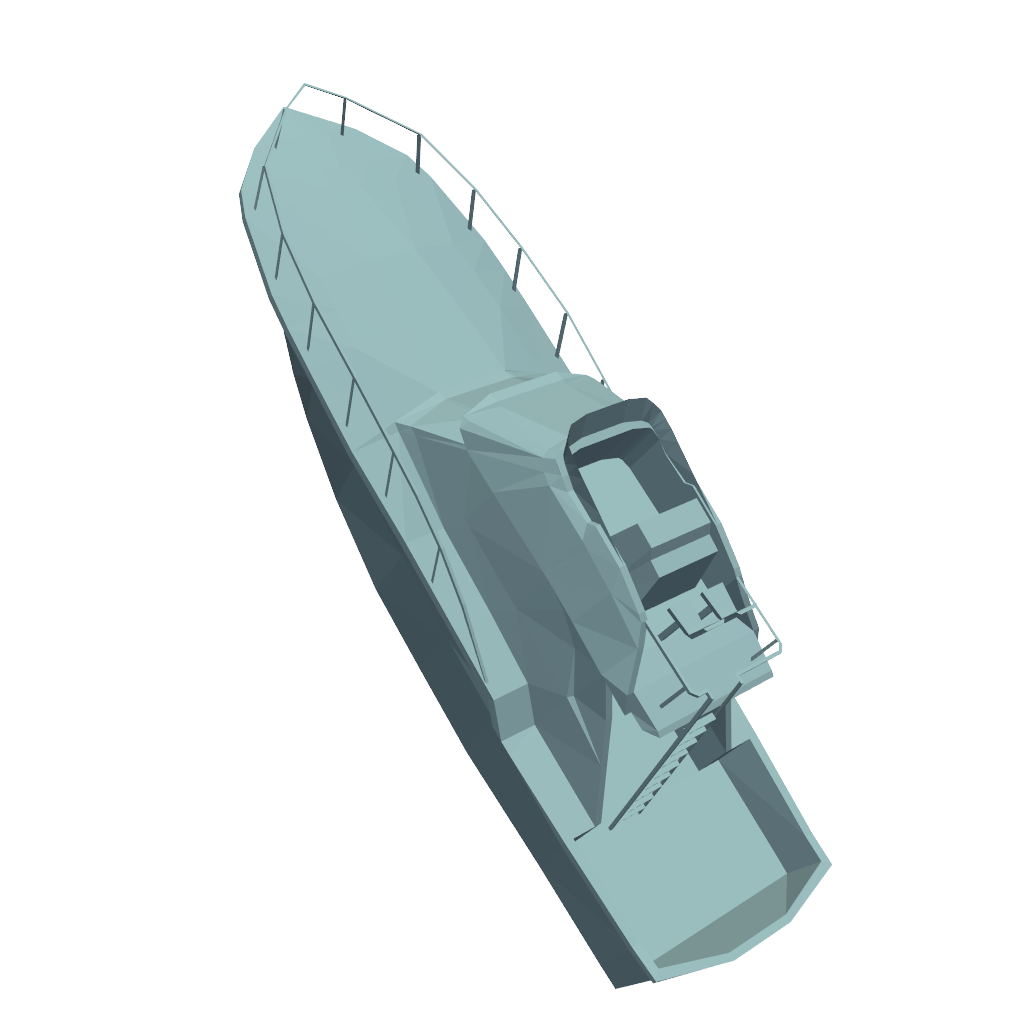}
        \\

    (a) & (b) & (c) & (d)
    \end{tabular}
    \caption{Qualitative results for optimization on rotating objects. (a) One of blurred input images. (b) State of the Art \cite{rozumnyi2021shape} result. (c) Our optimization result. (d) Ground-truth object. Our method also yields a significant superior result than the state-of-the-art work.}
    \label{fig:qualitative-optim-rotation}
    \vspace{-15pt}
\end{figure}
\begin{figure}
    \centering
    \begin{spacing}{1.0}
    \setlength\tabcolsep{0pt}
    \begin{tabular}{ccc}
    \imagecell[0.32]{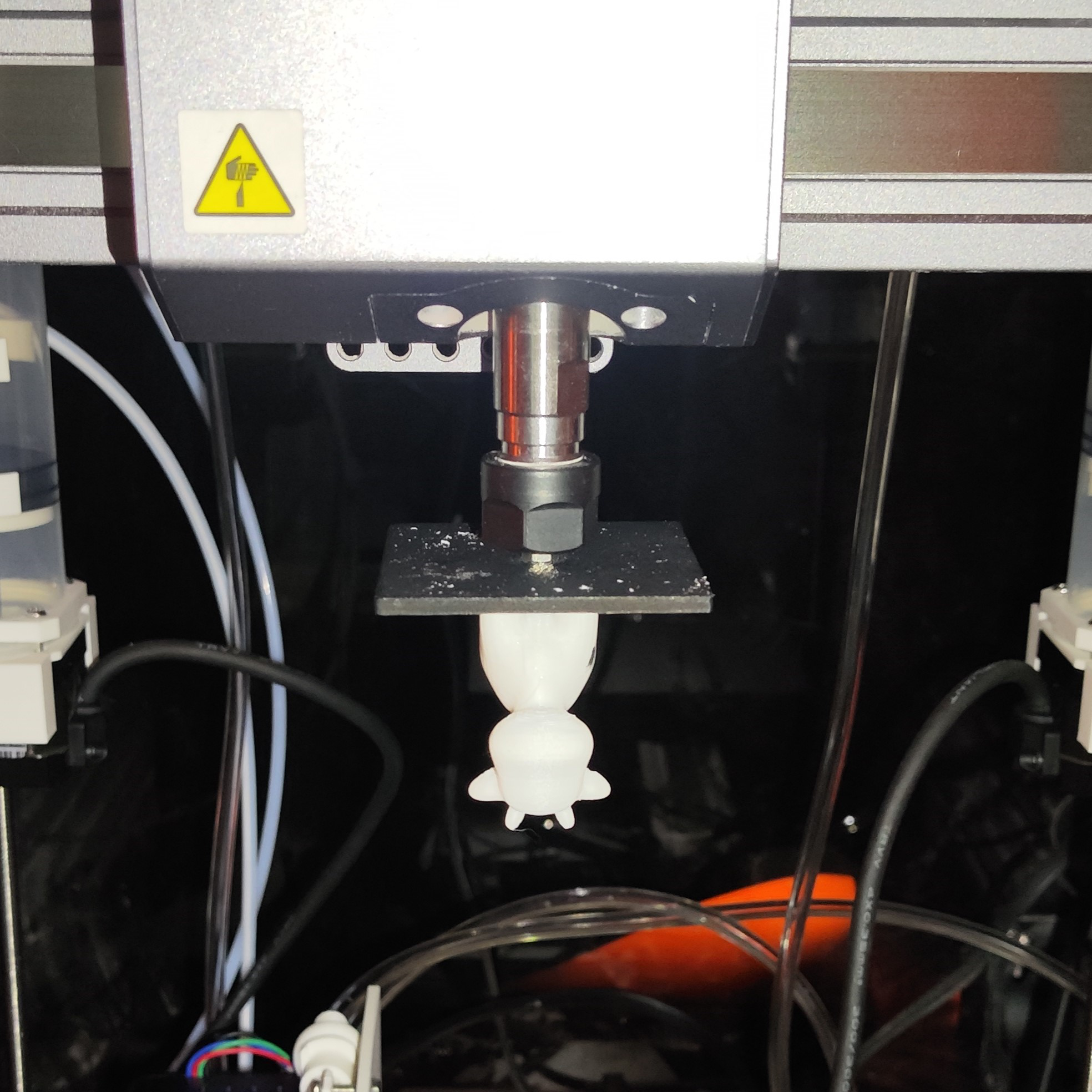} & 
    \imagecell[0.32]{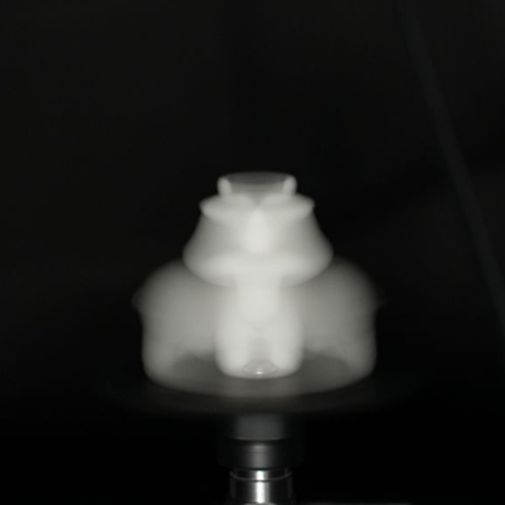} &
    \imagecell[0.32]{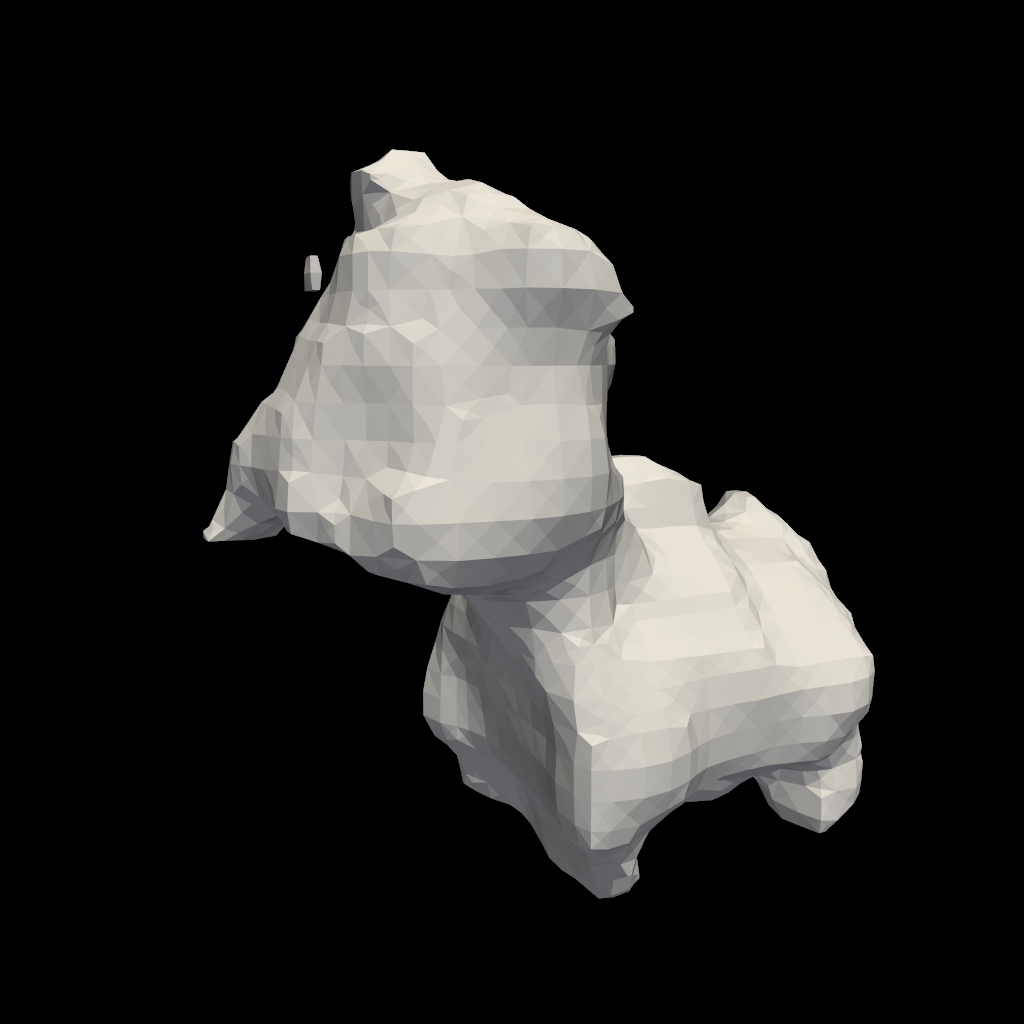} \\
    (a) Experimental & (b) Image & (d) Ours \\
    Setup & Captured & Result \\

    \end{tabular}
    \end{spacing}
    \vspace{3pt}
    \hrule
    \vspace{3pt}
    \begin{spacing}{1.0}
    \setlength\tabcolsep{0pt}
    \begin{tabular}{cccc}

    Captured & \cite{rozumnyi2021shape} & Ours & G.T. \\
    
    \imagecell[0.229]{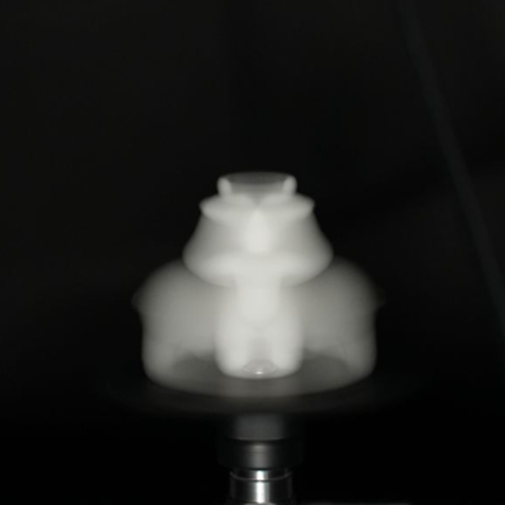} &
    \imagecell[0.229]{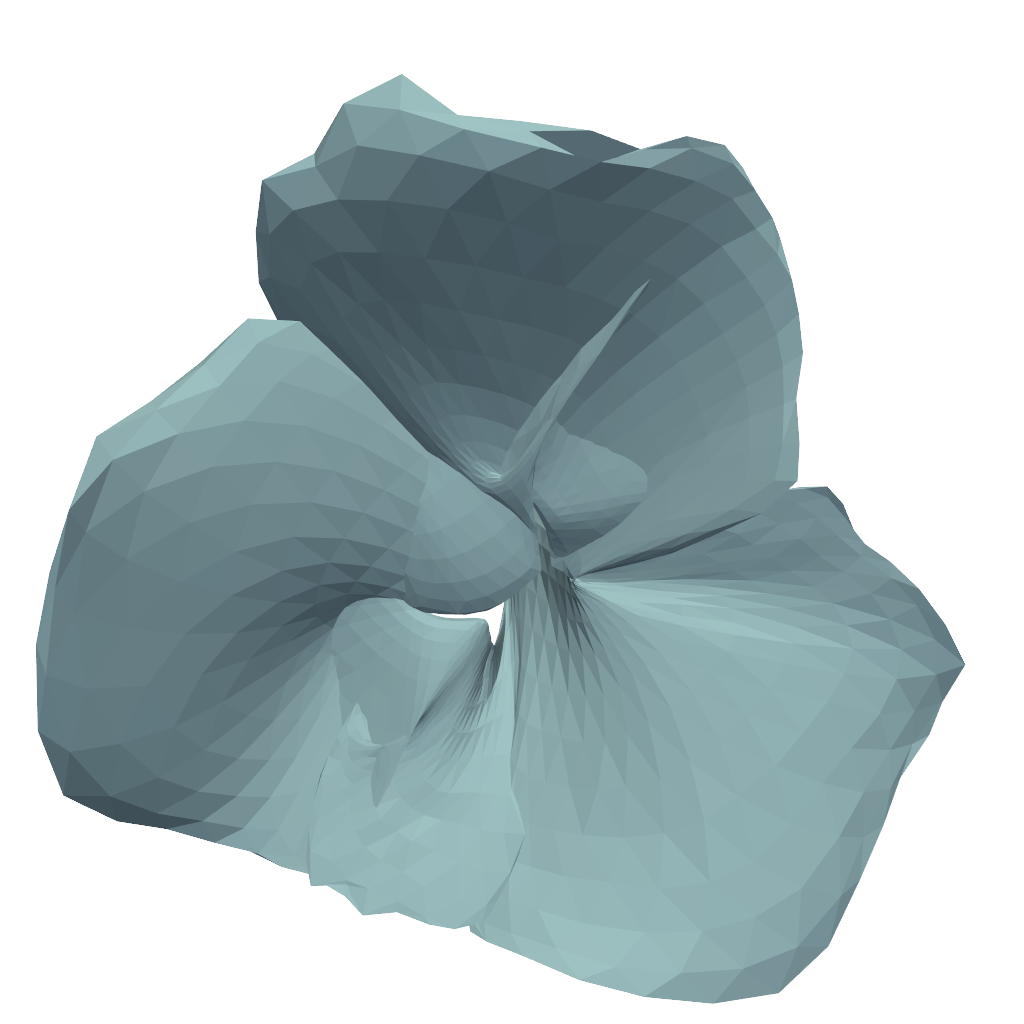} &
    \imagecell[0.229]{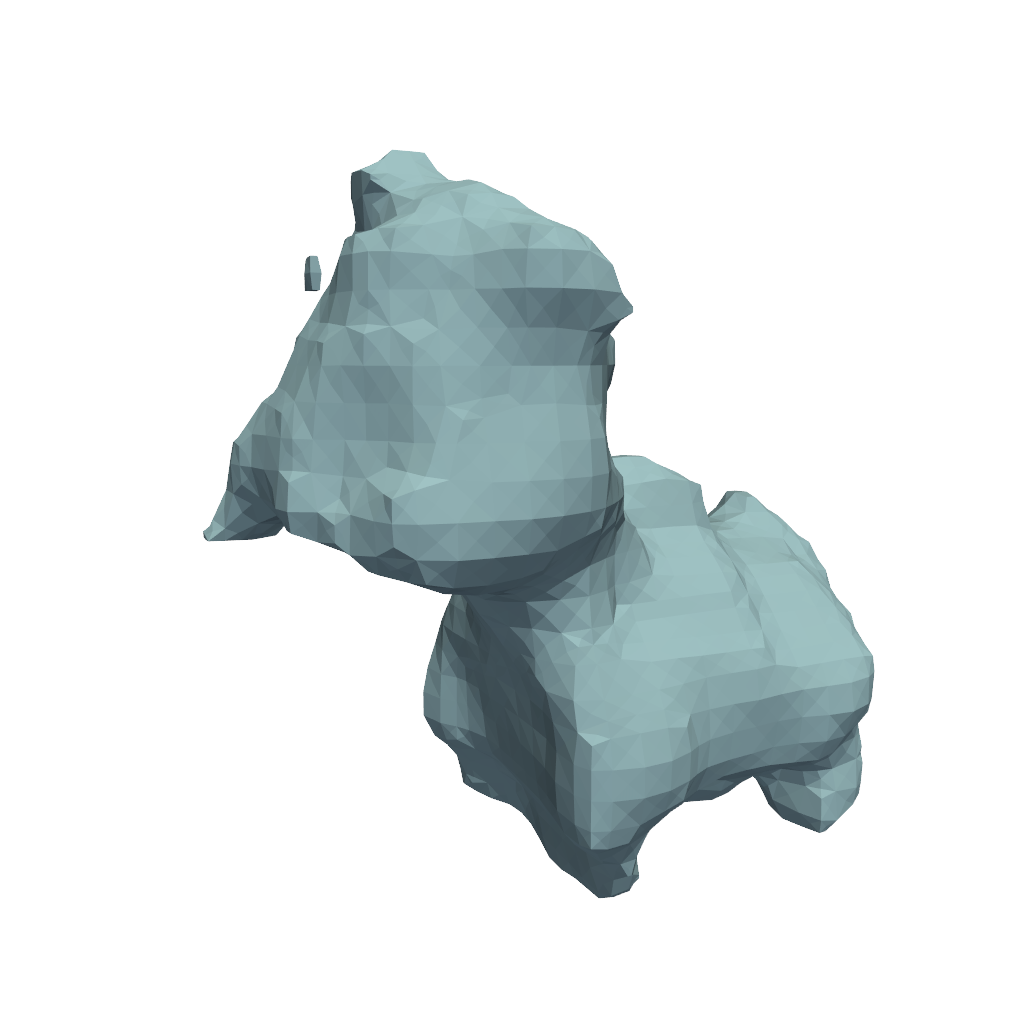} &
    \imagecell[0.229]{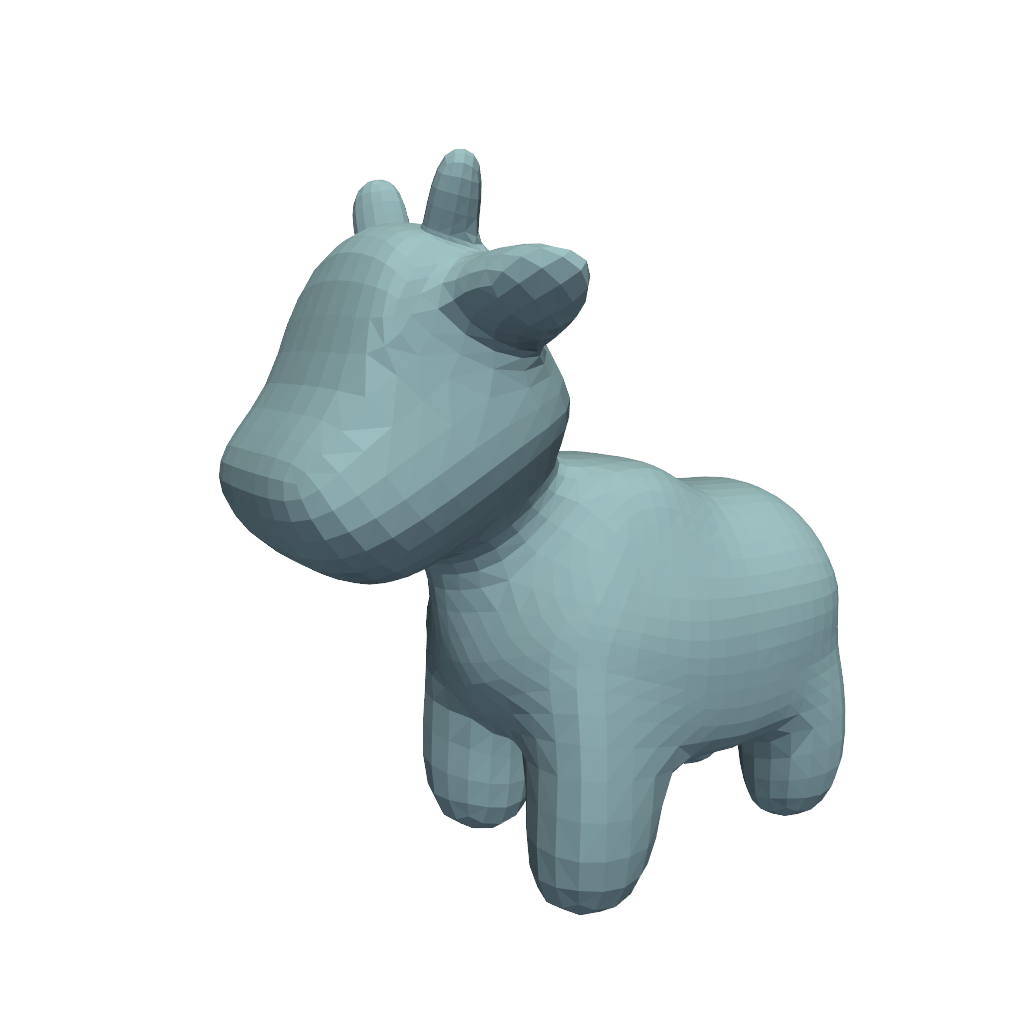} \\

    \vspace{-10pt} \\

    \imagecell[0.229]{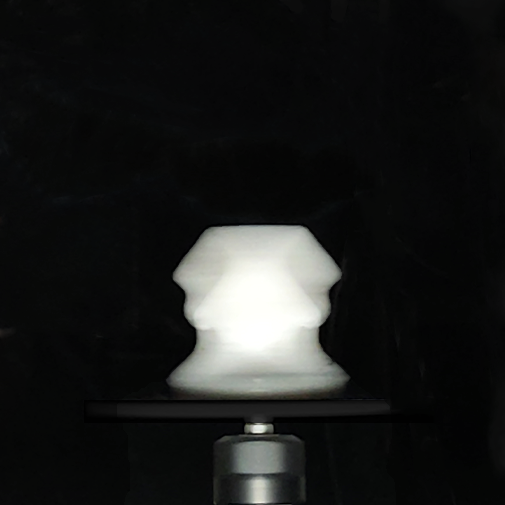} &
    \imagecell[0.229]{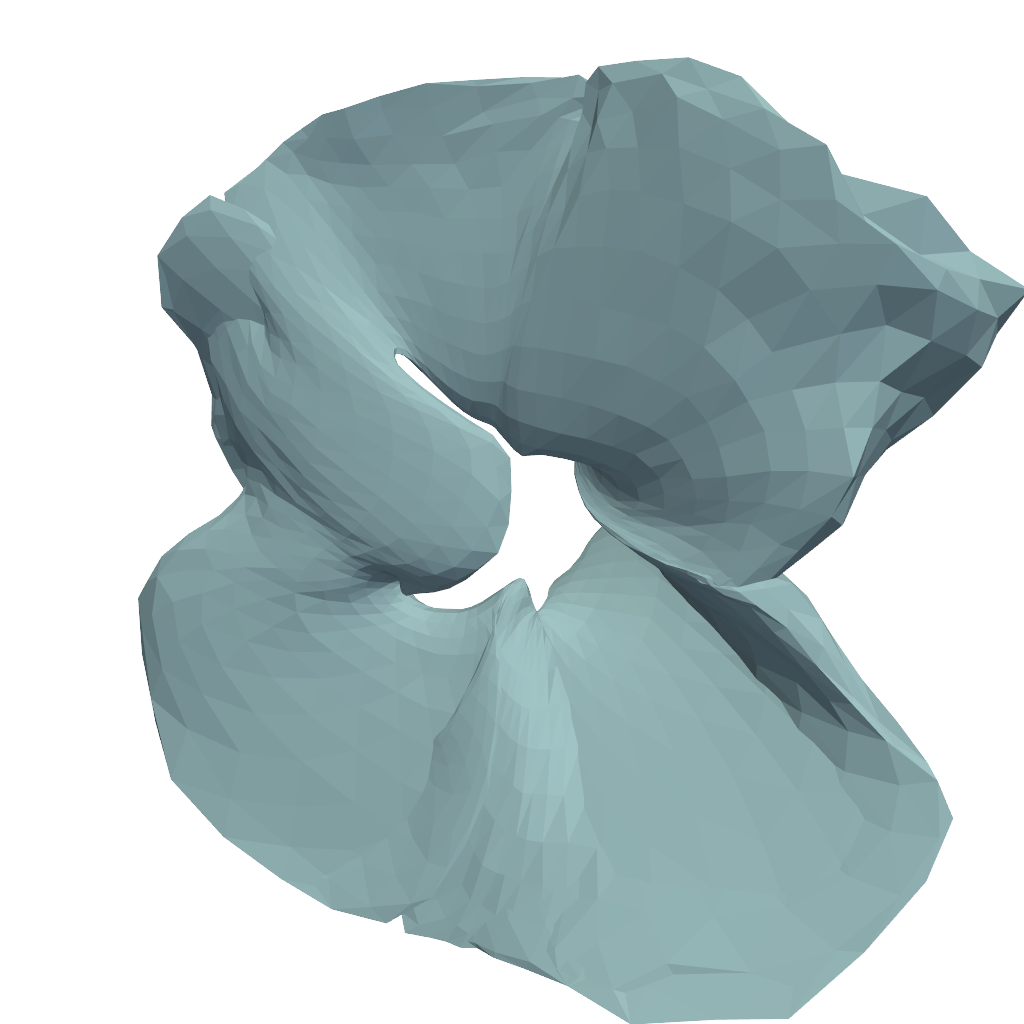} &
    \imagecell[0.229]{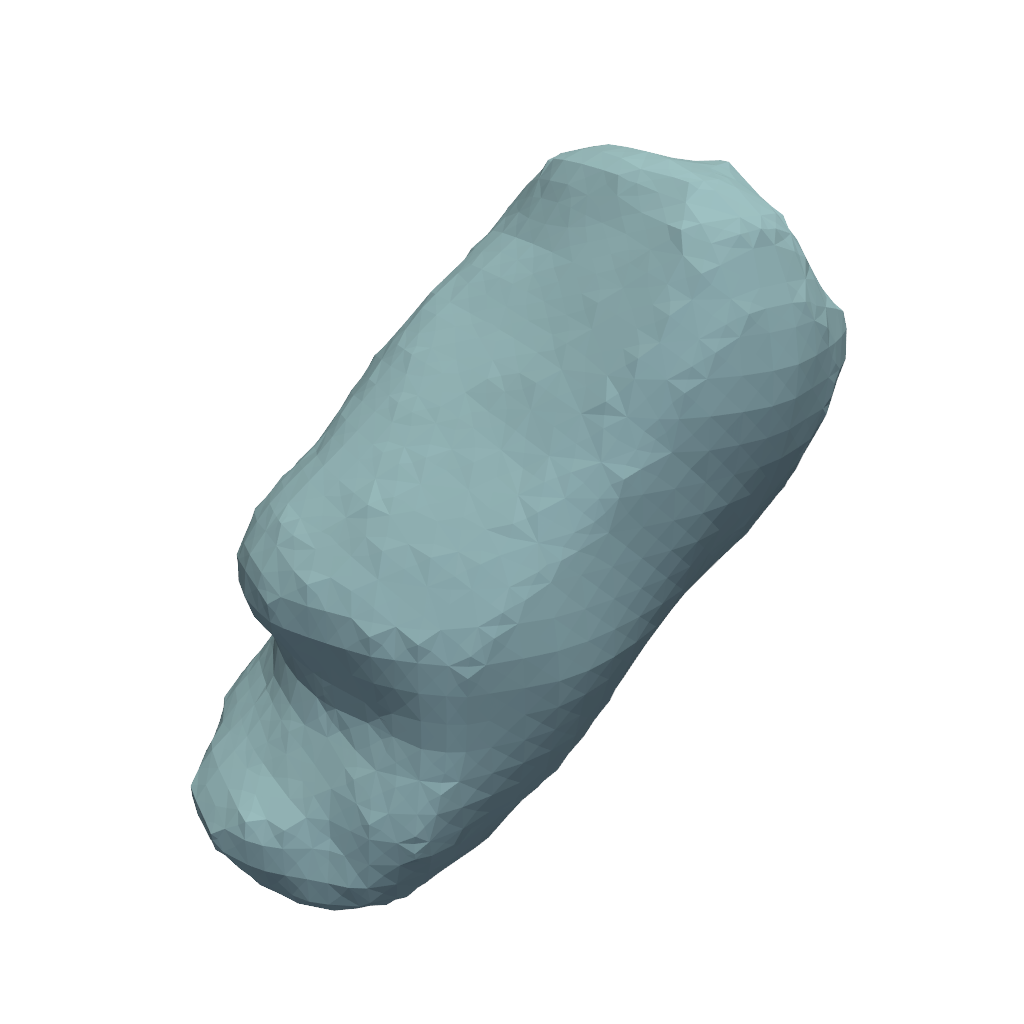} &
    \imagecell[0.229]{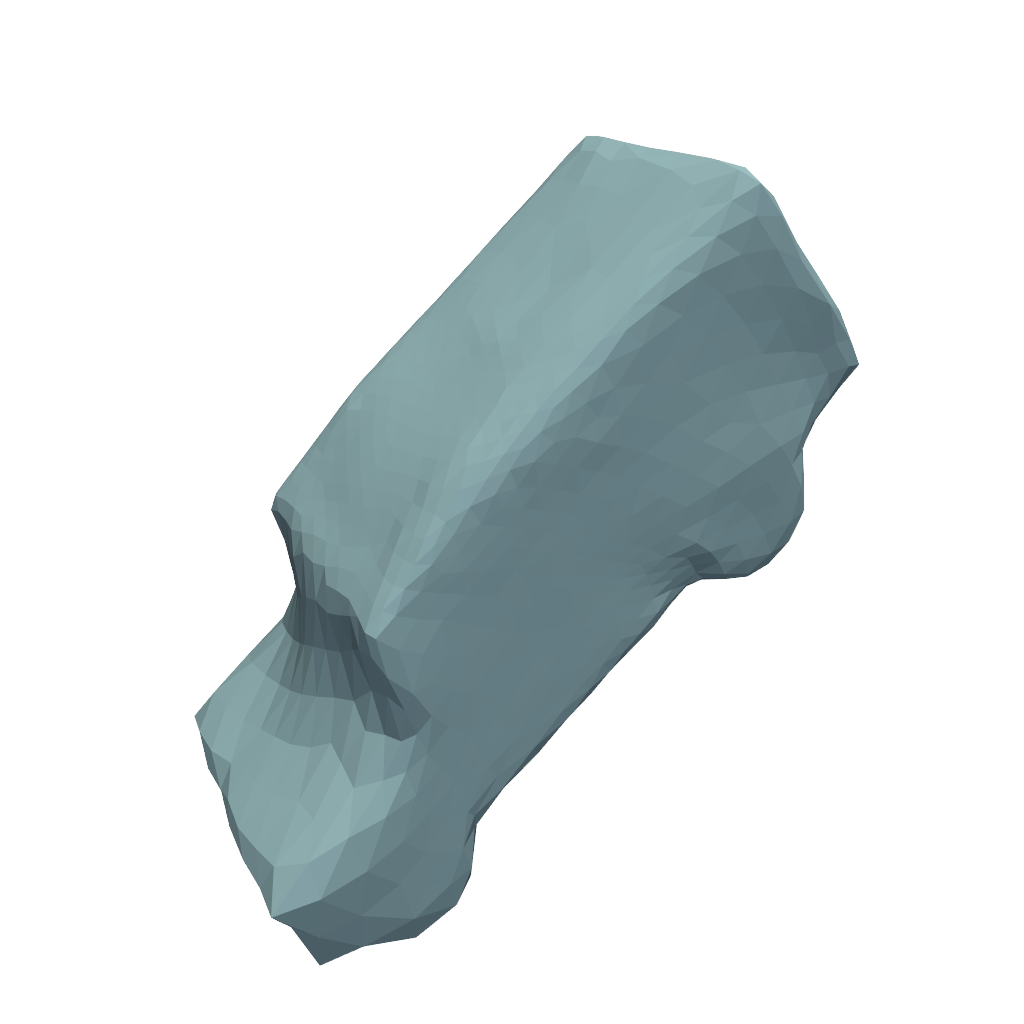} \\

    \vspace{-10pt} \\

    \imagecell[0.229]{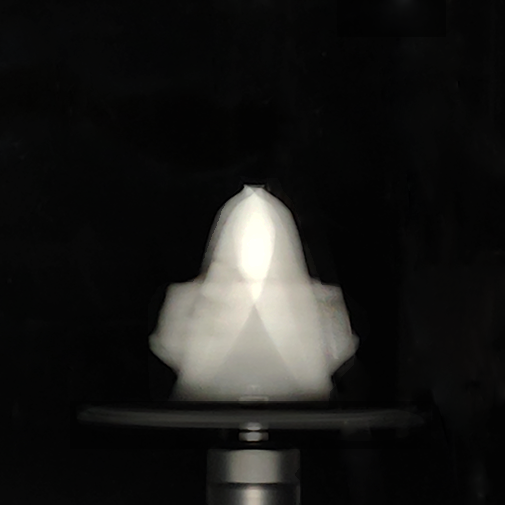} &
    \imagecell[0.229]{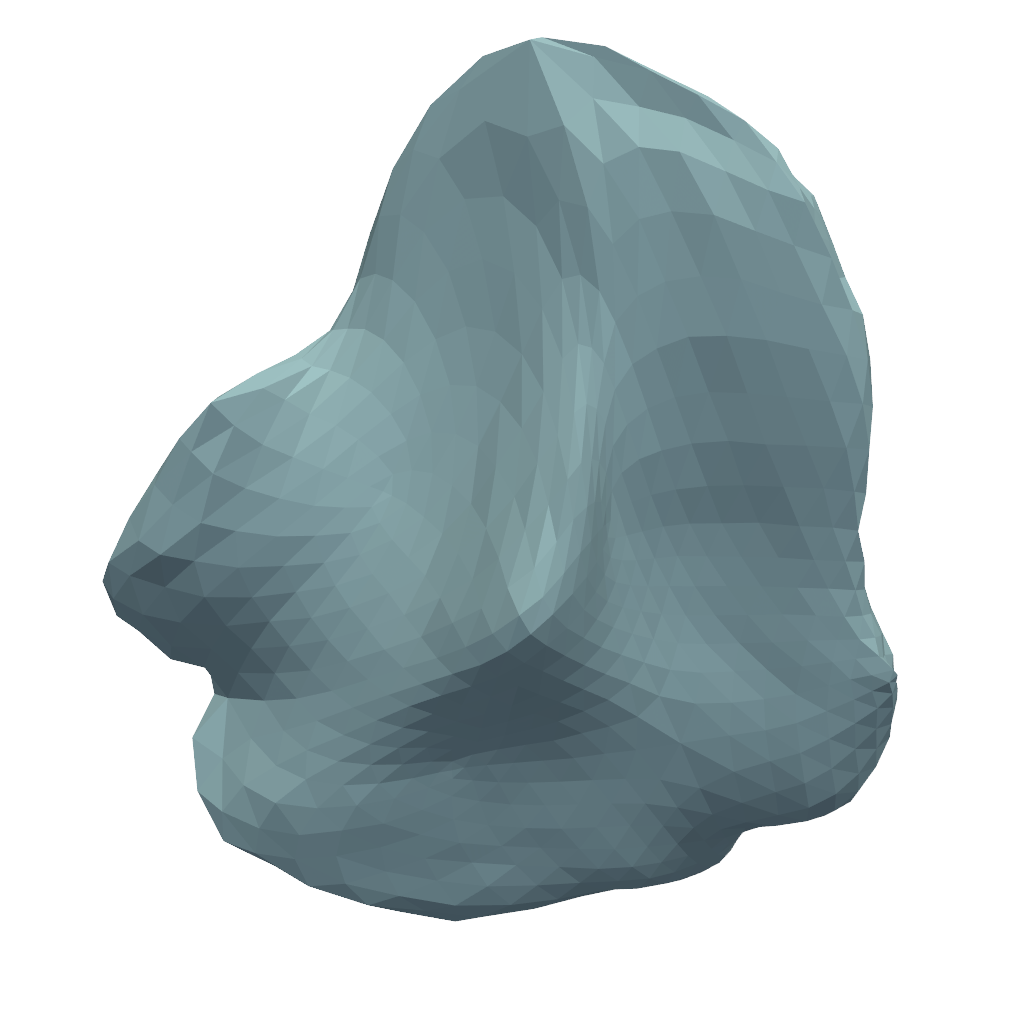} &
    \imagecell[0.229]{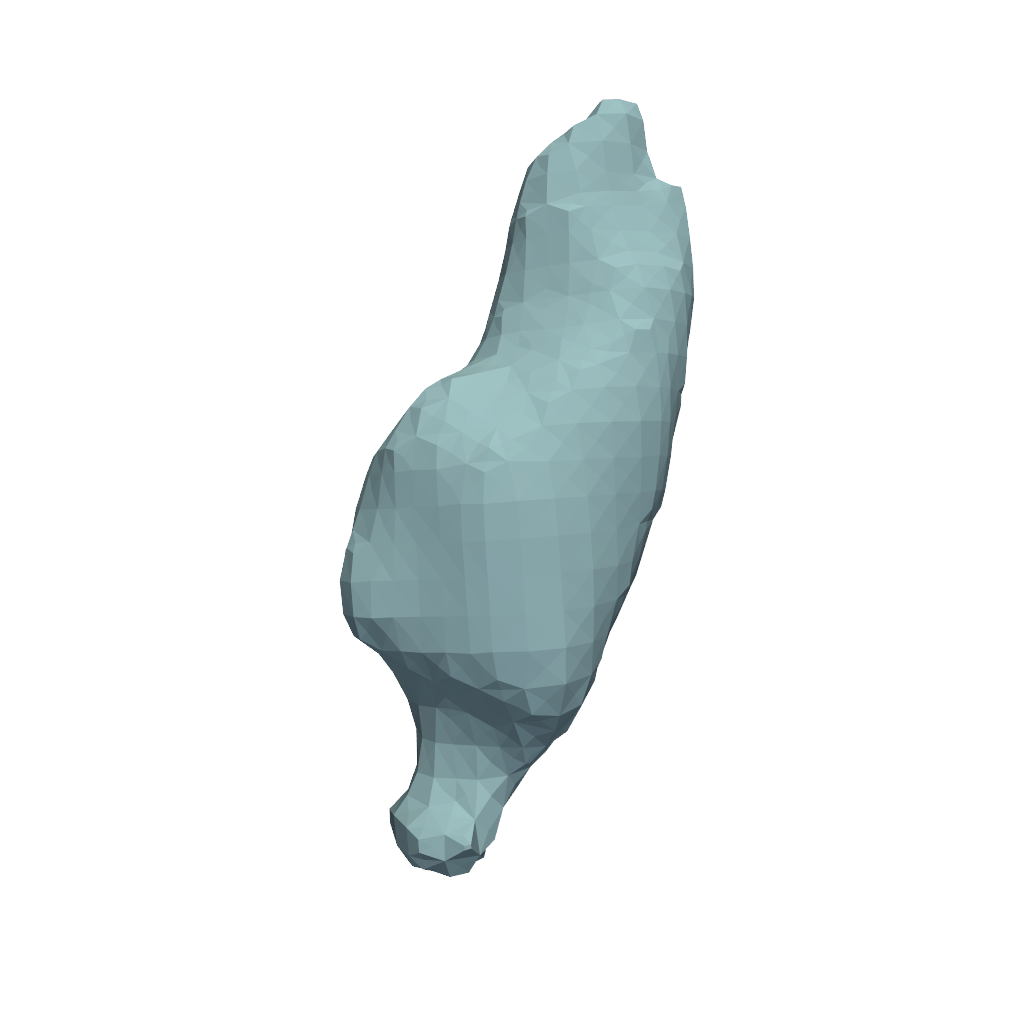} &
    \imagecell[0.229]{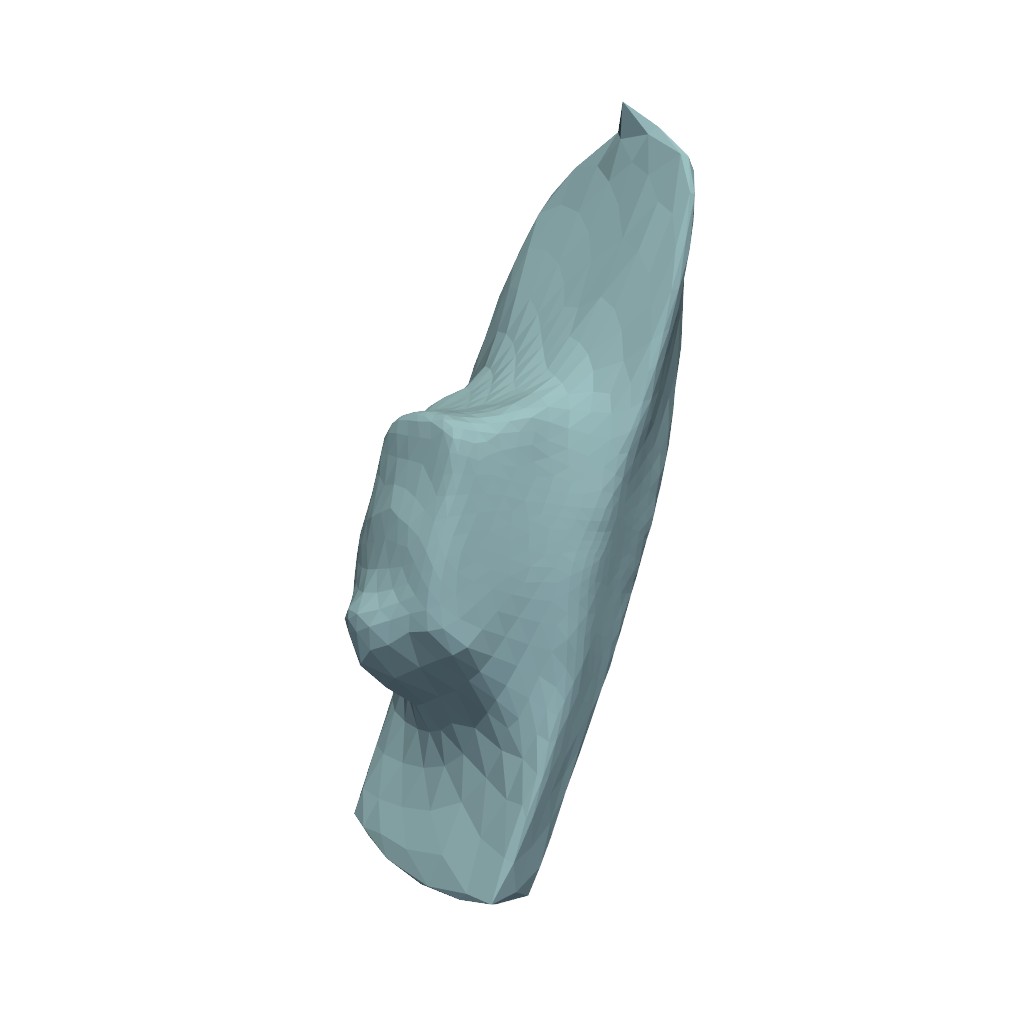} \\

    \vspace{-10pt} \\

    \imagecell[0.229]{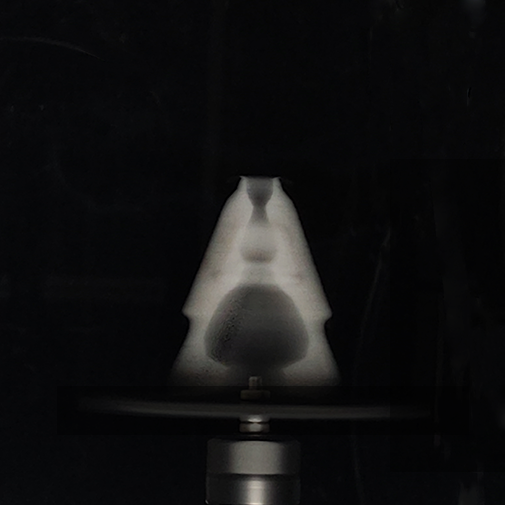} &
    \imagecell[0.229]{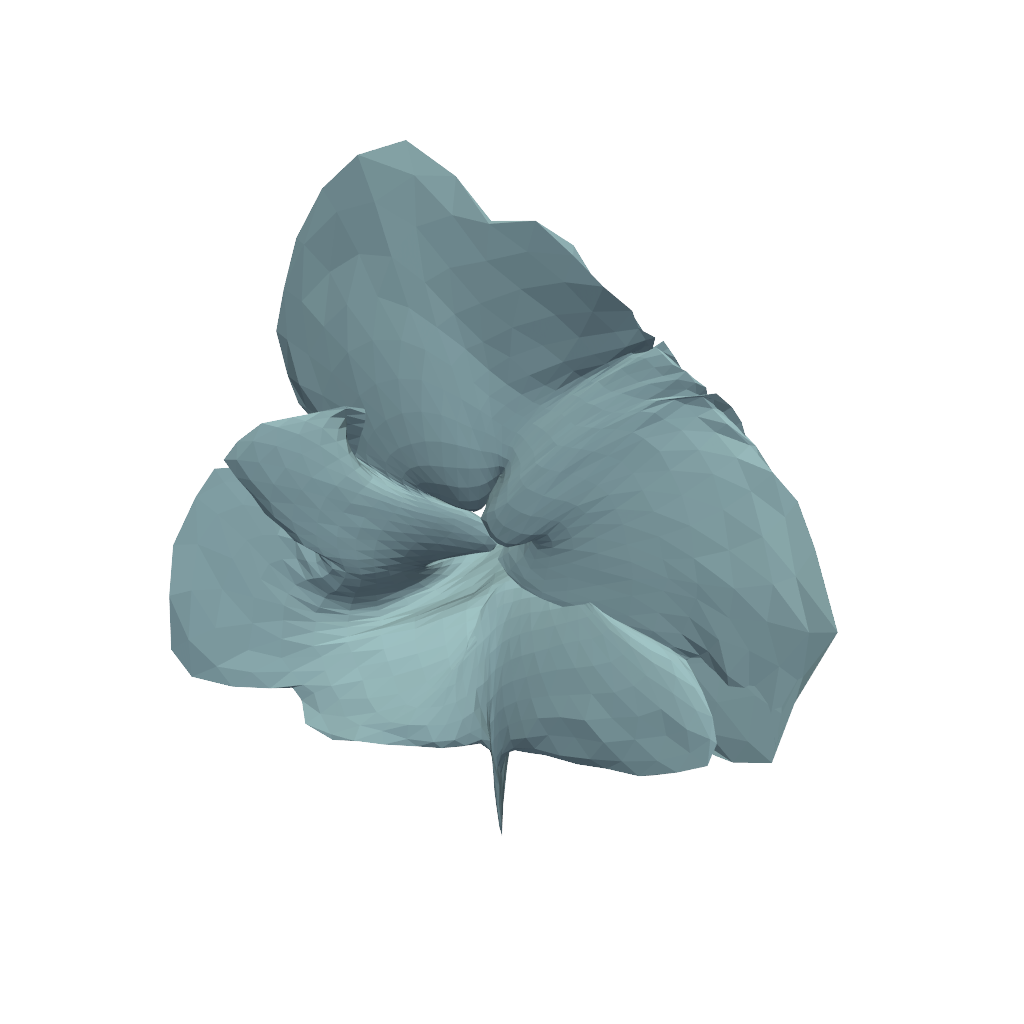} &
    \imagecell[0.229]{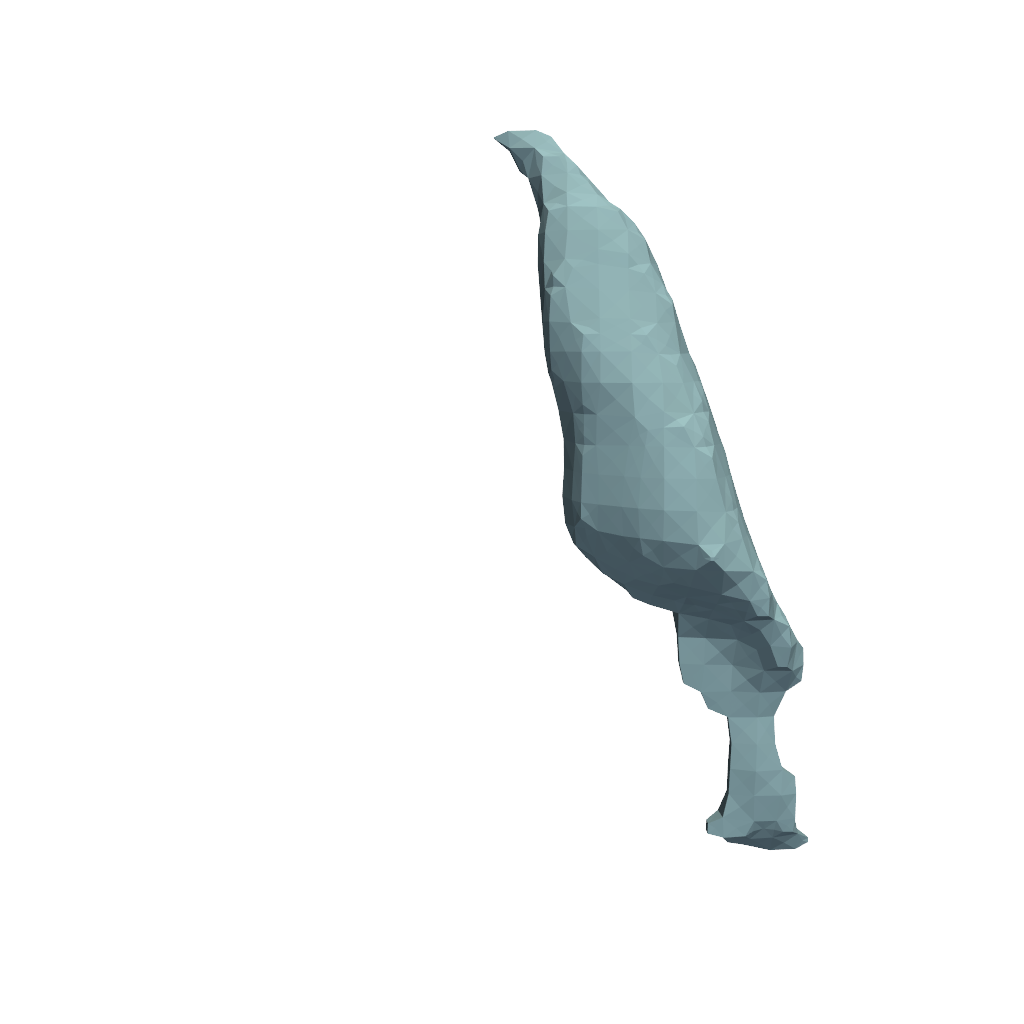} &
    \imagecell[0.229]{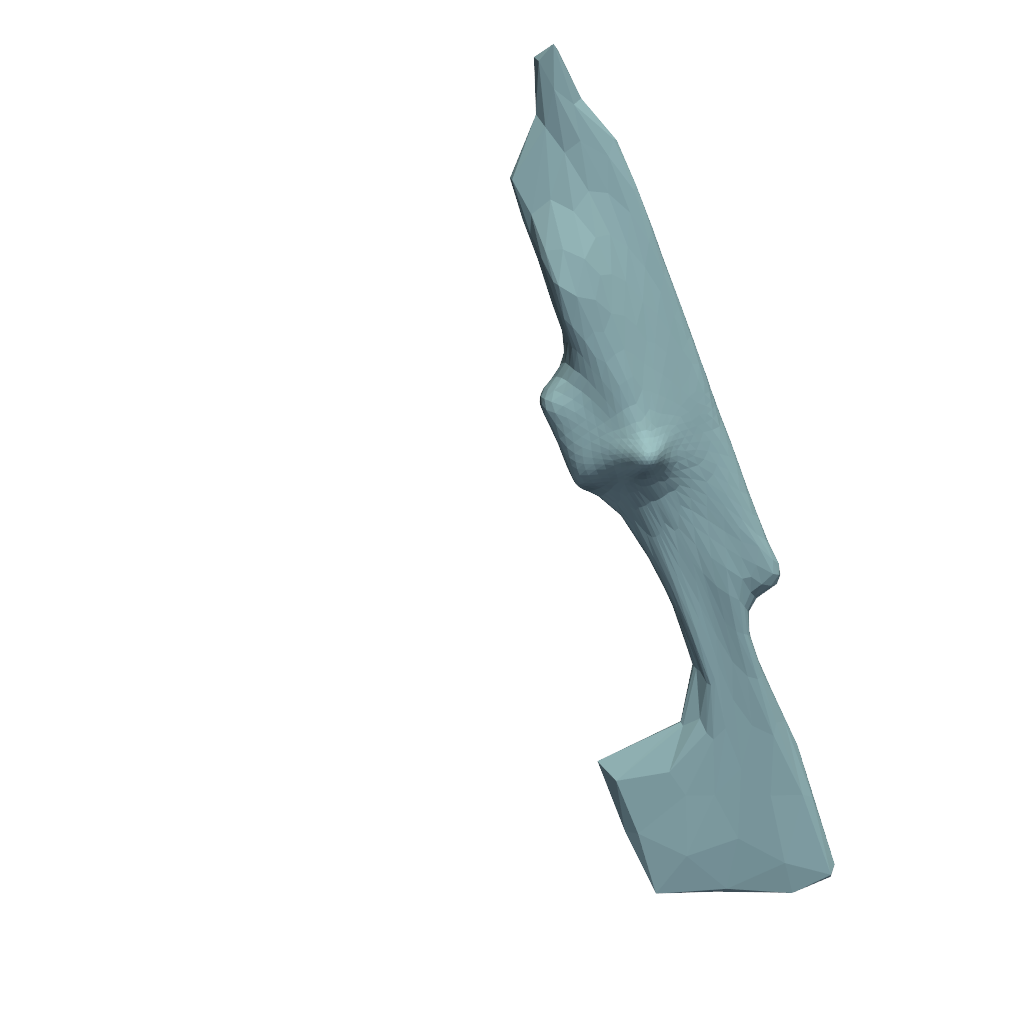} \\

    \end{tabular}
    \end{spacing}
    \caption{
    Real-world experiment. (a) Our experimental setup. (b) The original image captured. (c) Ours optimization result. 
    The remaining rows: More real-world examples. 
    Quantitatively, our method achieves a superior blurred PSNR score of 24.52 dB compared to 12.51 dB for \cite{rozumnyi2021shape}.
    Our method is capable of reconstructing 3D objects from real-world motion-blurred images.
    }
    
    \label{fig:real-world-exp}
    \vspace{-15pt}
\end{figure}

In this experiment, we evaluate shape recovery for objects undergoing ultra-fast rotational motion, benchmarking against the state-of-the-art \cite{rozumnyi2021shape}.
Here, we observe that multiple feasible 3D shape solutions may correspond to the same blurred image; a detailed analysis is provided in \Cref{sec:supp:3d-losses}.
Consequently, traditional 3D evaluation metrics (\eg, 3D IoU, Chamfer Distance) are not suitable.
Therefore, we assess the similarity between the rotational-blurred images rendered from the reconstructed objects and the ground truth, quantified using PSNR.

Similar to \cref{sub:rozumnyi-comparison-details}, for rotational motion, a similar trend of superior performance is observed for our method compared to \cite{rozumnyi2021shape}. 
As shown in \cref{tab:comparison-sfb,fig:qualitative-optim-rotation}, our method achieves a significantly superior performance and succeeds in a high-quality 3D reconstructions even under extreme rotational scenarios.

\vspace{-5pt}

\subsection{Real-World Results} 
\label{sub-sec:exp-real-world}

\vspace{-5pt}

Next, we evaluate our method on real-world images. We capture a front view of 3D-printed rotating objects at 100 Hz with a camera exposure time of $1/100$ s.
Since the data was captured in a controlled studio environment with a black background, we extract the object alpha masks based on pixel intensity thresholds to serve as supervision signals.
After preprocessing, including cropping and brightness correction, we perform rotational shape optimization using the same settings as described in \cref{sub-sub-sec:optim-rotation}.

Results and evaluations are presented in \cref{fig:real-world-exp}. 
The 3D print technique allows us to establish ground truth for evaluation.
As summarized, our method achieves a superior blurred PSNR score of 24.52 dB compared to 12.51 dB for \cite{rozumnyi2021shape}, demonstrating a significant improvement.
Qualitatively, due to the imperfect pose and noise introduced in real data, the recovered shape exhibits slight artifacts compared to the synthetic recovery results.
Nevertheless, our method successfully recovers reasonable shapes, demonstrating its capability to handle real-world data effectively.

\vspace{-5pt}

\subsection{Ablation Study}
\label{sub-sec:exp-ablation}

\vspace{-5pt}

Finally, we conduct an ablation study to validate our key design choices. 
Specifically, we analyze the efficiency improvements of our method compared to the codebase SoftRas, and assess the robustness and gradient quality of our method against high-performance Nvdiffrast under the extreme motion blur scenarios.

\vspace{-13pt}

\paragraph{Comparison with SoftRas}
\begin{figure}
    \centering
    \includegraphics[width=\linewidth]{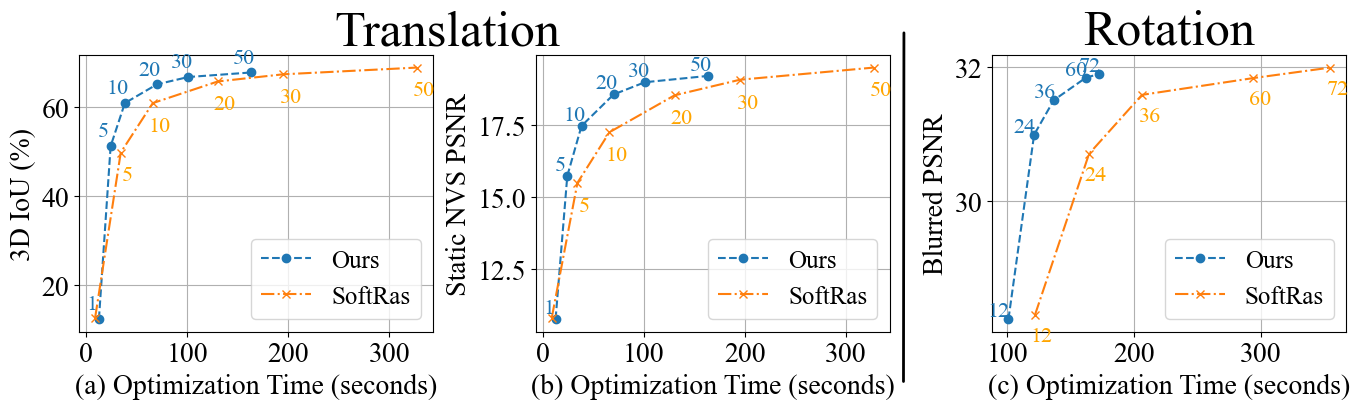}
    \caption{Optimization results for objects undergoing translation and rotation. We draw optimization time v.s. performance curves for our method and SoftRas, where each point indicate different number of samples.}
    \label{tab:quantitative-optim}
    \vspace{-15pt}
\end{figure} 

Our method is built upon the SoftRas framework. Therefore, we compare our method against SoftRas in terms of optimization time and reconstruction quality.
As shown in \cref{tab:quantitative-optim}, several interesting points can be observed.  
First, increasing the number of samples improves performance but also increases optimization time. This is reasonable as more samples result in better blur simulation, yielding better shape recovery performance and longer time.
Second, our method exhibits strong efficiency over SoftRas.
Given the same number of samples, our method achieves significantly faster optimization time. 
Conversely, for the same optimization time, our method yields superior reconstruction quality.

\vspace{-13pt}

\paragraph{Comparison with Nvdiffrast} 
\label{para:failure-cases-for-nvdiffrast}

\begin{figure}
\centering
\begin{spacing}{0.5}
\includegraphics[width=0.7\linewidth]{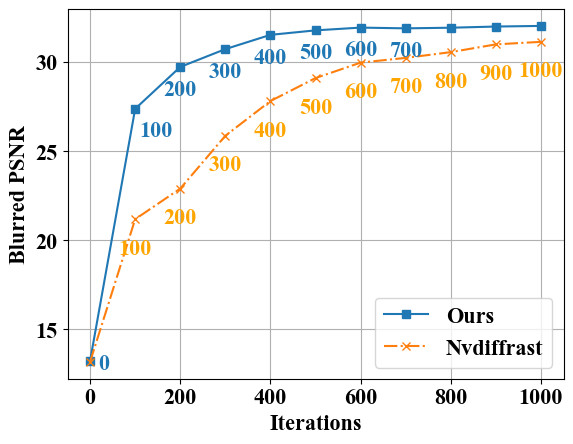}
\end{spacing}
\caption{Comparison of convergence rates between our method and Nvdiffrast w.r.t. number of iterations. Labels denote numbers of iterations. The convergence rate of Nvdiffrast is significantly slower than that of our method, which we attribute to its weaker pixel-wise gradients.}

\label{fig:nvdiffrast-convergence-speed}
\vspace{-10pt}
\end{figure}

Next, we present a comparison of our method's gradient quality and robustness against Nvdiffrast.
We first quantitatively assess the convergence behaviors of both methods as a function of the number of iterations. As illustrated in \cref{fig:nvdiffrast-convergence-speed}, our method demonstrates a significantly faster convergence rate. This is attributable to the stronger and more stable gradients generated by our approach, in contrast to the comparatively weaker gradients produced by Nvdiffrast.

\begin{figure}
    \centering
    \begin{spacing}{1.0}
    \setlength\tabcolsep{0pt}
    \begin{tabular}{cccc}
    \imagecell[0.23]{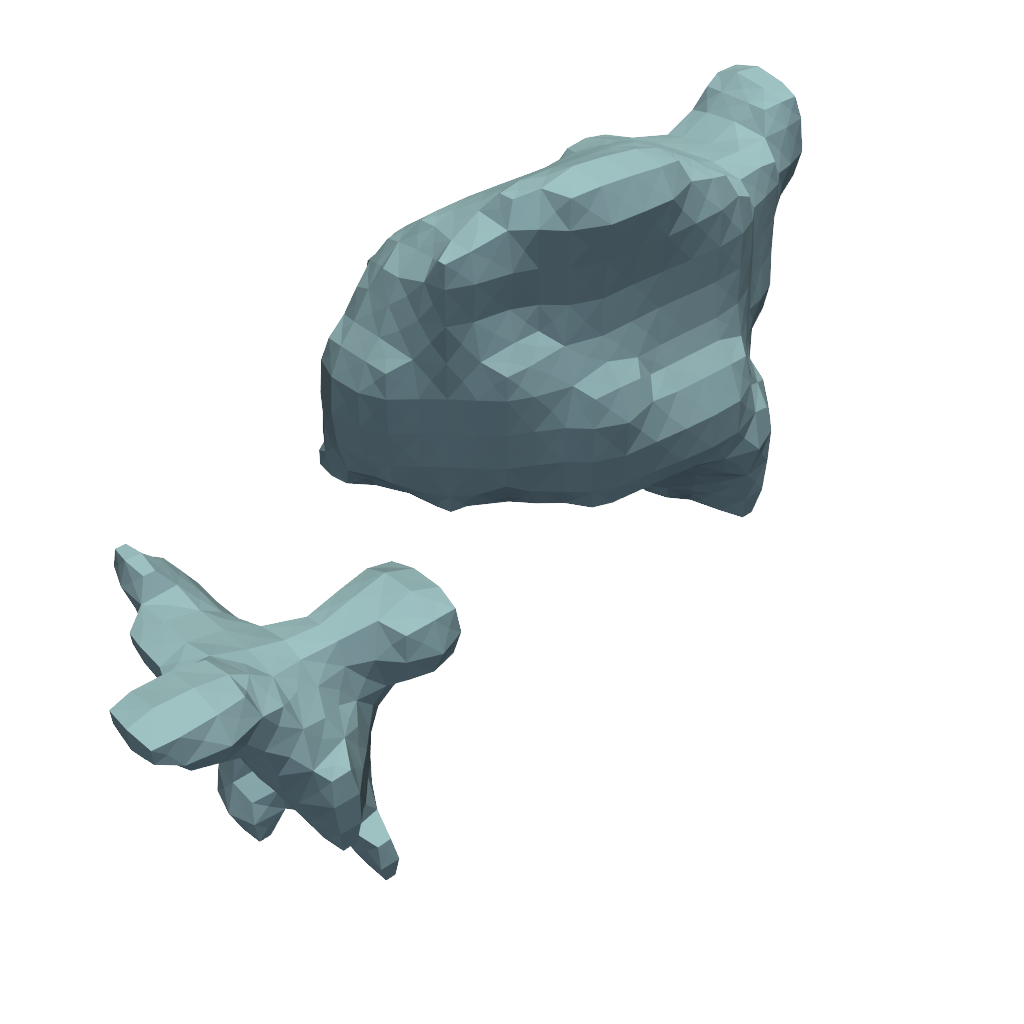} & 
    \imagecell[0.23]{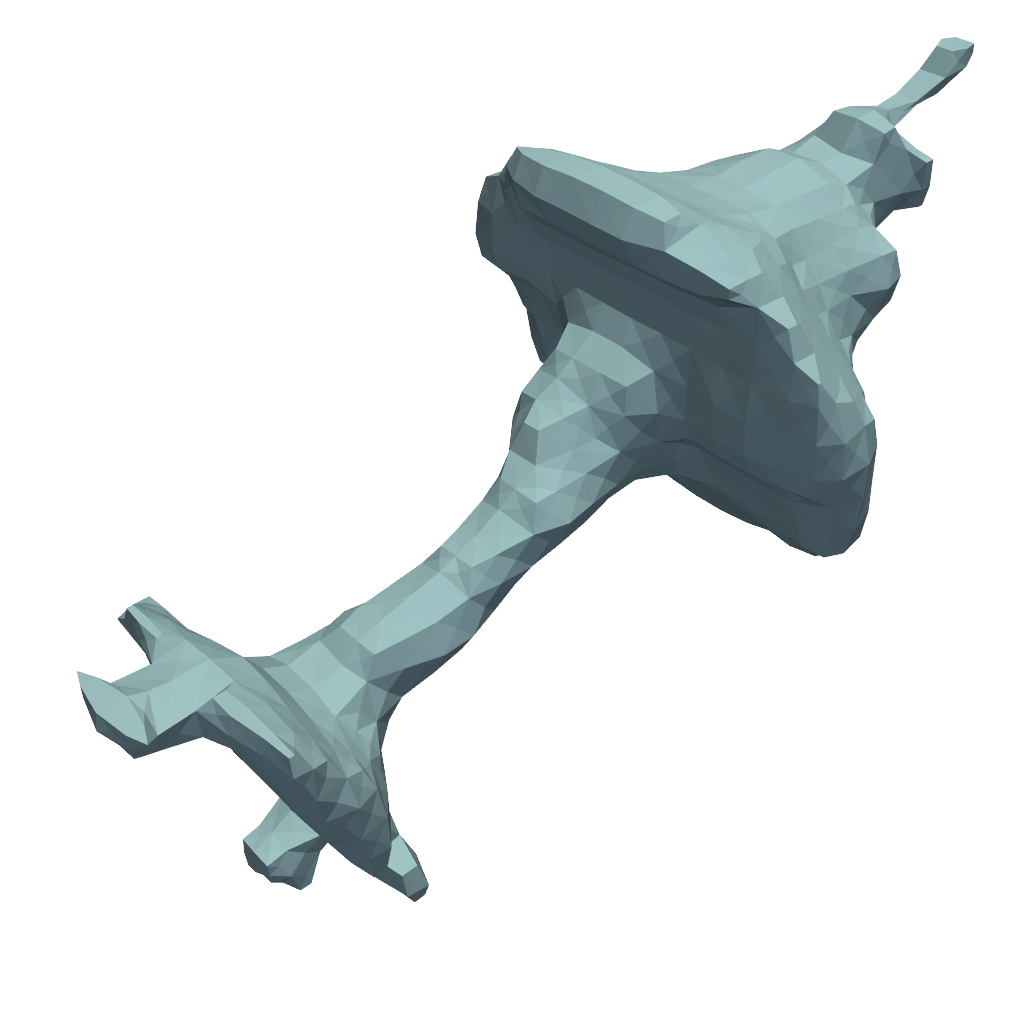} & 
    \imagecell[0.23]{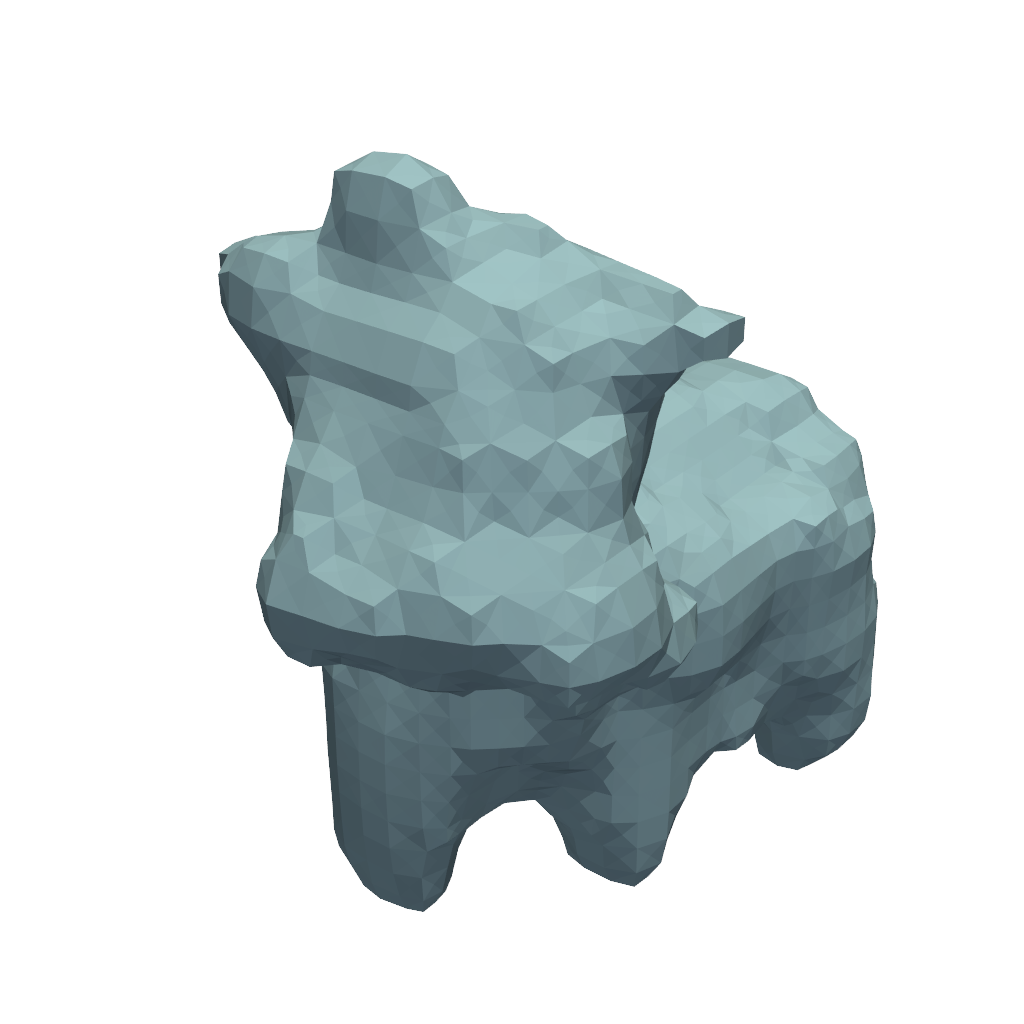} & 
    \imagecell[0.23]{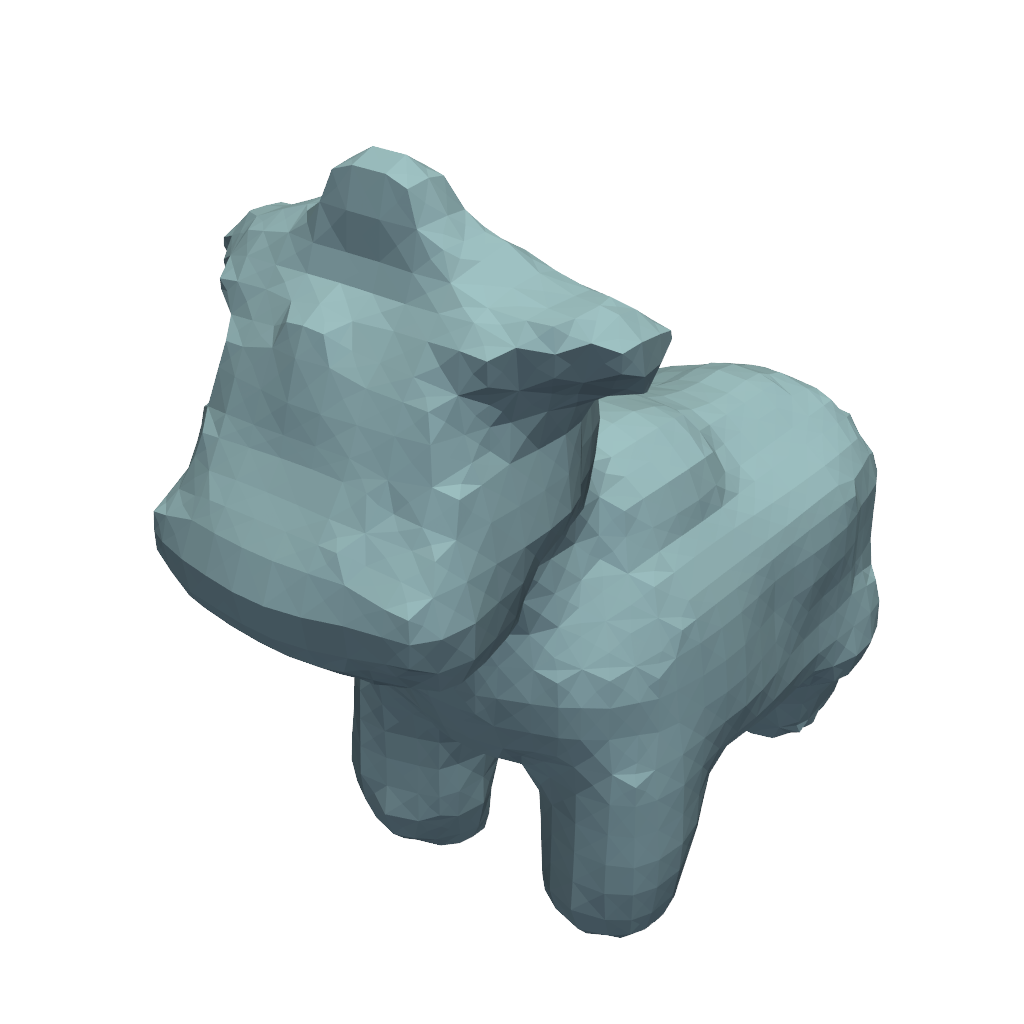} \\
    (a) Nvdiffrast & (b) Ours & (c) Nvdiffrast & (d) Ours
    \end{tabular}
    \end{spacing}
    \vspace{-7pt}
    \caption{Failure cases of Nvdiffrast. (a, c) Nvdiffrast optimization results. (b, d) Ours optimization results. In this task, our method can successfully recover a well-shaped object in most scenarios, whereas Nvdiffrast frequently fails, or producing distorted and uneven objects.
    }
    \label{fig:failure-cases-for-nvdiffrast}

    \vspace{-20pt}
\end{figure}

Beyond convergence speed, we observe that Nvdiffrast \cite{laine2020modular} frequently encounters catastrophic failures, leading to the inability to reconstruct valid meshes. For example, in our experiments on rotational recovery, Nvdiffrast failed to successfully complete the process (manifested as program crashes) in any of the 10 repeated attempts within 8 out of 25 test data cases, which even precluded a quantitative comparison with our method.
Furthermore, even in cases where Nvdiffrast manages to complete the reconstruction, its output quality often exhibits lower fidelity compared to ours, as illustrated in \cref{fig:failure-cases-for-nvdiffrast}. In stark contrast, our method consistently achieves successful, well-shaped, and high-fidelity reconstructions.
More analysis is provided in \Cref{sec:supp:nvdiffrast-gradients-concise}.


\vspace{-5pt}

\section{Discussion}
\label{sec:conclusion}

\vspace{-5pt}

\paragraph{Limitations}
Our method currently relies on known camera poses and motion parameters.
Additionally, our physical formation model assumes linear motion segments and a linear, noise-free photometric response.
While effective, these assumptions may deviate from in-the-wild scenarios characterized by complex non-linear motion, camera response functions (tone mapping), or sensor noise.
We provide a comprehensive discussion on these limitations and future works in \Cref{sec:supp:limitations}.

\vspace{-15pt}

\paragraph{Conclusion} 
In this paper, we propose a novel inverse rendering approach for 3D shape recovery from ultra-fast motion-blurred images. Our fast barycentric coordinate solver accelerates rendering while preserving accuracy, enabling efficient and fully differentiable shape reconstruction. Experimental results validate the effectiveness of our method on both synthetic and real-world data, advancing 3D reconstruction under ultra-fast motion blur.
\vspace{-5pt}

\section{Acknowledgements}

\vspace{-5pt}

This work is supported in part by grants from National Key Research and Development Program of China (Grant No. 2024YFB3309500), National Natural Science Foundation of China (Grant No. U23A20312, 62472257).

{
    \small
    \bibliographystyle{ieeenat_fullname}
    \bibliography{main}
}

\clearpage
\setcounter{page}{1}
\maketitlesupplementary

\appendix
\setcounter{table}{0}
\setcounter{figure}{0}
\setcounter{equation}{0}
\renewcommand{\thetable}{A\arabic{table}}
\renewcommand{\thefigure}{A\arabic{figure}}
\renewcommand{\theequation}{A\arabic{equation}}

\section{Appendix Overview}

In this appendix, we provide a comprehensive explanation of the technical details and additional experimental results of our work.

We start with the derivation of our method in \Cref{sec:supp:derivation}, followed by implementation details in \Cref{sec:supp:impl-details}. \Cref{sec:supp:segmentation-analysis} discusses segmentation analysis, and \Cref{sec:supp:visualization-details} covers visualization details. Scene settings are outlined in \Cref{sec:supp:scene-settings}. We then address limitations with a failure case of rotational optimization in \Cref{sec:supp:failure-mesh-rotational-optim} and detail all loss terms in \Cref{sec:supp:loss-terms}. \Cref{sec:supp:3d-losses} analyzes the suitability of 3D losses for evaluation. Finally, \Cref{sec:supp:hyperparameter} provides hyperparameter settings, \cref{sec:supp:further-analysis-of-baselines} provides further analysis of baselines, and \Cref{sec:supp:more-results} presents more results.

\section{Derivation of Our Method}
\label{sec:supp:derivation}

In this section, we show the derivation of our method.

\paragraph{Barycentric Coordinate Solver} We detail the derivation of $\vect{A}_1$, $\vect{A}_2$, $\vect{A}_3$, $a_1$, $a_2$, $a_3$.

First we introduce the definition of $\operatorname{adj}(\mat{F}_j(t))$ and $\det(\mat{F}_j(t))$. For a $3 \times 3$ matrix $\mat{A}$, given by:
\begin{equation}
    \mat{A} =
    \begin{bmatrix}
    a_{11} & a_{12} & a_{13} \\
    a_{21} & a_{22} & a_{23} \\
    a_{31} & a_{32} & a_{33}
    \end{bmatrix},
\end{equation}
its determinant $ \det(\mat{A}) $ is computed as:
\begin{equation}
    \det(\mat{A}) =
    a_{11} \begin{vmatrix} a_{22} & a_{23} \\ a_{32} & a_{33} \end{vmatrix}
    - a_{12} \begin{vmatrix} a_{21} & a_{23} \\ a_{31} & a_{33} \end{vmatrix}
    + a_{13} \begin{vmatrix} a_{21} & a_{22} \\ a_{31} & a_{32} \end{vmatrix}
\end{equation}
where each $ 2 \times 2 $ determinant (called a minor) is computed as:
\begin{equation}
    \begin{vmatrix} a & b \\ c & d \end{vmatrix} = ad - bc.
\end{equation}

For the matrix $\mat{A}$, its adjugate matrix $ \operatorname{adj}(\mat{A}) $ can be defined by:
\begin{equation}
    \operatorname{adj}(\mat{A}) =
    \begin{bmatrix}
    C_{11} & C_{21} & C_{31} \\
    C_{12} & C_{22} & C_{32} \\
    C_{13} & C_{23} & C_{33}
    \end{bmatrix},
\end{equation}
where $ C_{ij} $ are defined as:
\begin{equation*}
    C_{ij} = (-1)^{i+j} M_{ij}
\end{equation*} and $ M_{ij} $ is the determinant of the minor matrix obtained by deleting the $ i $-th row and $ j $-th column of $\mat{A}$.

If $\mat{A}$ is invertible (\ie, $ \det(\mat{A}) \neq 0 $), the inverse of $\mat{A}$ can be expressed in terms of its adjugate matrix:
\begin{equation}
\mat{A}^{-1} = \frac{\operatorname{adj}(\mat{A})}{\det(\mat{A})}.
\end{equation}

Now we consider derivation on a triangle. For a triangle matrix $\mat{F}_j(t)$ represented as:
\begin{equation}
    \mat{F}_j(t) = \begin{bmatrix} 
        x_0(t) & x_1(t) & x_2(t) \\ 
        y_0(t) & y_1(t) & y_2(t) \\ 
        1 & 1 & 1 
    \end{bmatrix},
    \label{eq:supp-1-1}
\end{equation}
where the vertex position $\vect{v}(t)$ can be defined by:
\begin{equation}
    \vect{v}(t) = (1 - t)\vect{v}(0) + t \vect{v}(1),
    \label{eq:supp-1-2}
\end{equation}
its determinant $\det(\mat{F}_j(t))$ is
\begin{equation}
\begin{aligned}
    \det(\mat{F}_j(t)) = & ~ x_0(t) \bigl(y_1(t) - y_2(t)\bigr) \\ & - x_1(t) \bigl(y_0(t) - y_2(t)\bigr) \\  & + x_2(t) \bigl(y_0(t) - y_1(t)\bigr),
\end{aligned}
    \label{eq:supp-1-3}
\end{equation}
and the adjugate matrix $\operatorname{adj}(\mat{A})$ is
\begin{footnotesize}
\begin{equation}
\begin{matrix}
 \operatorname{adj}(\mat{F}_j(t)) = \\
  \begin{bmatrix}
  y_1(t) - y_2(t) & x_2(t) - x_1(t) & x_1(t)y_2(t) - x_2(t)y_1(t) \\
  y_2(t) - y_0(t) & x_0(t) - x_2(t) & x_2(t)y_0(t) - x_0(t)y_2(t) \\
  y_0(t) - y_1(t) & x_1(t) - x_0(t) & x_0(t)y_1(t) - x_1(t)y_0(t)
  \end{bmatrix}.
\end{matrix}
    \label{eq:supp-1-4}
\end{equation}
\end{footnotesize}

Given a pixel $\vect{p}_i = \begin{bmatrix} u & v & 1 \end{bmatrix}^T$, we have
\begin{equation}
\begin{aligned}
    \vect{w}(t) &= \mat{F}_j(t)^{-1} \vect{p}_i =\frac{\operatorname{adj} (\mat{F}_j(t)) \times \vect{p}_i}{\det(\mat{F}_j(t))} \\
    &= \frac{\vect{A}_1 t^2 + \vect{A}_2 t + \vect{A}_3}{a_1 t^2 + a_2 t + a_3}
\end{aligned}
    \label{eq:supp-1-5}
\end{equation}

By \cref{eq:supp-1-1,eq:supp-1-2,eq:supp-1-3,eq:supp-1-4,eq:supp-1-5}, we can represent $\vect{A}_1$, $\vect{A_2}$, $\vect{A}_3$, $a_1$, $a_2$, $a_3$ using $x_i(0/1), y_i(0/1), u, v$:
\begin{footnotesize}
\begin{equation}
\begin{matrix}
     \vect{A}_1 =
      \begin{bmatrix}
          \begin{matrix}
          -(x_2(0) - x_2(1)) (y_1(0) - y_1(1)) \\
          + (x_1(0) - x_1(1))  (y_2(0) - y_2(1)) 
          \end{matrix}
        \\ 
        \\
          \begin{matrix}
          ((x_2(0) - x_2(1))  (y_0(0) - y_0(1)) \\
          - (x_0(0) - x_0(1))  (y_2(0) - y_2(1)))
          \end{matrix}
        \\ 
        \\
          \begin{matrix}
          (-((x_1(0) - x_1(1))  (y_0(0) - y_0(1))) \\
          + (x_0(0) - x_0(1))  (y_1(0) - y_1(1)))
          \end{matrix}
        \\ 
      \end{bmatrix},
\end{matrix}
\end{equation}
\end{footnotesize}
\begin{footnotesize}
\begin{equation}
\begin{matrix}
     \vect{A}_2 =
      \begin{bmatrix}
          \begin{matrix}
          u (-y_1(0) + y_2(0) + y_1(1) - y_2(1)) \\
          + v (x_1(0) - x_2(0) - x_1(1) + x_2(1)) \\
          + (y_2(0)  x_1(1) - y_1(0)  x_2(1) + \\ x_2(0) (2  y_1(0) - y_1(1)) + \\ x_1(0)  (-2  y_2(0) + y_2(1))) 
          \end{matrix}
        \\ 
        \\
          \begin{matrix}
          u (y_0(0) - y_2(0) - y_0(1) + y_2(1)) \\
          + v (-x_0(0) + x_2(0) + x_0(1) - x_2(1)) \\
          + (-(y_2(0)  x_0(1)) + y_0(0)  x_2(1) + \\
          x_2(0)  (-2  y_0(0) + y_0(1)) + \\
          
          x_0(0)  (2  y_2(0) - y_2(1)))
          \end{matrix}
        \\ 
        \\
          \begin{matrix}
          u (-y_0(0) + y_1(0) + y_0(1) - y_1(1)) \\
          + v (x_0(0) - x_1(0) - x_0(1) + x_1(1)) \\ 
          + (y_1(0)  x_0(1) - y_0(0)  x_1(1) + \\
          x_1(0)  (2  y_0(0) - y_0(1)) + \\
          x_0(0)  (-2  y_1(0) + y_1(1)))
          \end{matrix}
        \\ 
      \end{bmatrix},
\end{matrix}
\end{equation}
\end{footnotesize}
\begin{footnotesize}
\begin{equation}
\begin{matrix}
     \vect{A}_3 =
      \begin{bmatrix}
          \begin{matrix}
          u (y_1(0) - y_2(0)) + v (-x_1(0) + x_2(0)) \\ 
          + (-(x_2(0)  y_1(0)) + x_1(0)  y_2(0))
          \end{matrix}
        \\ 
        \\
          \begin{matrix}
          u (-y_0(0) + y_2(0)) + v (x_0(0) - x_2(0)) \\ 
          + (x_2(0)  y_0(0) - x_0(0)  y_2(0))
          \end{matrix}
        \\ 
        \\
          \begin{matrix}
          u (y_0(0) - y_1(0)) + v (-x_0(0) + x_1(0)) \\
          + (-(x_1(0)  y_0(0)) + x_0(0)  y_1(0))
          \end{matrix}
        \\ 
      \end{bmatrix},
\end{matrix}
\end{equation}
\end{footnotesize}
%
\begin{footnotesize}
\begin{equation}
\begin{matrix}
     a_1 = \\
      \begin{matrix}
        - (y_1(0)  x_0(1)) + y_2(0)  x_0(1) \\
        - y_2(0)  x_1(1) +  y_0(0)  (x_1(1) - x_2(1)) \\
        + y_1(0)  x_2(1) - x_1(1)  y_0(1) \\
        + x_2(1)  y_0(1) + x_0(1)  y_1(1) - x_2(1)  y_1(1) \\
        + x_2(0)  (y_0(0) - y_1(0) - y_0(1) + y_1(1)) \\
        + x_1(0)  (-y_0(0) + y_2(0) + y_0(1) - y_2(1)) \\
        - x_0(1)  y_2(1) + x_1(1)  y_2(1) \\
        + x_0(0)  (y_1(0) - y_2(0) - y_1(1) + y_2(1)),
      \end{matrix}
\end{matrix}
\end{equation}
\end{footnotesize}
%
\begin{footnotesize}
\begin{equation}
\begin{matrix}
     a_2 = \\
      \begin{matrix}
        y_1(0)  x_0(1) - y_2(0)  x_0(1) \\
        + y_2(0)  x_1(1) - y_1(0)  x_2(1) \\
        + y_0(0) (-x_1(1) + x_2(1)) \\
        + x_2(0) (-2 y_0(0) + 2 y_1(0) + y_0(1) - y_1(1)) \\
        + x_0(0) (-2  y_1(0) + 2  y_2(0) + y_1(1) - y_2(1)) \\
        + x_1(0) (2  y_0(0) - 2  y_2(0) - y_0(1) + y_2(1)),
      \end{matrix}
\end{matrix}
\end{equation}
\end{footnotesize}
%
\begin{footnotesize}
\begin{equation}
\begin{matrix}
     a_3 = \\
      \begin{matrix}
        x_2(0) (y_0(0) - y_1(0)) \\
        + x_0(0)  (y_1(0) - y_2(0)) \\
        + x_1(0)  (-y_0(0) + y_2(0)).
      \end{matrix}
\end{matrix}
\end{equation}
\end{footnotesize}

\paragraph{Euclidean Distance Approximation} 
\label{sec:supp:para:euc-distance-approx}

In this section, we detail the derivation of $\hat{w}^*$.

Given a triangle $\mat{F} = \begin{bmatrix} \vect{v_0} & \vect{v_1} & \vect{v}_2 \end{bmatrix} = \begin{bmatrix} x_0 & x_1 & x_2 \\ y_0 & y_1 & y_2 \\ 1 & 1 & 1\end{bmatrix}$ and pixel barycentric coordinates $\vect{w} = \begin{bmatrix} w_0 & w_1 & w_2 \end{bmatrix}$, we consider finding $\hat{w}^* = \begin{bmatrix} w_0^* & w_1^* & w_2^* \end{bmatrix}$ such that
\begin{equation}
    \vect{\hat{w}^*} = \mathop{\arg\min}_{\vect{\hat{w}} \in [0, 1]^3} \left| \left|  \mat{F} \vect{w} - \mat{F} \vect{\hat{w}}  \right| \right|_2^2.
\end{equation}

If the pixel is inside the triangle, it's obvious that $\vect{\hat{w}^*} = \vect{w}$, so we only consider the scenario where the pixel is outside the triangle. 

First we calculate the pixel position $\vect{p} = \begin{bmatrix} u & v & 1 \end{bmatrix}^T = \mat{F} \vect{w}$. If $\vect{p}$ is outside the triangle $\mat{F}$, the closest point $\vect{p}^*$ must lie on one of the triangle’s edges. Therefore, we need to compute the closest point from $\vect{p}$ to each of the 3 edges of the triangle and select the one with the minimum distance. 

For each edge $\vect{v}_i \vect{v}_j$, we first compute the parameter $t$ such that the projection (closest) point $\vect{p}^{'} = \vect{v}_i + t (\vect{v}_j - \vect{v}_i)$. We have
\begin{equation}
    t = \frac{(u - x_i)(x_j - x_i) + (v - y_i)(y_j - y_i)}{(x_j - x_i)^2 + (y_j - y_i)^2}.
\end{equation}

If $0 \leq t \leq 1$, the projection point lies on the edge, and the barycentric coordinates of $\vect{p}^{'}$ can be represented as $\vect{w}^{'} = \begin{bmatrix} w_0^{'} & w_1^{'} & w_2^{'} \end{bmatrix}$, where $w_i^{'} = 1 - t^{'}, w_j^{'} = t, \text{the rest one} = 0$. If $t < 0$, then $\vect{p}^{'} = \vect{v}_i$, $w_i^{'} = 1, \text{the rest} = 0$. If $t > 1$, then $\vect{p}^{'} = \vect{v}_j$, $w_j^{'} = 1, \text{the rest} = 0$.

Perform these computations for the three edges $\vect{v}_0 \vect{v}_1$, $\vect{v}_1 \vect{v}_2$, $\vect{v}_2 \vect{v}_0$, then choose the $\vect{p}^{'}$ with the smallest distance. Its barycentric coordinates $\vect{w}^{'}$ are the desired solution $\hat{w}^*$.

\section{Implementation Details}
\label{sec:supp:impl-details}

In this section, we present implementation details of our method. Our codebase is built on SoftRas. 
However, we follow DIB-R \cite{DIBR19} and separately compute the foreground and background pixels. 
Moreover, in the original Softras implementation, the probability map $A_j^i$ is defined as: 
\begin{equation}
    A_i^j = \operatorname{sigmoid} \left(-\frac{d \left(\vect{p}_i, \mat{F}_j \right)}{\delta} \right).
\end{equation}
We change it to exponential function for smoother gradients~\cite{DIBR19}:
\begin{equation}
    A_i^j = \operatorname{exp} \left(-\frac{d \left(\vect{p}_i, \mat{F}_j \right)}{\delta} \right).
\end{equation}

In addition, we find that enabling \emph{Aggregate Function} in SoftRas \cite{liu2019softras} results in a total reconstruction failure in the optimization task, so we disable it in all of our experiments.

\section{Segmentation Analysis}
\label{sec:supp:segmentation-analysis}

In the main paper, we decompose the entire rotation into 12 segments. In this section, we will illustrate the quality of forward rendered images and backward gradients with respect to the number of segments used.

\begin{figure}
    \centering
    \begin{spacing}{1}
    \setlength\tabcolsep{0pt}
    \begin{tabular}{ccccccc} 
    & & 6 {\small{Segments}} & 12 {\small{Seg.}}  & 24 {\small{Seg.}} & 48 {\small{Seg.}} \\
    \multirow{4}{*}{\rotatebox[origin=c]{90}{\textbf{Forward}} \hspace{1pt}} 
    & 
    \rotatebox[origin=c]{90}{SoftRas} \hspace{1pt} &
            \imagecell[0.22]{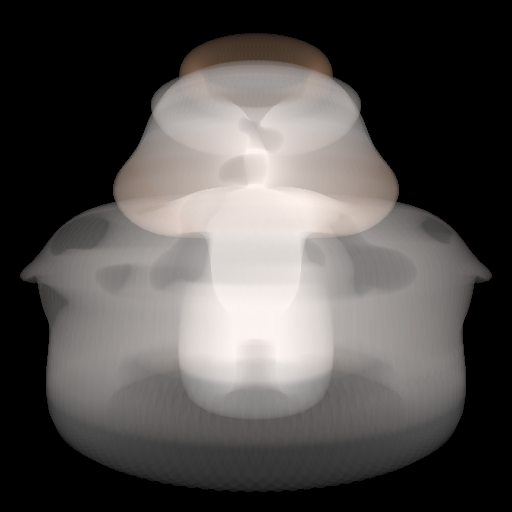} & 
            \imagecell[0.22]{figures/forward/softras/spot/rotation/12x20} &
            \imagecell[0.22]{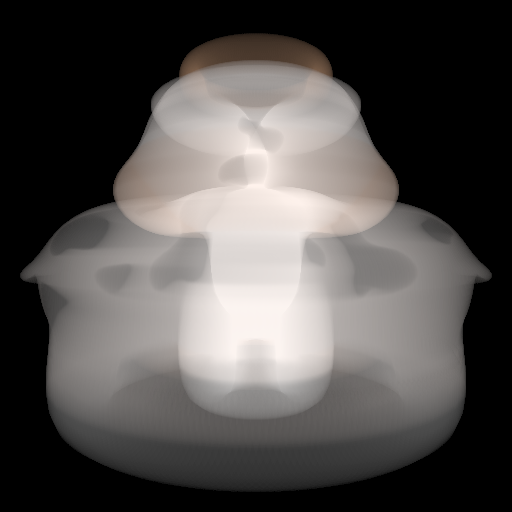} & 
            \imagecell[0.22]{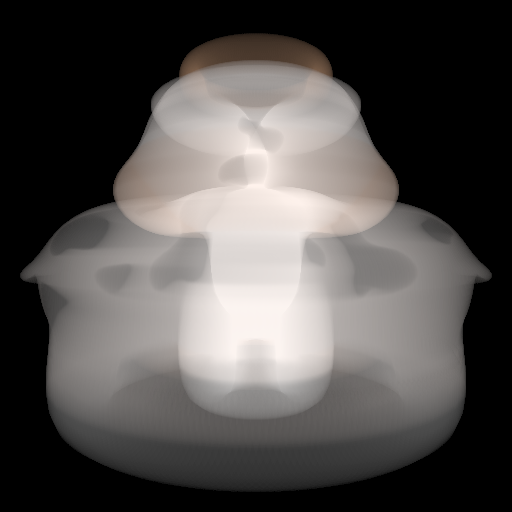} \vspace{3pt}
    \\ 
    &
    \rotatebox[origin=c]{90}{Ours} \hspace{1pt} &
            \imagecell[0.22]{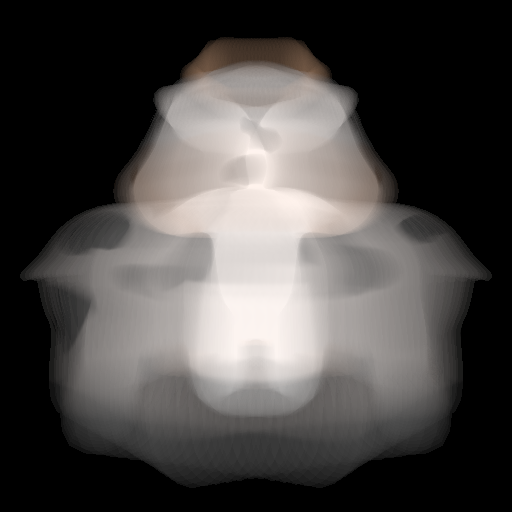} & 
            \imagecell[0.22]{figures/forward/ours/spot/rotation/12x20} &
            \imagecell[0.22]{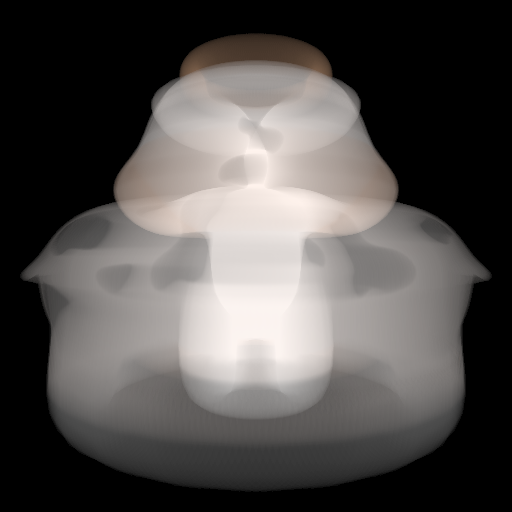} & 
            \imagecell[0.22]{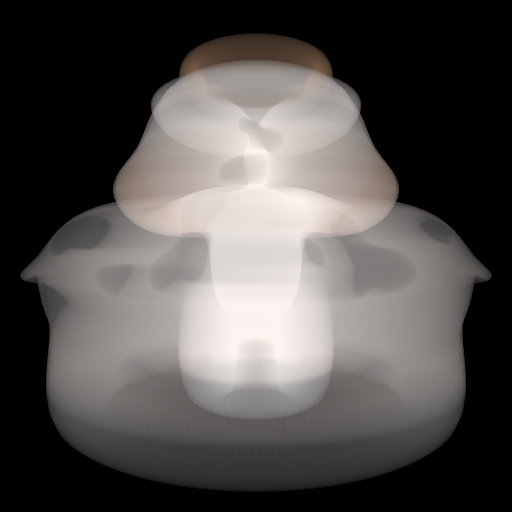} \vspace{3pt}
    \\ 

    \multirow{4}{*}{\rotatebox[origin=c]{90}{\textbf{Gradient}} \hspace{1pt}} 
    &
    \rotatebox[origin=c]{90}{SoftRas} \hspace{1pt} &
            \imagecell[0.22]{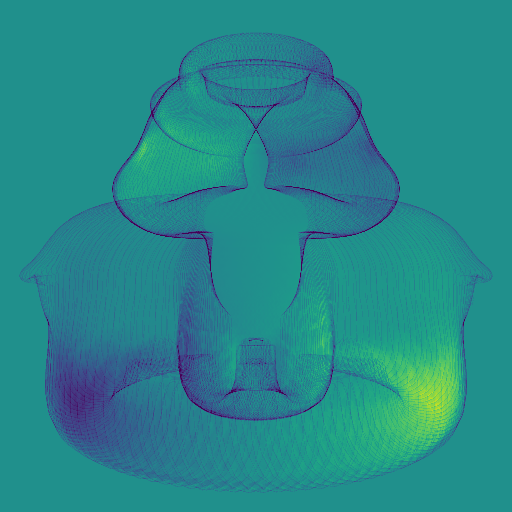} & 
            \imagecell[0.22]{figures/backward/softras/spot/rotation/12x20} &
            \imagecell[0.22]{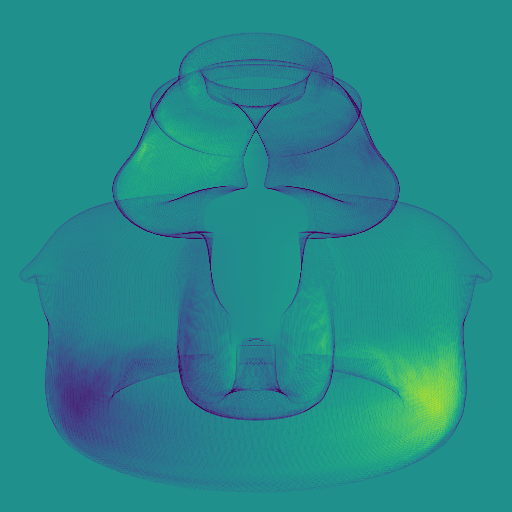} & 
            \imagecell[0.22]{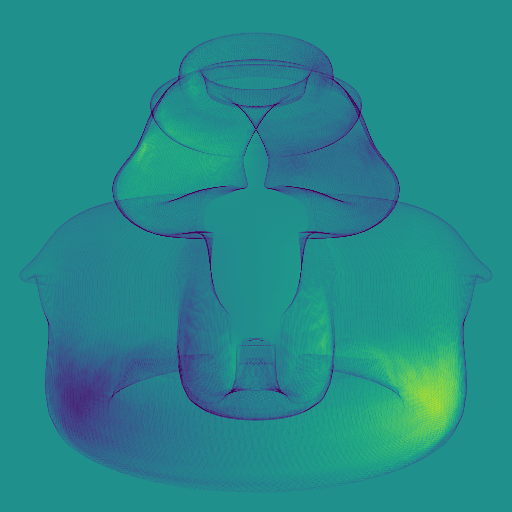} \vspace{3pt}
    \\
    &
    \rotatebox[origin=c]{90}{Ours} \hspace{1pt} &
            \imagecell[0.22]{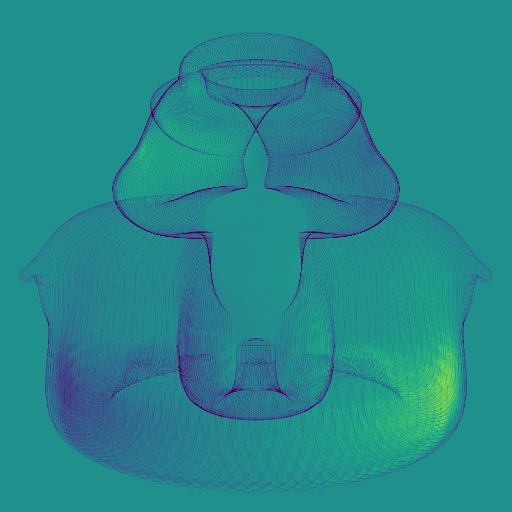} & 
            \imagecell[0.22]{figures/backward/ours/spot/rotation/12x20} &
            \imagecell[0.22]{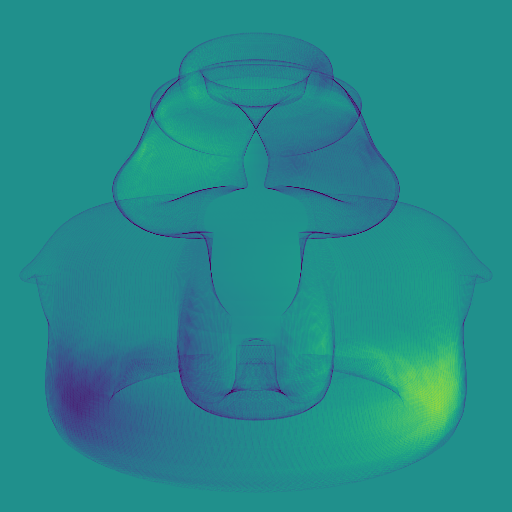} & 
            \imagecell[0.22]{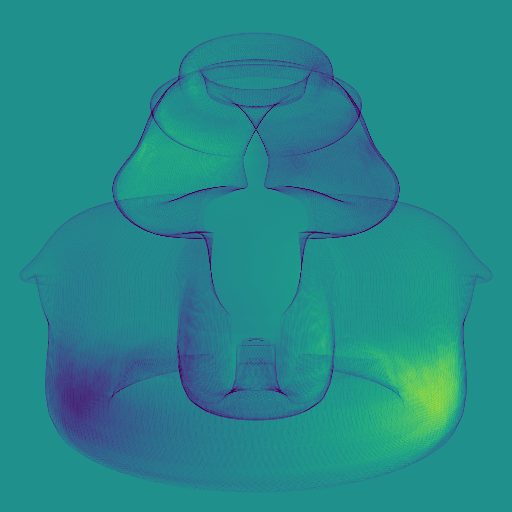} 
    \\
    
    \end{tabular}
    
    \end{spacing}
    
    \caption{Impact of Segment Count on Rendering and Gradient Quality.
    Forward rendering and backward gradient visualization demonstrating the effect of different segment counts on motion-blur synthesis.
    As illustrated, using fewer than 12 segments (e.g., 6 segments) introduces severe artifacts in forward rendering and compromises gradient quality. 
    Increasing the segment count to 12 significantly improves both rendering smoothness and gradient accuracy.}
    \label{fig-supp:image-analysis}
\end{figure}

Results are illustrated in \cref{fig-supp:image-analysis}. If using fewer than 12 segments (e.g., 6 segments) leads to severe artifacts in the forward rendering. Conversely, employing more than 12 segments increases the computational cost significantly, with only marginal improvement in rendering quality.

Therefore, as a trade-off between rendering quality and computational efficiency, 12 segments are chosen in our experiments.

\section{Visualization Details}
\label{sec:supp:visualization-details}

In this section, we provide detailed explanations of the rendering process for the images presented in \cref{fig:image-analysis}.

All forward images are rendered using the same camera parameters and blur settings (\ie, translation or rotation speed) as those used in our main experiments.
For translational motion, we do not decompose the motion into segments. For rotational motion, the entire rotation circle is always decomposed into 12 segments.
Consequently, for rotational motion blurred with a total of 12, 60 or 240 samples, these are respectively distributed as 1, 5 and 20 samples per segment, given our decomposition into 12 segments.

For gradient images, we compute gradients with respect to the X-positions of all vertices. 
After obtaining these per-vertex gradient scalars, we then render these scalars into single-channel grayscale images, which are subsequently color-mapped using the Viridis color-map for visualization. 

All images are rendered at a resolution of $512 \times 512$ pixels.

\section{Failures of Mesh in Rotational Optimization}
\label{sec:supp:failure-mesh-rotational-optim}

\begin{figure}
    \centering
    \begin{spacing}{1.0}
    \setlength\tabcolsep{0pt}
    \begin{tabular}{cccc}
    \imagecell[0.23]{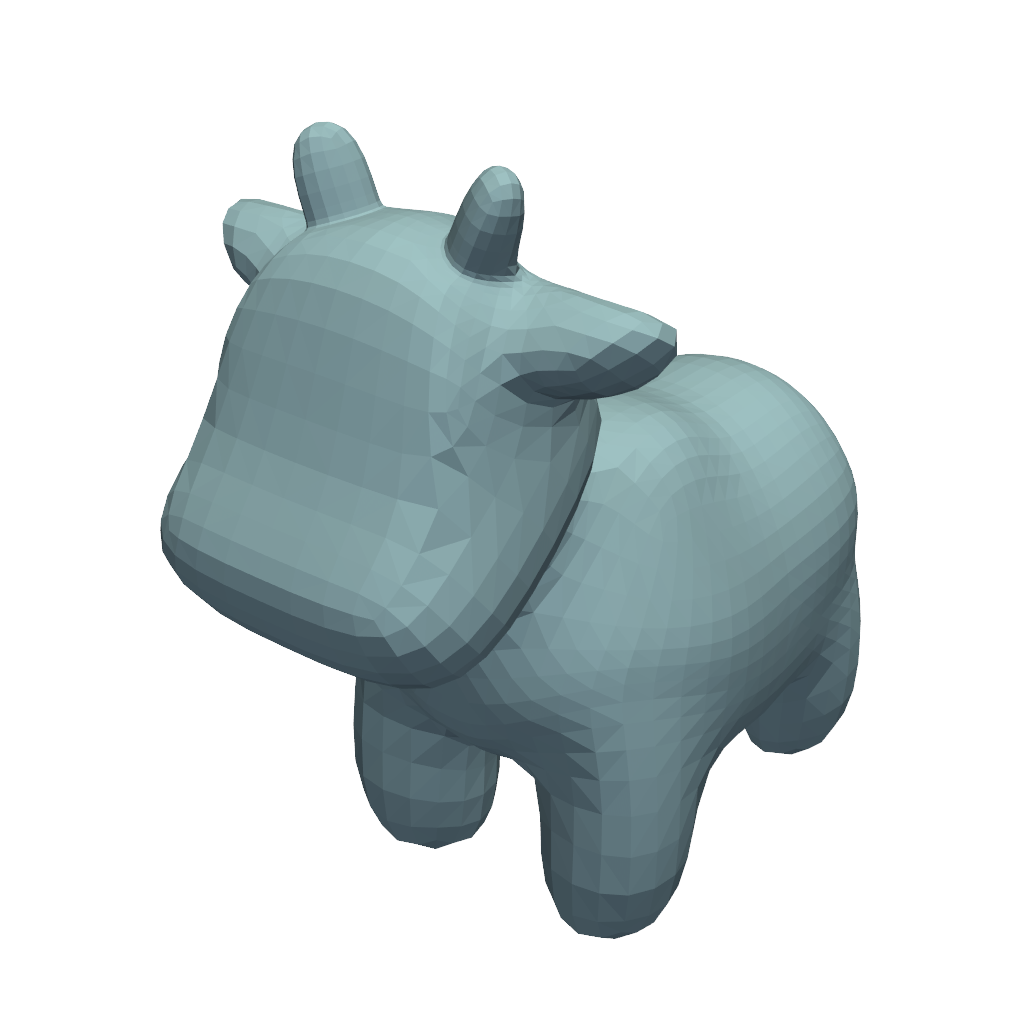} & 
    \imagecell[0.23]{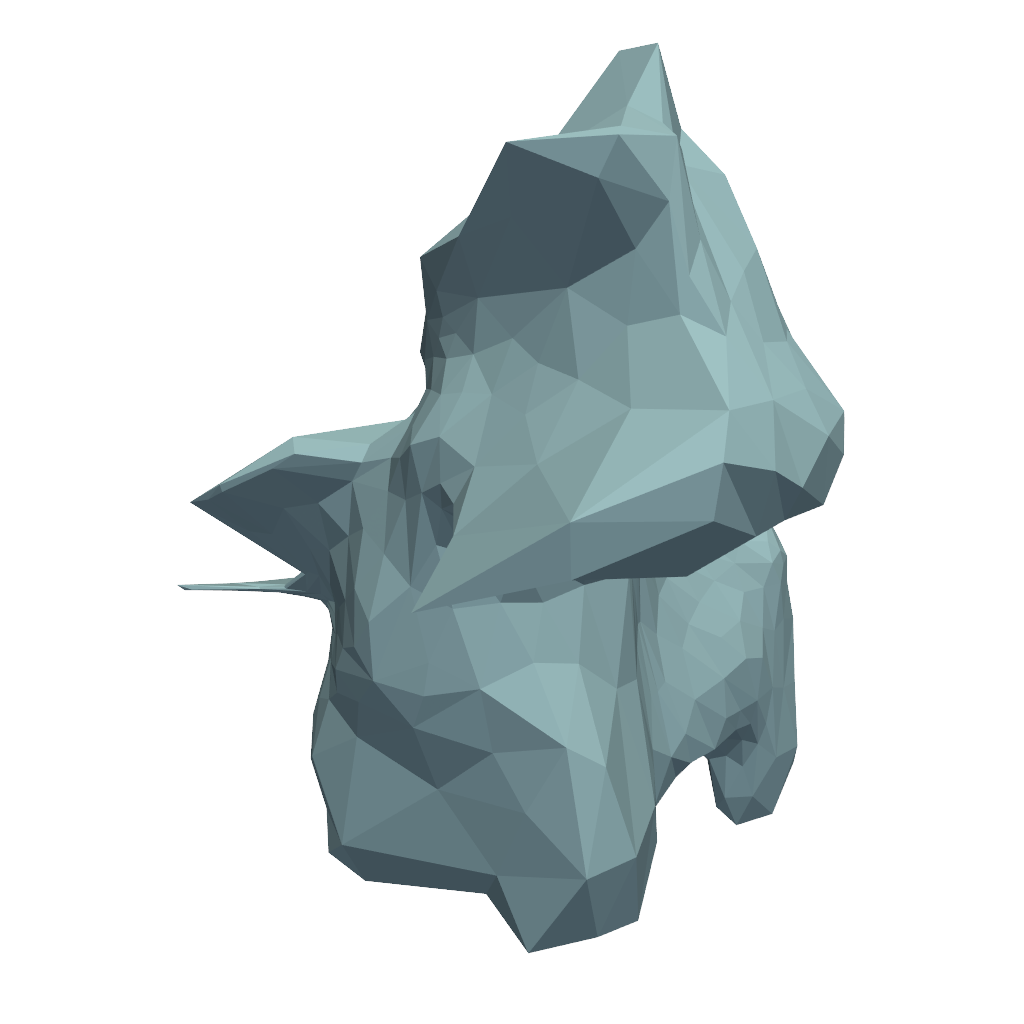} & 
    \imagecell[0.23]{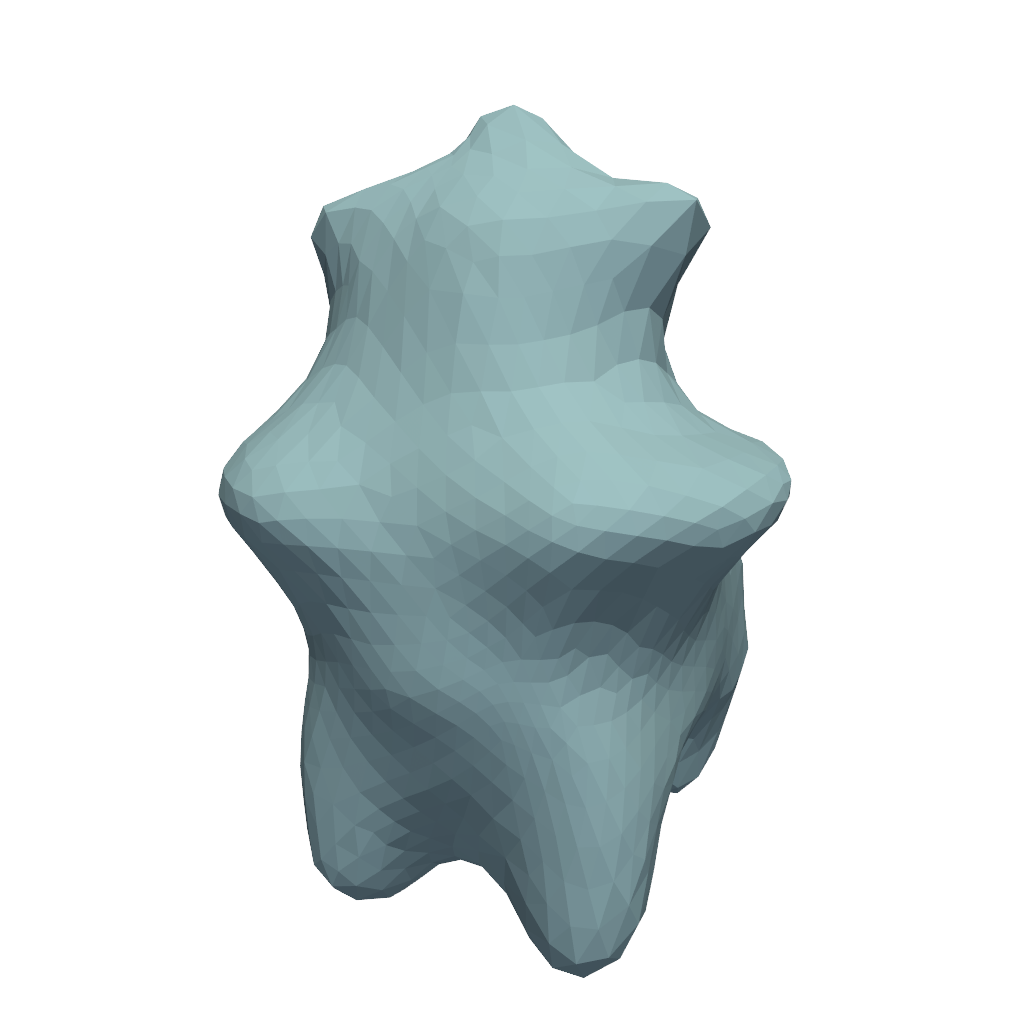} &
    \imagecell[0.23]{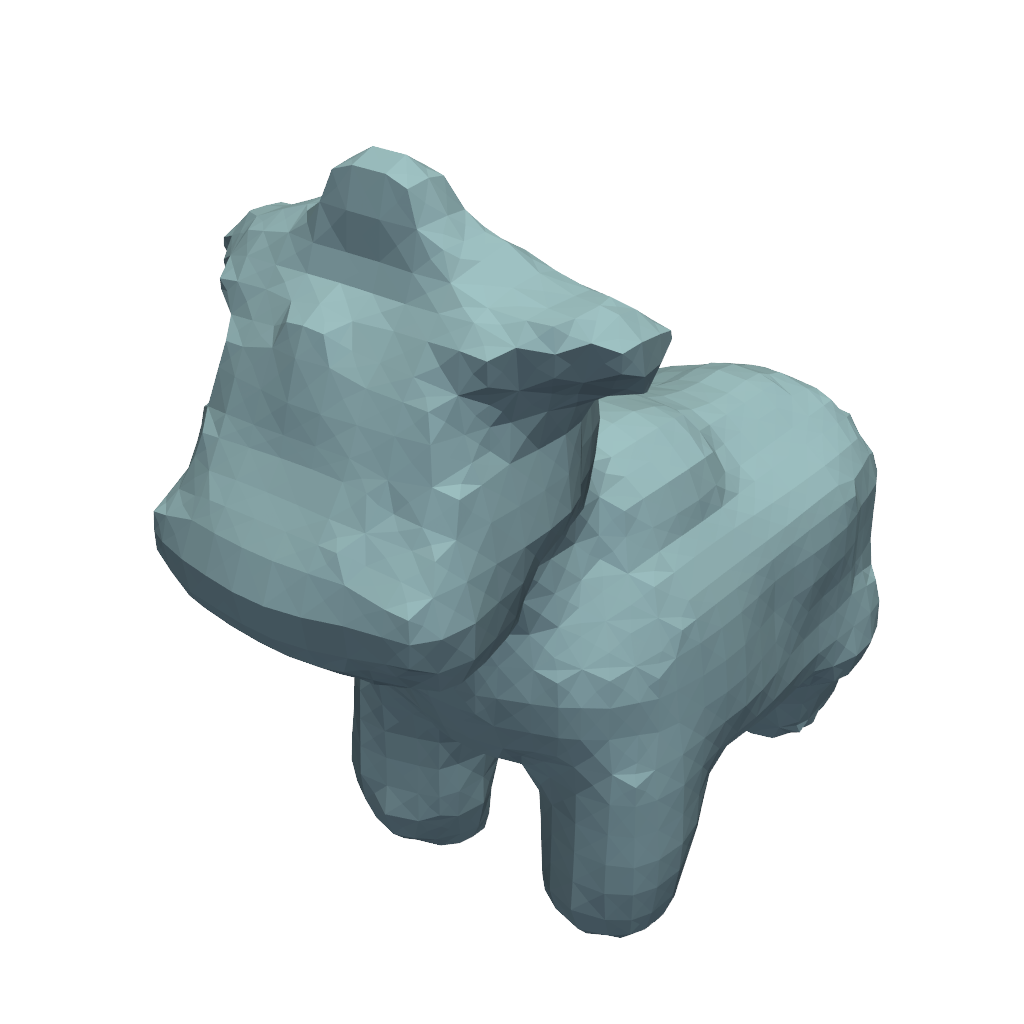} \\
    (a) G.T. & (b) $\begin{matrix}
        \text{Mesh} \\
        \text{Low Smo.}
    \end{matrix}$ & (c) $\begin{matrix}
        \text{Mesh} \\
        \text{High Smo.}
    \end{matrix}$ & (d) $\begin{matrix}
        \text{SDF}
    \end{matrix}$
    \end{tabular}
    \end{spacing}
    \caption{The results of rotational optimization. Mesh representation fails to recover a well-shaped Spot cow, no matter how smooth it is. Instead, SDF representation recovers a significantly better Spot cow.}
    \label{fig-supp:failure-mesh-rotational-optim}
\end{figure}

In rotational optimization, we observe that directly optimizing mesh vertices fails to recover well-shaped objects. One such failure case is illustrated in \cref{fig-supp:failure-mesh-rotational-optim}. 
The mesh representation consistently fails to recover a well-shaped object, even when incorporating smoothing regularization during optimization. 
In contrast, the results obtained with the SDF representation are substantially superior.

\section{Details of Loss Terms}
\label{sec:supp:loss-terms}

In this section, we provide detailed definitions of additional loss terms not explicitly covered in our main paper.

\paragraph{Laplacian Loss} 
Our definition of Laplacian loss follows \cite{liu2019softras}. For each vertex $v$, let $\mathcal{N}(v)$ be the set of adjacent vertices of $v$. The Laplacian loss is then defined as:
\begin{equation}
    \mathcal{L}_\text{L} = \sum_v \left\| \delta_v - \frac{1}{|\mathcal{N}(v)|} \sum_{v' \in \mathcal{N}(v)} \delta_{v^{'}} \right\|^2.
\end{equation}
where $\delta_v$ denotes the predicted movement of vector $v$. This Laplacian loss encourages adjacent vertices to move consistently, thereby promoting mesh deformation smoothness.

\paragraph{Smoothness Loss} 
Our definition of Smoothness loss is the same as \cite{liu2019softras}. For all two neighboring faces sharing the edge $e_i$, let $\theta_i$ be the dihedral  angle between the two faces. We have \begin{equation}
    \mathcal{L}_\text{s} = \sum_{e_i} \left(\cos (\theta_i) + 1 \right)^2.
\end{equation}

This smoothness loss encourages adjacent faces to have similar normal directions, thereby penalizing sharp edges.

\paragraph{Regularization Loss in FlexiCubes}
We incorporate the regularization loss provided in FlexiCubes \cite{shen2023flexicubes}. It is defined as: \begin{equation}
    \mathcal{L}_\text{reg} = \lambda_\text{dev} \mathcal{L}_\text{dev} + \lambda_\text{sign} \mathcal{L}_\text{sign}.
\end{equation}

For $\mathcal{L}_\text{dev}$, it is defined as: \begin{equation}
    \mathcal{L}_\text{dev} = \sum _{v \in V} \text{MAD} \left[ \left\{ 
        |v - u_e|_2 : u_e \in \mathcal{N}(v)
    \right\} \right],
\end{equation}
where $V$ denotes the set of voxel grid vertices, $| \cdot |2$ denotes Euclidean distance, $\text{MAD}(Y) = \frac{1}{|Y|} \sum{y \in Y} |y - \text{mean}(Y)|$ is the Mean Absolute Deviation, and $\mathcal{N}(v)$ denotes the set of adjacent vertices of $v$. This term penalizes the variability of distances between a vertex $v$ and its neighbors $u_e \in \mathcal{N}(v)$.

For $\mathcal{L}_\text{sign}$, it is defined as: \begin{equation}
    \mathcal{L}_\text{sign} = \sum_{(s_a, s_b) \in \mathcal{E}_g} H \left( \sigma(s_a), \text{sign} (s_b) \right),
\end{equation}
where $\mathcal{E}_g$ denotes the set of all edges $(a, b)$ where the scalar function values $(s_a, s_b)$ at grid vertices $a, b$ have differing signs (\ie, cross the zero-level set). $H$ and $\sigma$ denote the cross-entropy and sigmoid functions, respectively. This term discourages the appearance of spurious geometrical structures or internal cavities in regions where explicit shape supervision is absent.

We use the same weight parameters $\lambda_\text{dev}, \lambda_\text{sign}$ as specified in \cite{shen2023flexicubes}.

\paragraph{Regularization Loss in Neural-Singular-Hessian}
We use the regularization loss provided in Neural-Singular-Hessian \cite{wang2023neural}. It is defined as:
\begin{equation}
    \mathcal{L}_\text{crit} = \lambda_\text{Eikonal} \mathcal{L}_\text{Eikonal} + \lambda_\text{singularH} \mathcal{L}_\text{singularH}.
\end{equation}

The Eikonal loss $\mathcal{L}_\text{Eikonal}$ is defined as:
\begin{equation}
    \mathcal{L}_\text{Eikonal} = \int_\mathcal{P} || (||\nabla f(x) ||_2 - 1) ||_1 \dif x,
\end{equation}
where $f(\cdot)$ denotes the SDF function and $\mathcal{P}$ denotes the set of sampling points. The Eikonal loss encourages the gradient magnitude of the SDF field to be 1, which is crucial for maintaining global smoothness and a valid SDF property.

The singular Hessian loss $\mathcal{L}_\text{singularH}$ is defined as:
\begin{equation}
    \mathcal{L}_\text{singularH} = \int_{\mathcal{P}_\text{near}} || \det(\mat{H}_f(x)) ||_1 \dif x,
\end{equation}
where $f(\cdot)$ denotes the SDF function, $\mathcal{P}_\text{near}$ denotes the set of sampling points located near the zero-level set (surface), and $\det(\mat{H}_f(x))$ signifies the determinant of the Hessian matrix $\mat{H}f(x)$. The Hessian matrix is defined as the Jacobian of the gradient of $f$: \begin{equation}
    \mat{H}_f(x) = \begin{bmatrix}
        f_{xx}(x) & f_{xy}(x) & f_{xz}(x) \\
        f_{yx}(x) & f_{yy}(x) & f_{yz}(x) \\
        f_{zx}(x) & f_{zy}(x) & f_{zz}(x)
    \end{bmatrix}.
\end{equation}

We set the initial weighting parameters as $\lambda_\text{Eikonal} = \frac{50}{53}$ and $\lambda_\text{singularH} = \frac{3}{53}$. The same decay policy as described in \cite{wang2023neural} is adopted.

\section{Analysis of 3D Losses in Rotational Optimization}
\label{sec:supp:3d-losses}

In rotational optimization, we did not employ 3D losses for quantitative evaluation of reconstructed object shapes. The rationale behind this decision is detailed in this section.

\begin{figure}
    \centering
    \begin{spacing}{1.0}
    \setlength\tabcolsep{0pt}
    \begin{tabular}{cccc}
    \imagecell[0.23]{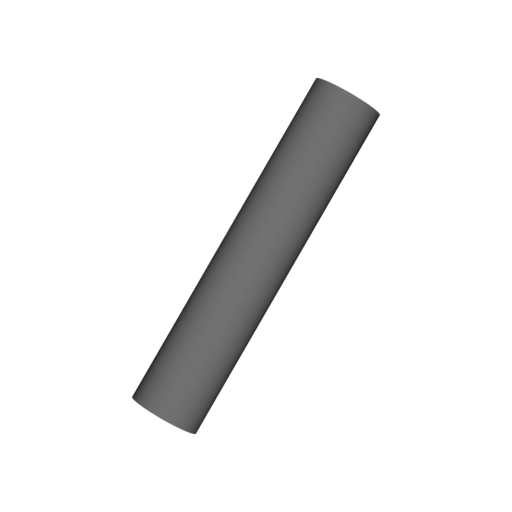} & 
    \imagecell[0.23]{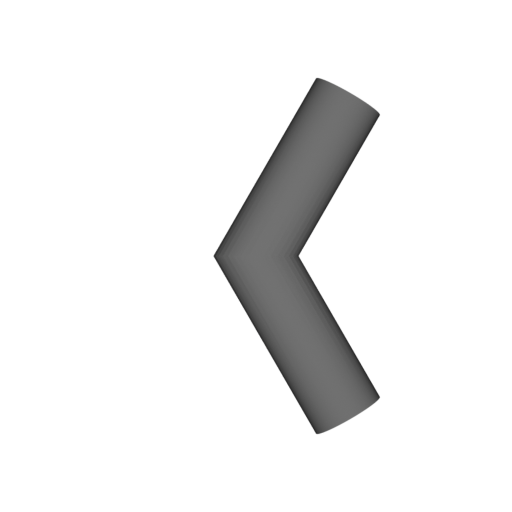} & 
    \imagecell[0.23]{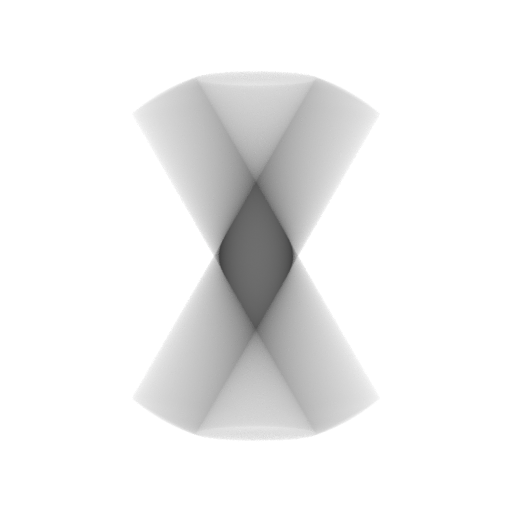} &
    \imagecell[0.23]{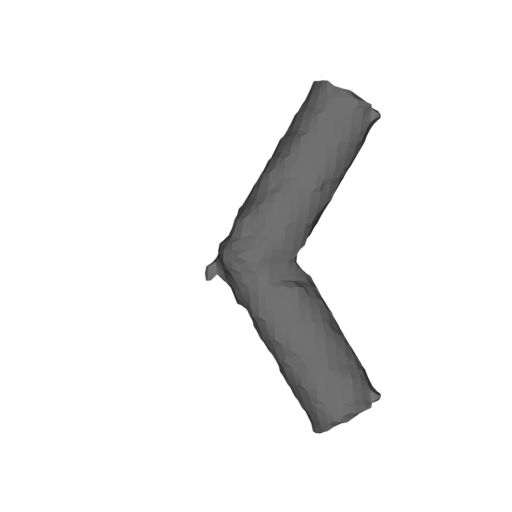} \\
    (a) Cylinder & (b) $\begin{matrix}
        \text{Twisted} \\
        \text{Cylinder}
    \end{matrix}$ & (c) $\begin{matrix}
        \text{Blur} \\
        \text{Image}
    \end{matrix}$ & (d) $\begin{matrix}
        \text{Ours} \\
        \text{Result}
    \end{matrix}$
    \end{tabular}
    \end{spacing}
    \caption{A cylinder and a twisted cylinder. They share a same rotational blurred image. Given (c) as input, our optimization result is (d).}
    \label{fig-supp:reasons-not-evaluating-3d-losses}
\end{figure}

In the rotational recovery task, it is common for multiple distinct 3D objects to produce rotational motion-blurred images that are indistinguishable from the input blurred image. 
An illustrative example is provided in \cref{fig-supp:reasons-not-evaluating-3d-losses}, where the rotational motion-blurred images of both objects in \cref{fig-supp:reasons-not-evaluating-3d-losses} (a, b) result in the same blurred image shown in \cref{fig-supp:reasons-not-evaluating-3d-losses} (c).

As demonstrated in \cref{fig-supp:reasons-not-evaluating-3d-losses} (a, b), the geometric discrepancies among feasible objects can be substantial, and it is unreasonable to designate any one of these feasible objects as the ground truth. Consequently, evaluating the results using 3D losses or static image losses is not suitable. 

To the best of our knowledge, the most effective evaluation for this task is to compute the differences between the rotational motion-blurred images, which is adopted in our main paper.

\section{Scene Settings}
\label{sec:supp:scene-settings}

In this section, we detail the specific configurations for our scenes, covering object initialization, motion parameters, and camera extrinsic and intrinsic properties.

\subsection{Object Initialization}

All 3D objects utilized in our experiments (\eg, for gradient visualization and optimization evaluation) undergo a two-step initialization process. 
First, Each object is uniformly scaled such that the maximum Euclidean norm of any vertex does not exceed 1.
Subsequently, each object is rotated around its local X-axis by a random angle uniformly sampled from the range $[-90^\circ, 90^\circ]$.

\subsection{Motion Parameters}

\paragraph{Translation}

For all translational motion, objects undergo a linear translation along the X-axis. The position $P(t) = (x(t), y(t), z(t))$ of a vertex that was initially at $P_0 = (x_0, y_0, z_0)$ is defined by:
\begin{equation}
\begin{cases}
x(t) = x_0 + (0.5 - t) \\
y(t) = y_0 \\
z(t) = z_0
\end{cases}
\quad \text{for } t \in [0, 1]
\end{equation}
This leads to the object translating linearly from an X-coordinate of $x_0 + 0.5$ at $t=0$ to $x_0 - 0.5$ at $t=1$.

\paragraph{Rotation}

For rotational motion, objects are rotated around the Y-axis. The angular displacement is $\theta(t) = 2\pi t$, where $t \in [0, 1]$. The position $P(t) = (x(t), y(t), z(t))$ of a vertex that was initially at $P_0 = (x_0, y_0, z_0)$ is defined by:
\begin{equation}
P(t) = \mathbf{R}_y(2\pi t) P_0,
\end{equation}
where $\mathbf{R}_y(\theta)$ is the 3D rotation matrix around the Y-axis by an angle $\theta$:
\begin{equation}
\mathbf{R}_y(\theta) = \begin{pmatrix}
\cos\theta & 0 & -\sin\theta \\
0 & 1 & 0 \\
\sin\theta & 0 & \cos\theta
\end{pmatrix}
\end{equation}

\subsection{Camera}

Following SoftRas, our camera setup employs a standard perspective model. The camera's eye point $\mathbf{E}=(E_x, E_y, E_z)$, from which it observes the scene origin $(0,0,0)$, is defined by spherical coordinates: a radial distance $d$, an elevation angle $\phi$, and an azimuth angle $\theta$. The conversion to Cartesian coordinates is given by:
\begin{equation}
\begin{cases}
E_x = d \cos(\phi) \cos(\theta) \\
E_y = d \cos(\phi) \sin(\theta) \\
E_z = d \sin(\phi)
\end{cases}
\end{equation}

For all experiments, $d=2.232$, and $\phi \in \{-60^\circ, $ $ -30^\circ, 0^\circ, 30^\circ, 60^\circ\}$. The azimuth angle $\theta$ varies with the motion type: (1) For translational motion, $\theta \in \{-315^\circ, -270^\circ, -225^\circ $ $, -180^\circ, -135^\circ, -90^\circ, -45^\circ, 0^\circ\}$; (2) For rotational motion, $\theta = 0^\circ$.

The camera's intrinsic parameters define a perspective projection with a fixed half-angular field of view $\alpha=30^\circ$. A 3D point $P=(x,y,z)$ in camera coordinates is projected to an image point $P_p=(x_p, y_p)$ as:
\begin{equation}
x_p = \frac{x}{z \cdot \tan(\alpha)} \quad \text{and} \quad y_p = \frac{y}{z \cdot \tan(\alpha)}
\end{equation}

\section{Hyperparameter Settings}
\label{sec:supp:hyperparameter}

In this section, we detail the hyperparameter settings used in our experiments.

\subsection{Overall Settings}

Following \cite{liu2019softras}, we set $\delta = 1 \times 10^{-4}$ in the probability function. Unless otherwise stated, we randomly select 25 objects from ShapeNet \cite{chang2015shapenet} for evaluation. The ADAM optimizer \cite{kingma2014adam} is employed for optimization. Each image is rendered at a resolution of $128 \times 128$ pixels. All experiments are conducted on a single NVIDIA RTX 4090 GPU with 24GB of memory.

\subsection{Translational Optimization}

In this experiment, each object is rendered from 40 different viewpoints. We set $\lambda_\text{S} = 3 \times 10^{-2}$, $\lambda_\text{L} = 3 \times 10^{-4}$, and $\alpha = 0.01$, $\beta_1 = 0.5$, $\beta_2 = 0.99$ (following \cite{liu2019softras}) for the ADAM optimizer. 
The batch size for input views is set to 16, and each object is optimized for 1000 iterations. 
A sphere consisting of 1352 vertices and 2700 faces is utilized as a template mesh for deformation.
We use the same method as the official SoftRas implementation for mesh texturing.

\subsection{Rotational Optimization}

In this experiment, each object is rendered from 5 viewpoints with varying elevations. The ADAM optimizer is configured with $\alpha = 5 \times 10^{-4}$, $\beta_1 = 0.9$, and $\beta_2 = 0.999$. The batch size for input views is set to 5, and each object is optimized for 1000 iterations. We set the initial $\lambda_\text{crit} = 3 \times 10^{-3}, \lambda_\text{reg} = 1$. The voxel-grid resolution in FlexiCubes \cite{shen2023flexicubes} is set to 32. In our approach, we decompose the entire rotation into 12 segments, all of which are uniformly sampled.

We first pretrain the SDF field on an inclined ellipsoid (defined by $4x^2 + 2.5y^2 + 2.5z^2 - 3yz = 1$) for 500 iterations, followed by an additional 1000 iterations of optimization.

\section{Further Analysis of Baselines}
\label{sec:supp:further-analysis-of-baselines}

\subsection{Shape From Blur \cite{rozumnyi2021shape}}
\label{sec:supp:sfb-analysis}

In this section, we provide more details and analysis of our comparative experiments against Shape from Blur (SFB)~\cite{rozumnyi2021shape}. 

\paragraph{Experimental Setup Adaptations for Comparison}
The problem formulation and experimental setup of SFB differ from our inverse rendering approach.
SFB is designed to recover 3D shape and motion parameters directly from a single RGB blurred image, leveraging a pre-trained neural network (DeFMO, \cite{rozumnyi2021defmo}) for intermediate guidance.
Specifically, SFB takes a single RGB image and an RGB background as input, and does not require or utilize explicit object motion information (\eg, translation or rotation velocities). Its core optimization loop involves:
\begin{enumerate}
    \item Using DeFMO to predict instance-level static masks (silhouettes) for multiple intermediate timestamps from the input. These masks represent the underlying static appearance of the object at various points along its motion path.
    \item Optimizing for mesh deformation (starting from a template mesh) and motion parameters (translation $t$, $\Delta t$, rotation $r$, $\Delta r$) through a single-view, differentiable rendering pipeline.
    \item In each optimization iteration, it renders RGB images and silhouette masks for multiple timestamps.
    \item The rendered silhouettes are compared against the static masks predicted by DeFMO.
    \item All rendered RGB images, masked by their silhouettes and composited with the input background, are averaged to form a synthetic motion-blurred image. This image is then compared against the input RGB image. These losses drive the backward propagation and optimization.
\end{enumerate}

\hspace{-14pt} In contrast, our method operates on multi-view RGB images with their corresponding non-binary transparency masks (alpha channels). 
In addition, our method requires and utilizes object motion information (\eg, trajectory, velocities) as input. 
During optimization, we render multi-view motion-blurred RGBA images by accumulating contributions from the object along its known motion path, which are then compared against the input images for gradient computation.

Despite these fundamental differences, SFB remains the most relevant benchmark due to the severe scarcity of alternative methods tackling 3D shape recovery from motion blur. 
To enable a best-effort comparison, we adapted our data for SFB. 
Specifically, for each input to SFB, we generate a single RGB image by masking our RGB images with corresponding transparency masks and compositing them onto a plain black background, to minimize the influence of the background to the greatest extent possible. 
This ensures SFB receives input that best aligns with its expected format (RGB image + background) while making our data compatible. 
We kept SFB's camera parameters consistent with those used in our setup and made no other modifications to SFB's internal configurations or parameters, aiming for the most straightforward comparison.

\paragraph{Why SFB Performs Not So Well in These Extreme Motion Scenarios}

As demonstrated in the main paper (\cref{tab:comparison-sfb}), our method significantly outperforms SFB for ultra-fast motion blur reconstruction. 
This disparity, particularly in extreme motion scenarios, primarily stems from a limitation in SFB's pipeline: its heavy reliance on the DeFMO \cite{rozumnyi2021defmo} neural network for deriving intermediate static masks.

\begin{figure*}[htbp]
    \centering
    \begin{spacing}{1}
    \setlength\tabcolsep{1pt}
    \begin{tabular}{ccc}
    
     & \rotatebox[origin=c]{90}{ \cite{rozumnyi2021defmo} RGB} \hspace{2pt} &
    \imagecell[1.62]{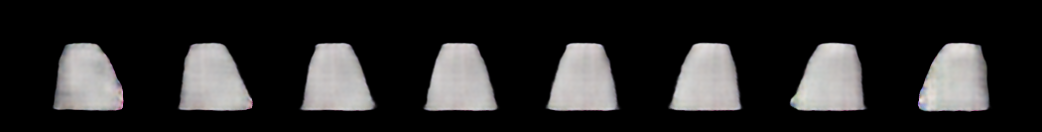} \\ \vspace{-18pt} \\
    \imagecell[0.3]{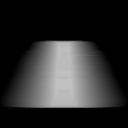} & 
    \rotatebox[origin=c]{90}{\cite{rozumnyi2021defmo} Sil.} \hspace{2pt} &
    \imagecell[1.62]{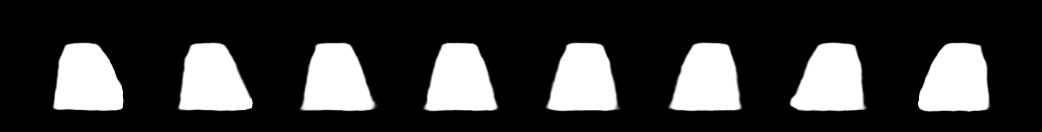} \\ \vspace{-18pt} \\
    $\begin{matrix}
        \text{Input}
    \end{matrix}$ & 
    \rotatebox[origin=c]{90}{G.T.} \hspace{2pt} &
    \imagecell[1.62]{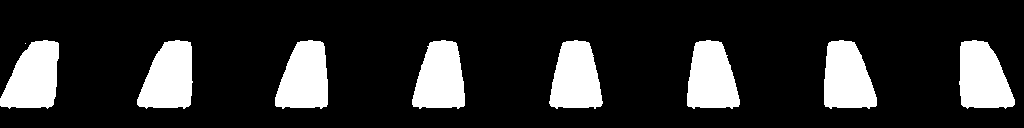} \\ \vspace{-9pt} \\
    \midrule \\ \vspace{-22pt} \\
    
    \vspace{-9pt} & \rotatebox[origin=c]{90}{ \cite{rozumnyi2021defmo} RGB} \hspace{2pt} &
    \imagecell[1.62]{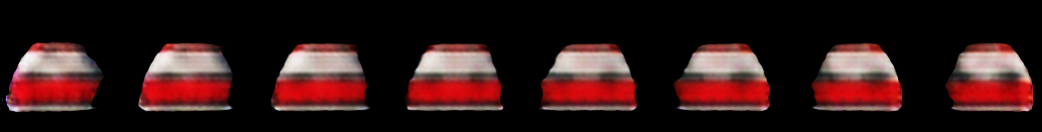} \\ \vspace{-9pt} \\
    \imagecell[0.3]{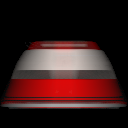} & 
    \rotatebox[origin=c]{90}{\cite{rozumnyi2021defmo} Sil.} \hspace{2pt} &
    \imagecell[1.62]{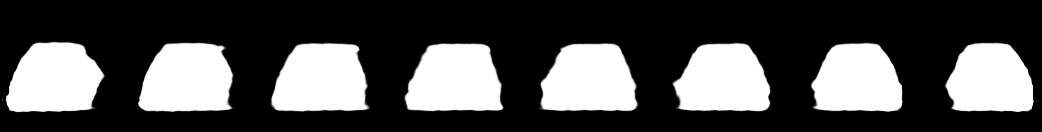} \\ \vspace{-18pt} \\
    $\begin{matrix}
        \text{Input}
    \end{matrix}$ & 
    \rotatebox[origin=c]{90}{G.T.} \hspace{2pt} &
    \imagecell[1.62]{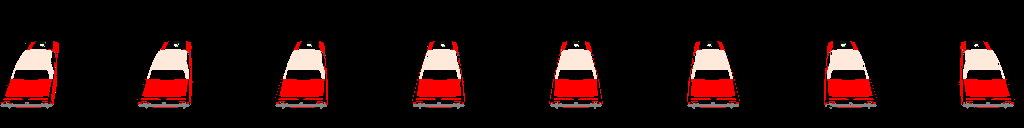} \\ \vspace{-9pt} \\
    \midrule \\ \vspace{-22pt} \\
    
    & \rotatebox[origin=c]{90}{ \cite{rozumnyi2021defmo} RGB} \hspace{2pt} &
    \imagecell[1.62]{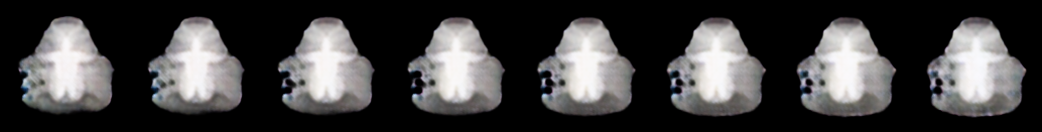} \\ \vspace{-18pt} \\
    \imagecell[0.3]{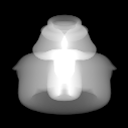} & 
    \rotatebox[origin=c]{90}{\cite{rozumnyi2021defmo} Sil.} \hspace{2pt} &
    \imagecell[1.62]{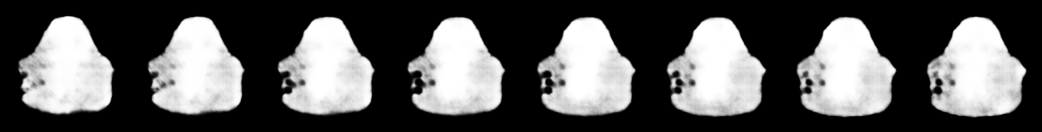} \\ \vspace{-18pt} \\
    $\begin{matrix}
        \text{Input}
    \end{matrix}$ & 
    \rotatebox[origin=c]{90}{G.T.} \hspace{2pt} &
    \imagecell[1.62]{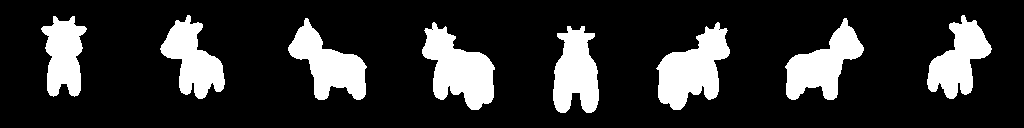} \\ \vspace{-18pt} \\
    
    \end{tabular}
    \end{spacing}
    \caption{
        \textbf{Failures of DeFMO \cite{rozumnyi2021defmo} in Extreme Motion Blur.}
        Each group displays: \textbf{Left.} Input motion-blurred image. \textbf{Right.} Three rows presenting results from DeFMO: \textbf{Top Row.} RGB images predicted by DeFMO at various timestamps. \textbf{Middle Row.} Corresponding static masks (silhouettes) predicted by DeFMO. \textbf{Bottom Row.} Ground Truth (G.T.) static masks at the respective timestamps.
        As illustrated, DeFMO \cite{rozumnyi2021defmo} fails to predict accurate static masks under these extreme motion conditions. 
        This fundamental inaccuracy in DeFMO's prior critically undermines the optimization guidance for SFB \cite{rozumnyi2021shape}, ultimately leading to its reconstruction failures in the challenging scenarios.
    }
    \label{fig:defmo-failure}
\end{figure*}

DeFMO, while generally effective for typical fast motion blur scenarios, fails when confronted with the highly diffused and ambiguous observations generated by ultra-fast motion. 
In such extreme cases, DeFMO struggles to accurately predict the static masks at timestamps along the motion path.
As illustrated in \cref{fig:defmo-failure}, the masks produced by DeFMO for our ultra-fast motion blurred images are often highly inaccurate and entirely non-representative of the underlying object's true silhouette.

Since the DeFMO-predicted static masks serve as a fundamental guidance signal for SFB's shape and motion recovery, their inaccuracy directly propagates through the entire pipeline. 
This makes SFB ineffective for the ultra-fast motion blur reconstruction challenge, despite any richness in the input image data provided.

However, we acknowledge that SFB is a pioneering and important work that significantly advances the field of shape-from-blur by introducing a novel, learning-assisted approach to tackle this challenging inverse problem. 
Our analysis of its limitations merely highlights the unique difficulties posed by extreme motion blur.
While our method demonstrates superior performance in this specific setting, the requirement for input transparency masks will be a limitation. 
We believe that addressing the challenges of extreme motion blur, particularly managing the ambiguity without explicit transparency, presents a significant and fertile ground for future research. 

\subsection{Analysis of Nvdiffrast's Gradient Computation}
\label{sec:supp:nvdiffrast-gradients-concise}

In the main paper (\Cref{para:failure-cases-for-nvdiffrast}), we demonstrated Nvdiffrast's limited performance in reconstructing shapes from extreme motion blur. This might stem from a fundamental difference in how geometry gradients are computed.

Nvdiffrast primarily derives geometry gradients from localized, pixel-wise anti-aliasing signals along triangle edges. 
This means a vertex's influence on the gradient is concentrated on a few pixels it directly affects. 
While efficient for rendering, these localized gradients are insufficient for optimizing shape deformations from highly ambiguous, severely blurred input images. 
It leads to slow convergence or catastrophic failures due to a lack of meaningful gradient signals.

In contrast, our method, built on from SoftRas \cite{liu2019softras}, enables each vertex to influence many pixels across a broader image region, effectively generating global and smoothed gradients. 
Such gradients provide a more stable signal for shape optimization. This fundamental difference in gradient computation contributes to robust 3D shape recovery in our challenging scenarios.

\section{More Results}
\label{sec:supp:more-results}

In this section, we present additional experimental results.

\subsection{Parabolic Recovery}
\label{sec:supp:parabolic-recovery}

We provide an evaluation on a more complex motion type: combined translational and rotational motion along a parabolic trajectory.

In this experiment, each vertex $P_0 = (x_0, y_0, z_0)$ undergoes a two-step transformation to define its motion path over time $t \in [0, 1]$. The vertex is first rotated around the Y-axis by an angle $\theta(t) = \pi t$, where $t \in [0, 1]$. The intermediate rotated position $P_{rot}(t)$ is given by $P_{rot}(t) = \mathbf{R}y(\pi t) P_0$. Specifically, for $P_{rot}(t) = (x_{rot}(t), y_{rot}(t), z_{rot}(t))$:
\begin{equation}
\begin{cases}
x_{rot}(t) = x_0 \cos(\pi t) + z_0 \sin(\pi t) \\
y_{rot}(t) = y_0 \\
z_{rot}(t) = -x_0 \sin(\pi t) + z_0 \cos(\pi t).
\end{cases}
\end{equation}
Subsequently, a translation vector $T(t) = (T_x(t), T_y(t), T_z(t))$ is applied to the rotated position $P_{rot}(t)$. Let $s(t) = 0.5 - t$. The components of this translation vector are:
\begin{equation}
\begin{cases}
T_x(t) = s(t) \\
T_y(t) = -4 s(t)^2 + 0.5 \\
T_z(t) = s(t).        
\end{cases}
\end{equation}
The final position $P(t) = (x(t), y(t), z(t))$ at time $t$ is then $P(t) = P_{rot}(t) + T(t)$. Specifically:
\begin{equation}
\begin{cases}
x(t) = x_{rot}(t) + (0.5 - t) \\
y(t) = y_{rot}(t) + (-4(0.5 - t)^2 + 0.5) \\
z(t) = z_{rot}(t) + (0.5 - t).
\end{cases}
\end{equation}

\begin{figure}
    \centering
    \setlength\tabcolsep{1pt}
    \renewcommand{\arraystretch}{1.0}
    \begin{tabular}{cc}
        \includegraphics[width=0.21\textwidth]{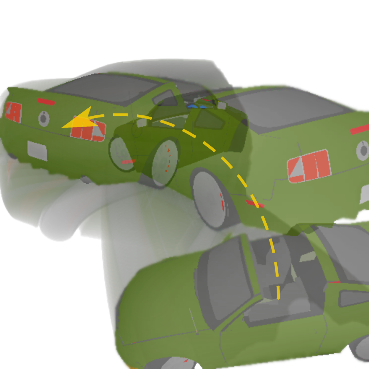} &
        \includegraphics[width=0.25\textwidth]{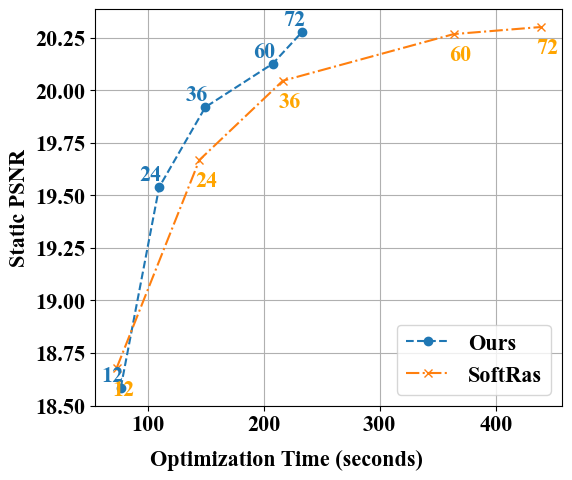} \\
        (a) Illustration & (b) Evaluation
    \end{tabular}
    \caption{
        Shape recovery for complex motion trajectories (parabolic translation + rotation). Labels indicate the corresponding number of samples. We achieve better performance than SoftRas.
    }
    \label{fig:eval-complex}
\end{figure}

Illustration and evaluation results are shown in \cref{fig:eval-complex}.

\subsection{Accelerating Existing Pipelines}
\label{sec:supp:combine34}

We further demonstrate the potential of our method as an accelerator for existing optimization-based inverse rendering pipelines.
We integrate our method into \cite{rozumnyi2021shape}, replacing its original rendering component.
We evaluate its performance by comparing total optimization time and reconstruction quality (TIoU, PSNR, SSIM) against the original \cite{rozumnyi2021shape}.

Quantitative comparison results are presented in \cref{tab:comparison-sfb-rw}.
Our integration reduces the optimization time while maintaining comparable reconstruction quality.
These results demonstrate our method's effectiveness in accelerating existing inverse rendering pipelines, thereby enabling them to tackle complex, real-world motion blur scenarios with greater efficiency.

Moreover, these results also demonstrate that our method can leverage existing pipelines (\eg, \cite{rozumnyi2021shape}) to handle diverse real-world scenarios.

\begin{table}[thbp]
\centering
\setlength\tabcolsep{3pt}
\begin{tabular}{cccccc}
\toprule
\multirow{2}{*}{\textbf{Method}} & \multicolumn{3}{c}{\textbf{Falling Objects Dataset}} & \multirow{2}{*}{\textbf{Time (s)}} & \\
& TIoU $\uparrow$ & PSNR $\uparrow$ & SSIM $\uparrow$ & \\
\midrule
\text{\cite{rozumnyi2021shape}} & 0.678 & 26.133 & 0.736 & 60.663 & \\
\text{\cite{rozumnyi2021shape}} + Ours & 0.678 & 26.010 & 0.731 & 47.227 & \\
\bottomrule
\end{tabular}
\vspace{-7pt}
\caption{
Evaluation on the FMO real-world benchmark. 
Note time contains both rendering and data processing steps. 
Our solver can be integrated into \cite{rozumnyi2021shape}'s pipeline, providing faster optimization with comparable performance. 
``+ Ours'' denotes replacing the Kaolin DIB-R rasterizer with ours but retaining the texture mapping module.
Since the time cost for per template mesh remains similar, as reported in \cite{rozumnyi2021shape}, we follow the best settings but use the Voronoi sphere as the template mesh only, and split the trajectory into 8 segments in our method. 
We have tried our best to make reproduction (the top row) but small discrepancy in performance still exists, which might impact little on our time-oriented evaluation.
Results show that with the complement of out method (the bottom row), a speedup can be achieved without significant losses of performance. 
In addition, Kaolin DIB-R is a highly-optimized CUDA renderer, while our method is lack of low level CUDA optimization. We believe that with more such optimization, our method can achieve a more significant acceleration. 
}
\label{tab:comparison-sfb-rw}
\vspace{-13pt}
\end{table}

\section{Detailed Limitations and Future Work}
\label{sec:supp:limitations}

In this section, we provide a detailed discussion on the limitations of our method and potential directions for future research.

\paragraph{Dependency on Known Motion and Poses}
Similar to many inverse rendering approaches, our current optimization pipeline requires known camera intrinsics, poses, and motion information.
In unconstrained settings, obtaining these parameters can be challenging.
A promising direction is to integrate our differentiable renderer with motion estimation modules (\eg, \cite{rozumnyi2021shape}) to jointly estimate motion trajectories and shape from the input image.
We present a preliminary trial of this integration in \cref{sec:supp:combine34}.

\paragraph{Motion Linearity Assumption}
Our fast barycentric solver assumes that motion within each time segment is linear.
While this approximation holds for short exposure times, highly complex non-linear motions may introduce errors.
Addressing this would require modeling higher-order motion trajectories or employing finer temporal segmentation, which we leave for future optimization.

\paragraph{Photometric Assumptions}
Our rendering model assumes a linear photometric relationship between the scene radiance and pixel intensity.
However, real-world camera ISPs (Image Signal Processors) typically apply non-linear tone mapping curves (\eg, Gamma correction) to compress high dynamic range data for display.
Furthermore, high-speed photography often necessitates high ISO settings to compensate for short exposure times (if not blurring intentionally), or operates in low-light conditions where the signal-to-noise ratio is low.
Our current model does not explicitly account for non-linear camera response functions or sensor noise.
Future work could incorporate learnable camera response functions (CRFs) and noise modeling to enhance reconstruction robustness in raw, in-the-wild footage.

\end{document}